\documentclass[10pt,twocolumn,letterpaper]{article}

\usepackage{cvpr}
\usepackage{times}
\usepackage{epsfig}
\usepackage{graphicx}
\usepackage[export]{adjustbox}
\usepackage{amsmath}
\usepackage{amssymb}
\usepackage{multirow}
\usepackage{array}
\usepackage{enumitem}
\usepackage{float}

\usepackage[pagebackref=true,breaklinks=true,colorlinks,bookmarks=false]{hyperref}

\cvprfinalcopy 

\def\cvprPaperID{7922} 

\ifcvprfinal\fi

\usepackage{overpic}
\usepackage{amssymb}
\usepackage{amsmath}
\usepackage{graphicx}

\newcommand{\repose}{\textbf{RePose}~}
\newcommand{\reposenospace}{\textbf{RePose}} 

\usepackage{color}
\definecolor{turquoise}{cmyk}{0.65,0,0.1,0.3}
\definecolor{purple}{rgb}{0.65,0,0.65}
\definecolor{dark_green}{rgb}{0, 0.5, 0}
\definecolor{green}{rgb}{0, 1.0, 0}
\definecolor{orange}{rgb}{0.8, 0.6, 0.2}
\definecolor{red}{rgb}{0.8, 0.2, 0.2}
\definecolor{blueish}{rgb}{0.0, 0.7, 1}
\definecolor{light_gray}{rgb}{0.7, 0.7, .7}
\definecolor{pink}{rgb}{1, 0, 1}

\newcommand{\hide}[1]{{}} 

\usepackage{currfile}

\usepackage{amsmath}

\usepackage{blindtext}

\renewcommand{\paragraph}[1]{{\vspace{1em}\noindent \textbf{#1.}}}

\newcommand{\kernel}{\mathcal{K}}
\newcommand{\stride}{\mathcal{S}}
\newcommand{\filters}{\mathcal{F}}
\newcommand{\keypoint}{\mathbf{p}}
\newcommand{\concat}{\mathbf{c}}
\newcommand{\feature}{\mathbf{f}}

\begin{document}

\title{RePose: Learning Deep Kinematic Priors for Fast Human Pose Estimation}

\author{Hossam Isack$^{1,2}$
\and Christian Haene$^{1}$
\and Cem Keskin $^{1}$
\and Sofien Bouaziz$^{1}$
\and Yuri Boykov$^{2}$
\and Shahram Izadi$^{1}$ \vspace{2ex} \\ 
$^{1}$Google Inc.
\and Sameh Khamis$^{1}$ \vspace{2ex} \\
$^{2}$University Of Waterloo
}

\maketitle

\begin{abstract}
We propose a novel efficient and lightweight model for human pose estimation from a single image. Our model is designed to achieve competitive results at a fraction of the number of parameters and computational cost of various state-of-the-art methods. To this end, we explicitly incorporate part-based structural and geometric priors in a hierarchical prediction framework. At the coarsest resolution, and in a manner similar to classical part-based approaches, we leverage the kinematic structure of the human body to propagate convolutional feature updates between the keypoints or body parts. Unlike classical approaches, we adopt end-to-end training to learn this geometric prior through feature updates from data. We then propagate the feature representation at the coarsest resolution up the hierarchy to refine the predicted pose in a coarse-to-fine fashion. The final network effectively models the geometric prior and intuition within a lightweight deep neural network, yielding state-of-the-art results for a model of this size on two standard datasets, Leeds Sports Pose and MPII Human Pose.
\end{abstract}

\vspace{-1.5ex}
\section{Introduction}

The recent progress in machine learning techniques has allowed computer vision to work beyond bounding box estimation and solve tasks necessitating a fine-grained understanding of scenes and people. In particular, estimating human poses in the wild has been an area of research which has seen tremendous progress thanks to the development of large convolutional neural networks (CNN) that can efficiently reason about human poses under large occlusions or poor image quality. However, most of the best-performing models have 30 to 60 millions parameters, prohibiting their usage in systems where compute and power are constrained, \eg mobile phones. Decreasing the size of these deep models most often results in a drastic loss of accuracy, making this a last resort option to improve the efficiency. One might then rightfully ask: \emph{could some form of structural priors be used to reduce the size of these models while keeping a good accuracy?}

\begin{figure}[t!]
\newcommand{\leedsa}{0.11}
\begin{tabular*}{\textwidth}{*{5}{c@{\hspace{0.2ex}}}c}
    \multirow{1}{*}[12ex]{\rotatebox[origin=c]{90}{\footnotesize Pre-updates}}& 
    \multirow{1}{*}[14ex]{\rotatebox[origin=c]{90}{\footnotesize Coarsest resolution}} &
    \includegraphics[height=\leedsa\textheight]{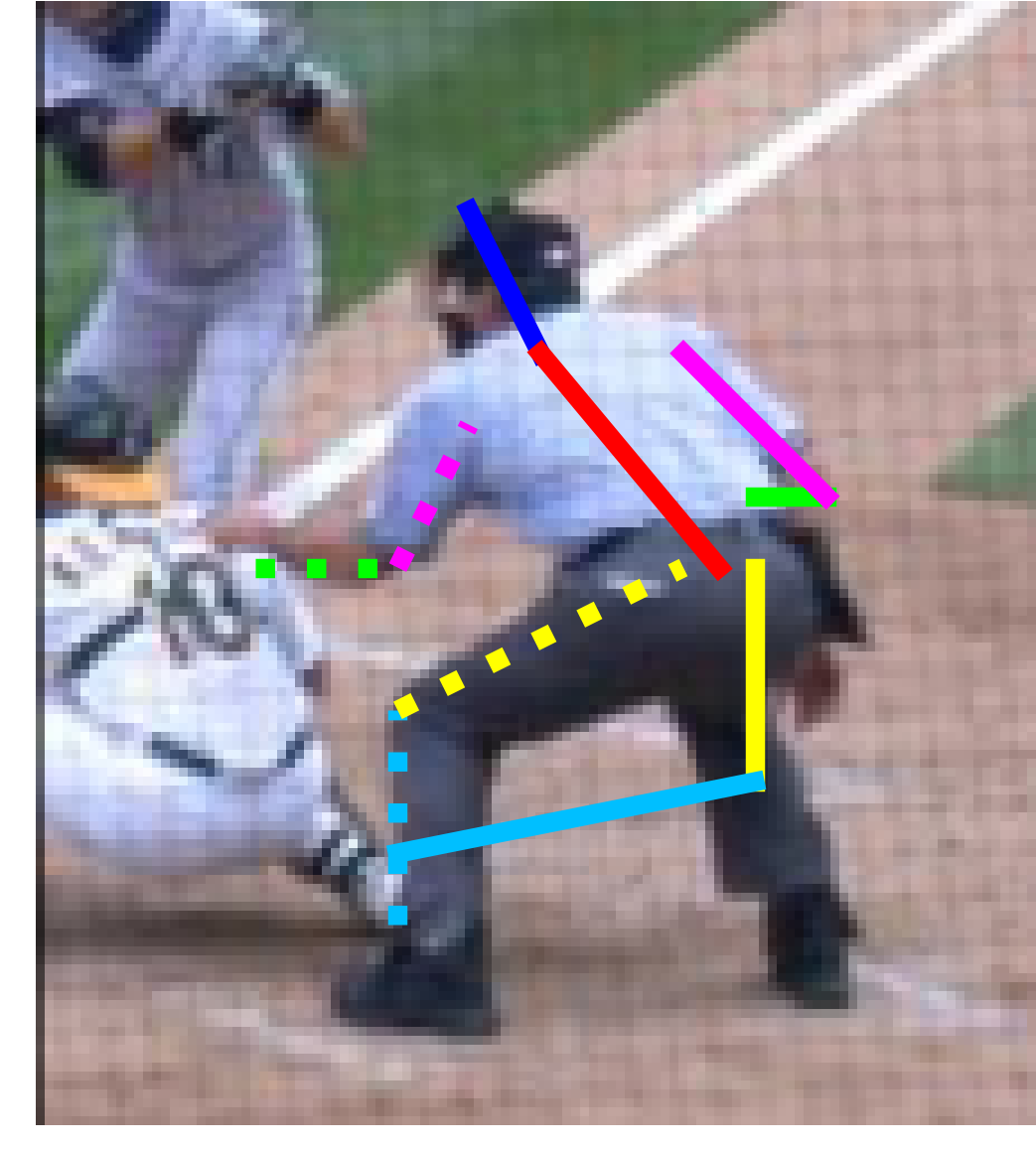} &
    \includegraphics[height=\leedsa\textheight]{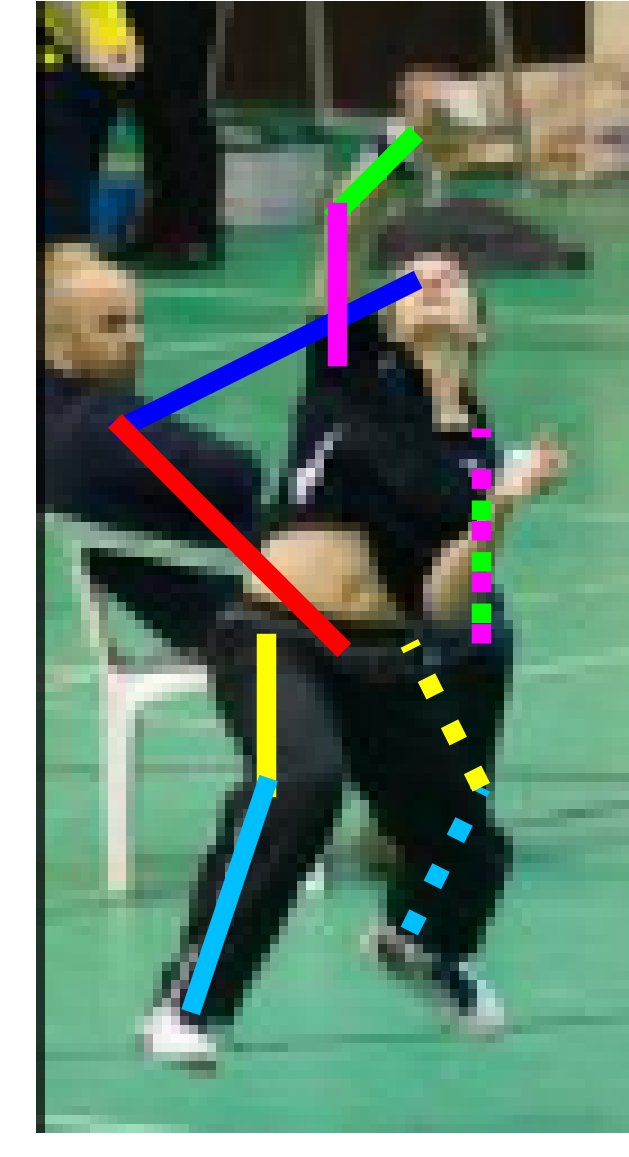} &
    \includegraphics[height=\leedsa\textheight]{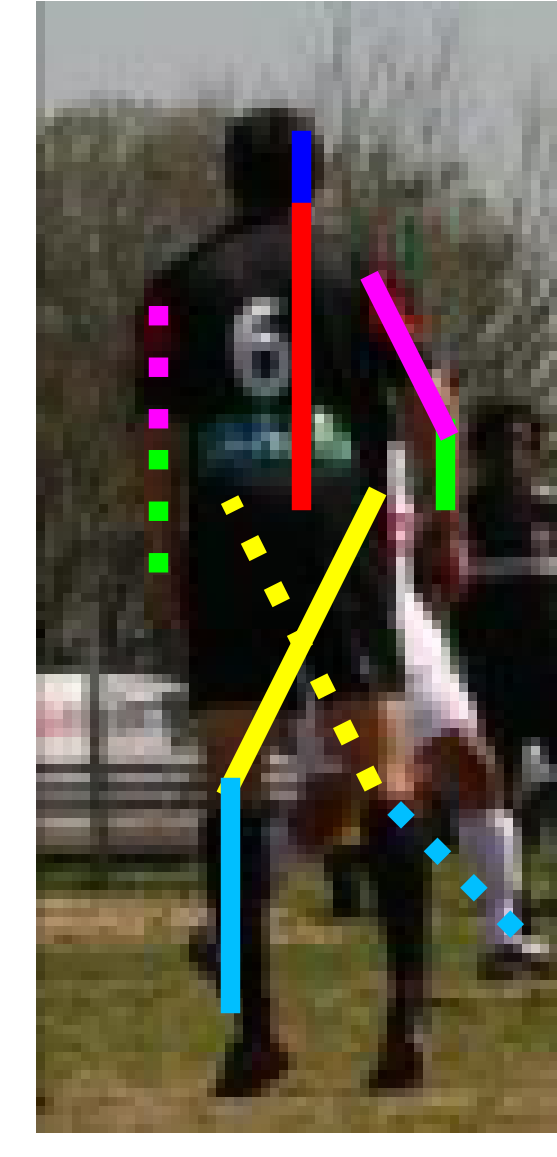} &
    \includegraphics[height=\leedsa\textheight]{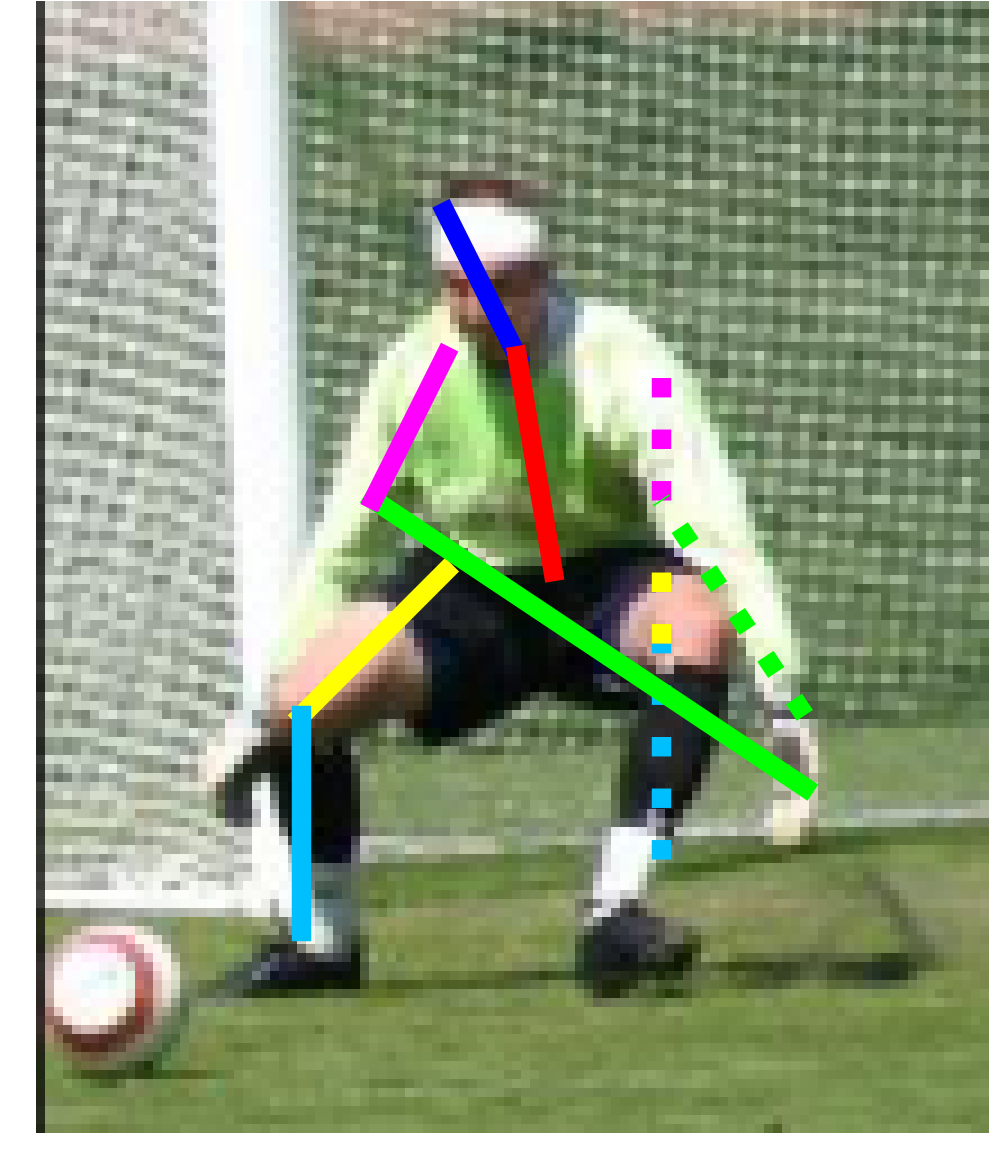} \\
    \multirow{1}{*}[12ex]{\rotatebox[origin=c]{90}{\footnotesize Post-updates}}& 
    \multirow{1}{*}[14ex]{\rotatebox[origin=c]{90}{\footnotesize Coarsest resolution}} &
    \includegraphics[height=\leedsa\textheight]{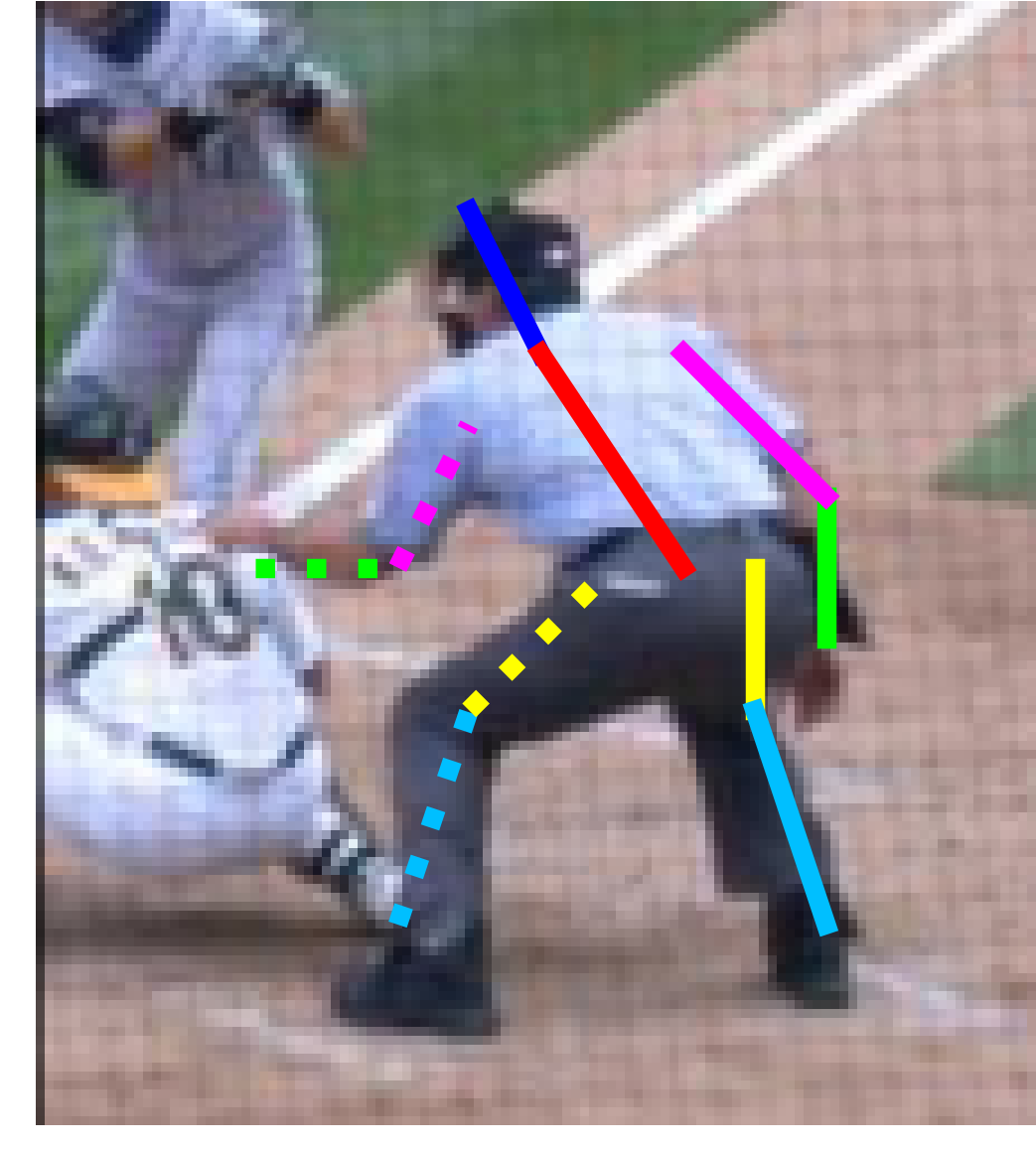} &
    \includegraphics[height=\leedsa\textheight]{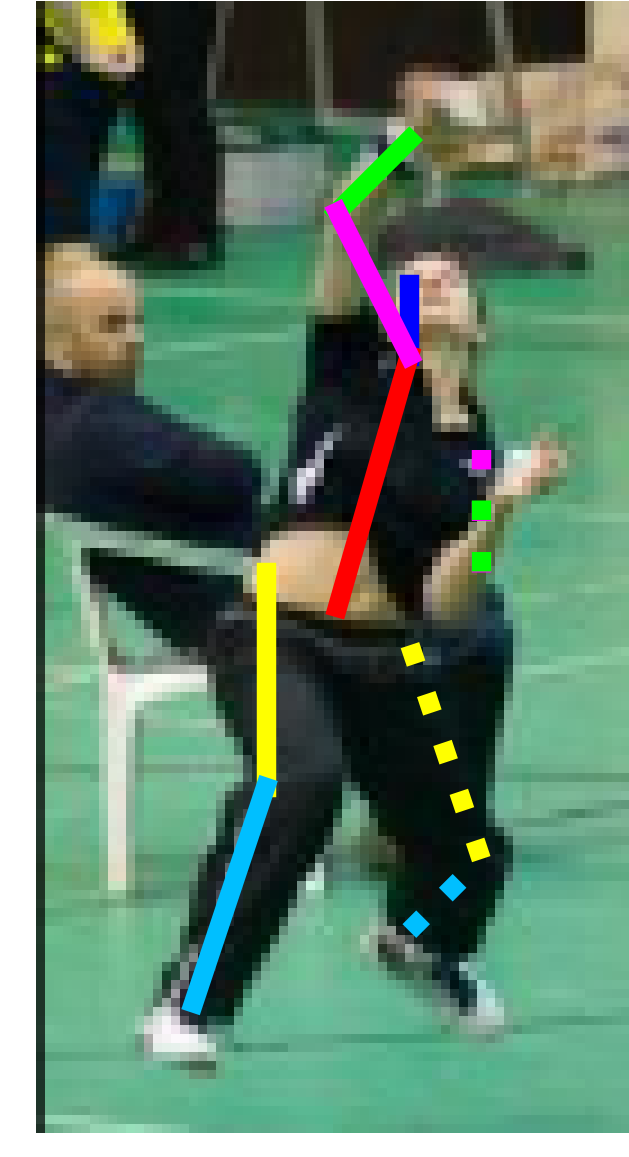} &
    \includegraphics[height=\leedsa\textheight]{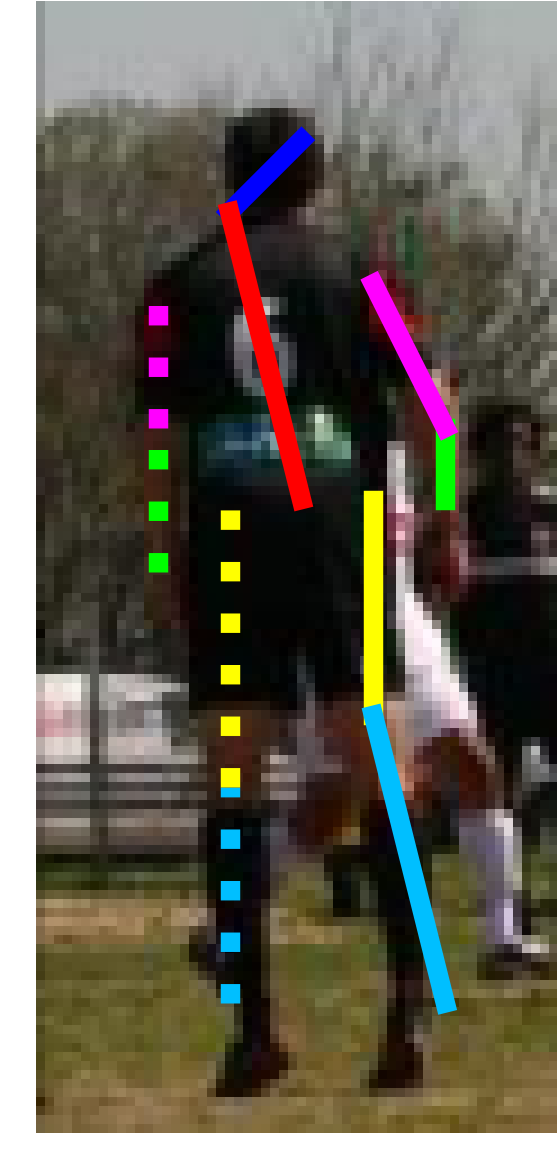} &
    \includegraphics[height=\leedsa\textheight]{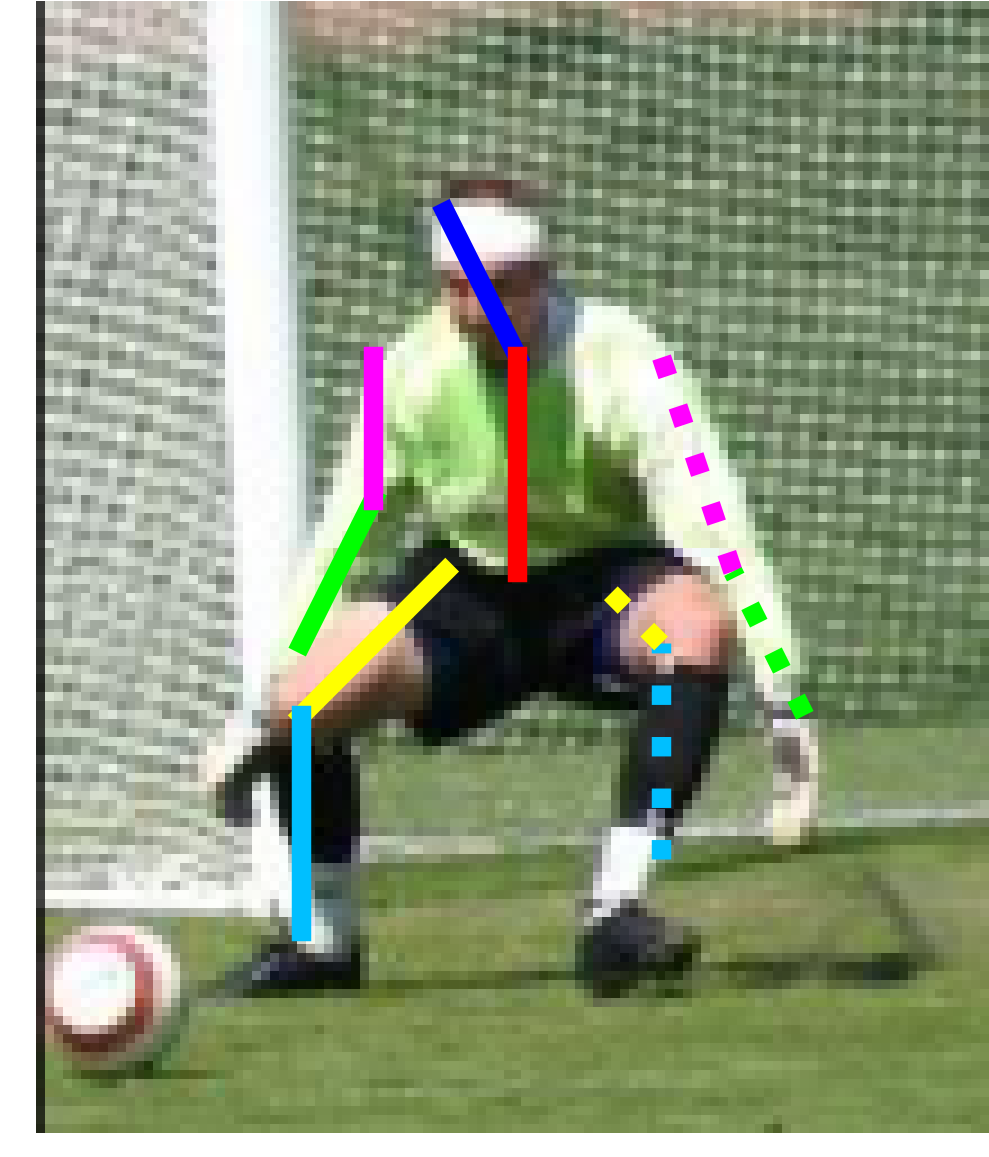} \\    
    \multirow{1}{*}[12ex]{\rotatebox[origin=c]{90}{\footnotesize Final prediction}}& 
    \multirow{1}{*}[12ex]{\rotatebox[origin=c]{90}{\footnotesize Finest resolution}} &
    \includegraphics[height=\leedsa\textheight]{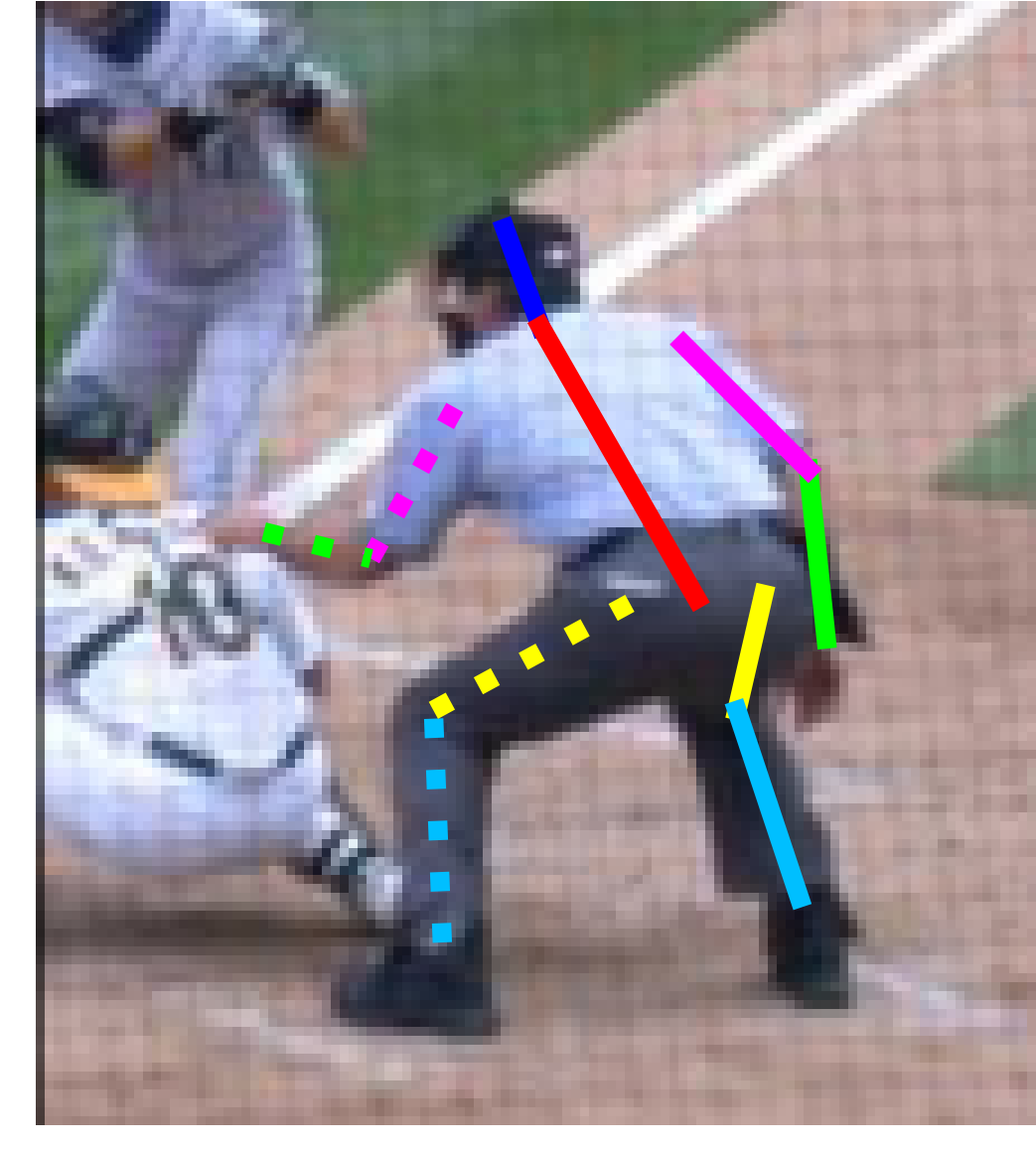} &
    \includegraphics[height=\leedsa\textheight]{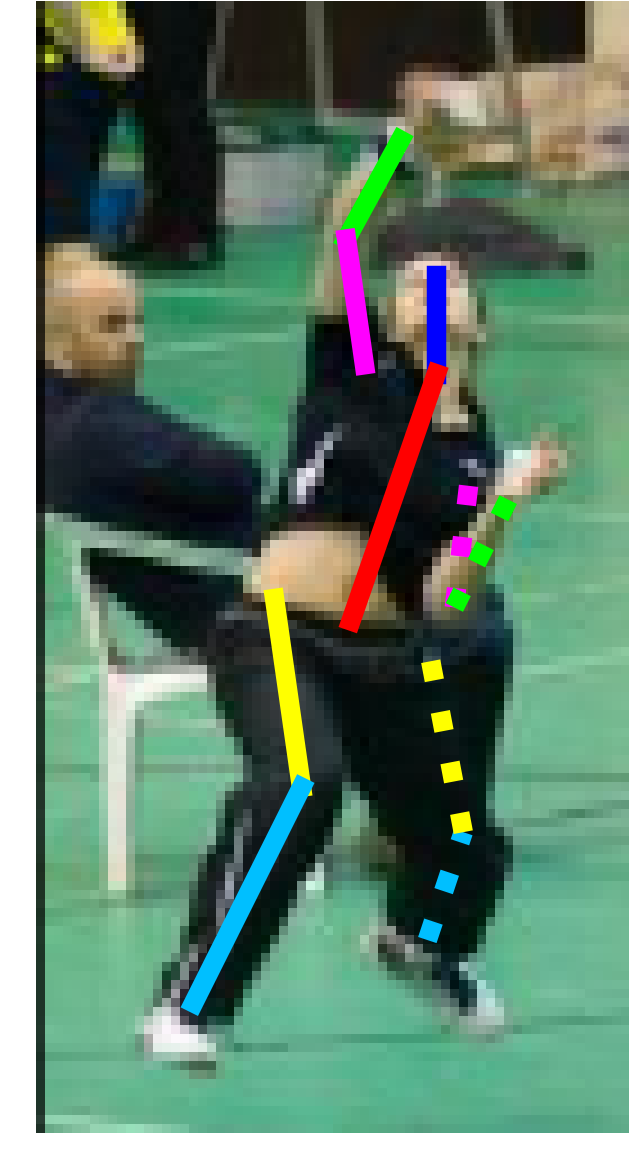} &
    \includegraphics[height=\leedsa\textheight]{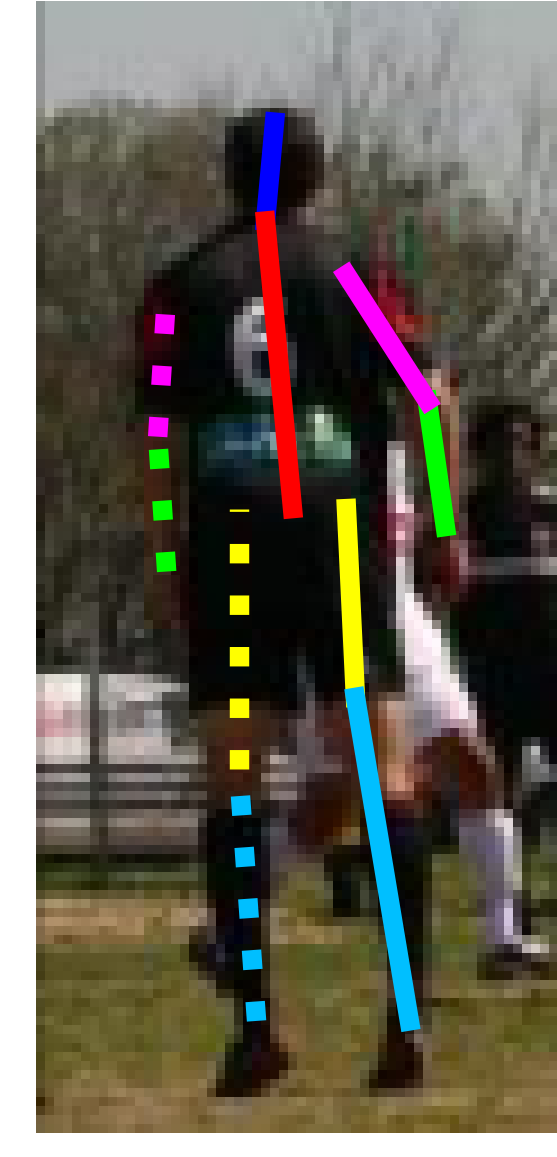} &
    \includegraphics[height=\leedsa\textheight]{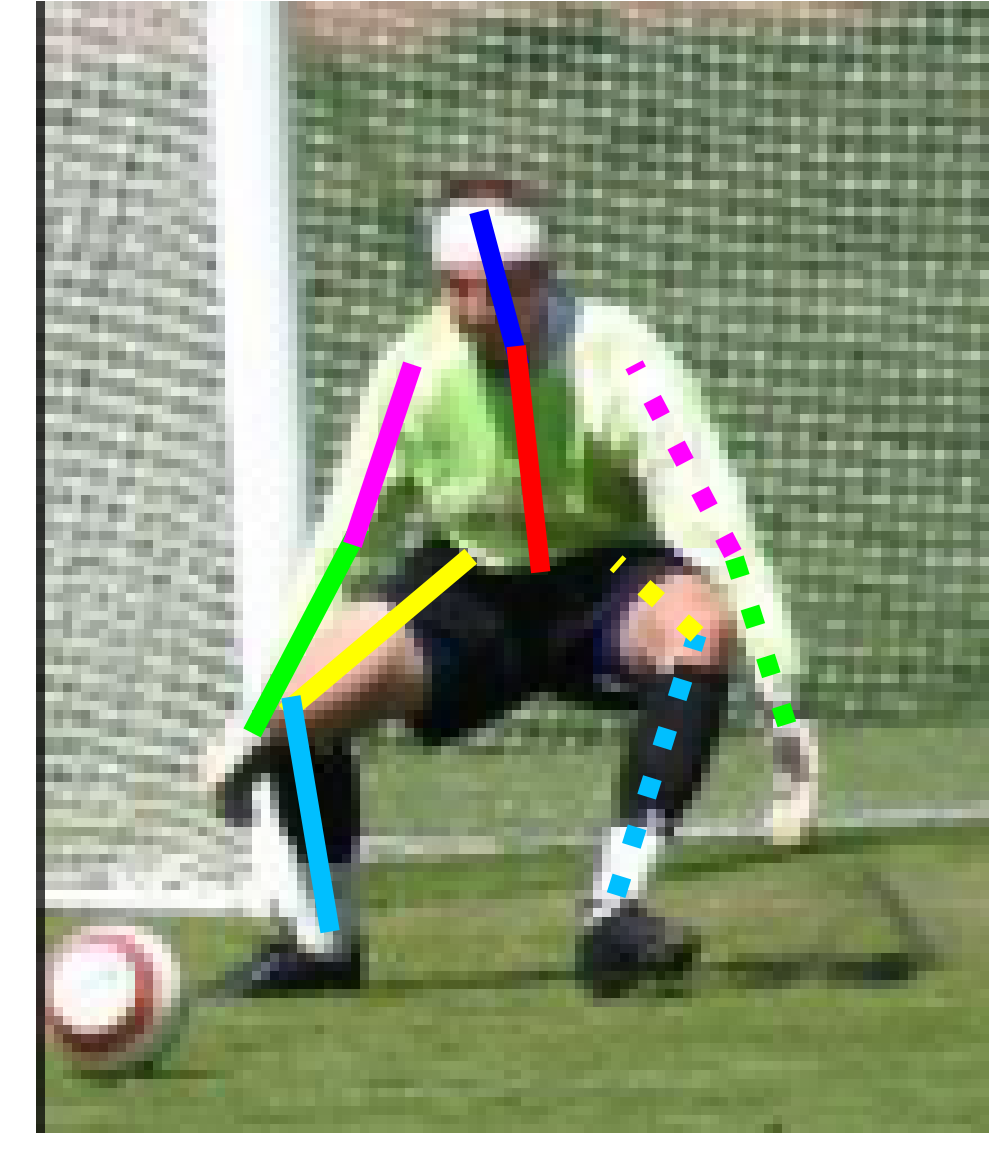} \\
    &
    \multirow{1}{*}[12ex]{\rotatebox[origin=c]{90}{\footnotesize Ground truth}} &
    \includegraphics[height=\leedsa\textheight]{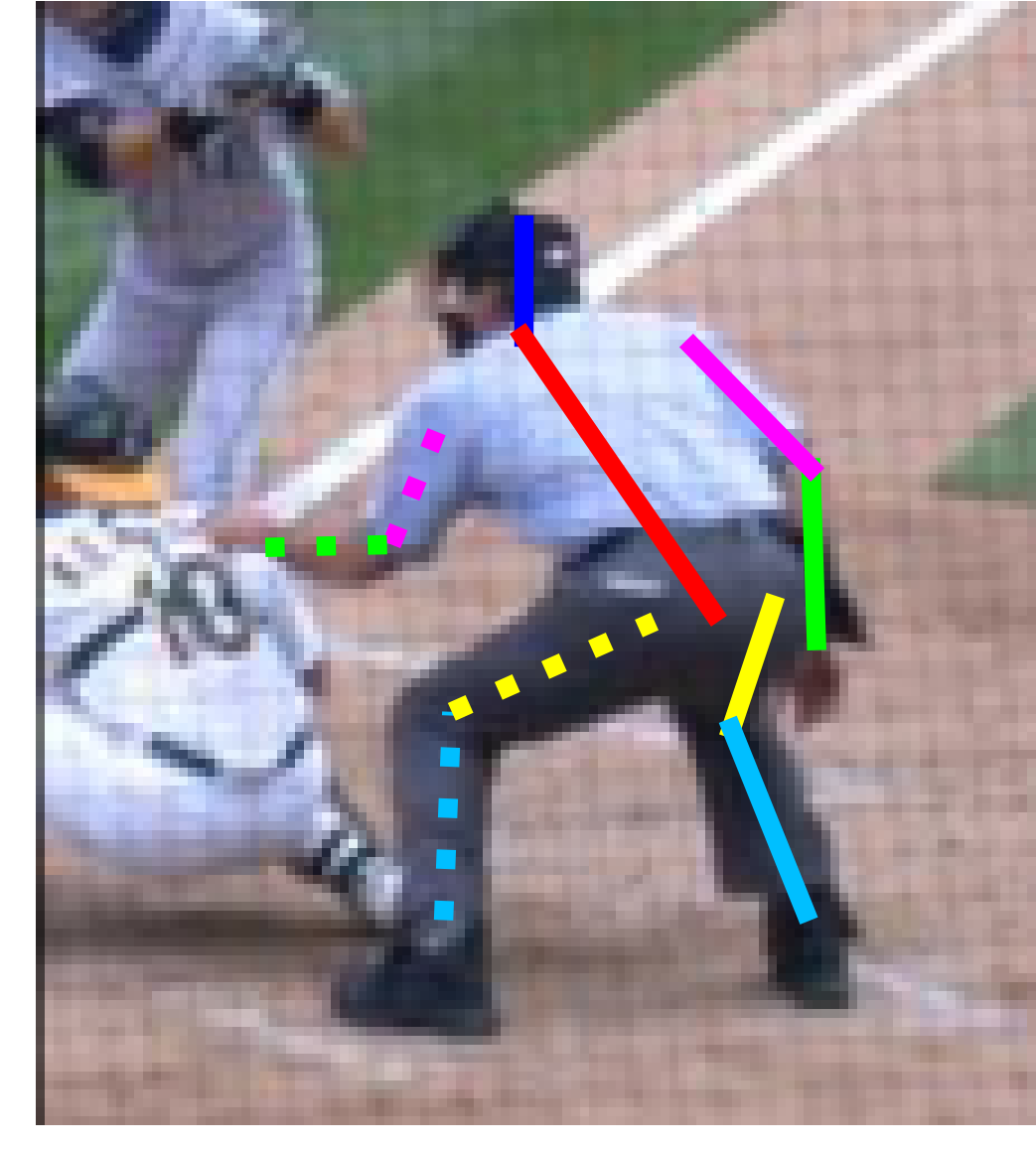} &
    \includegraphics[height=\leedsa\textheight]{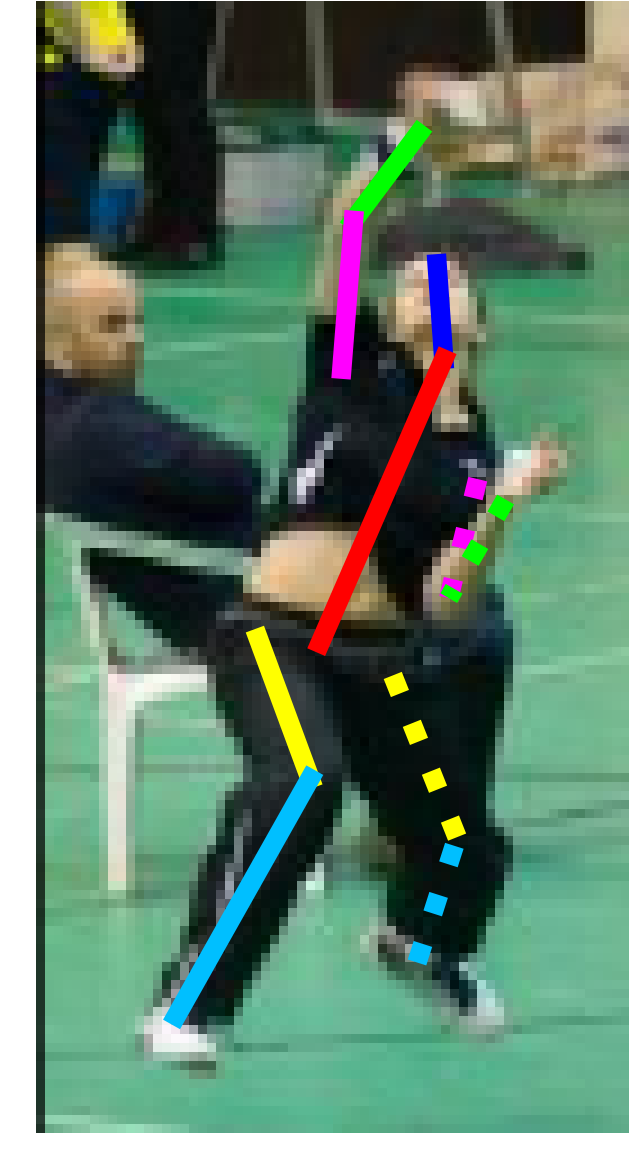} &
    \includegraphics[height=\leedsa\textheight]{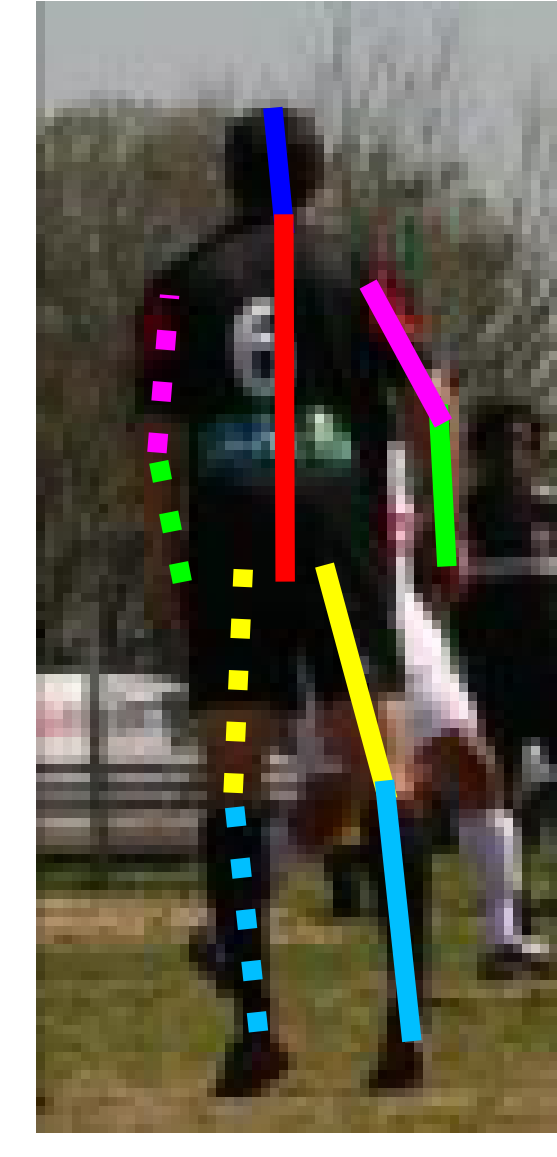} &
    \includegraphics[height=\leedsa\textheight]{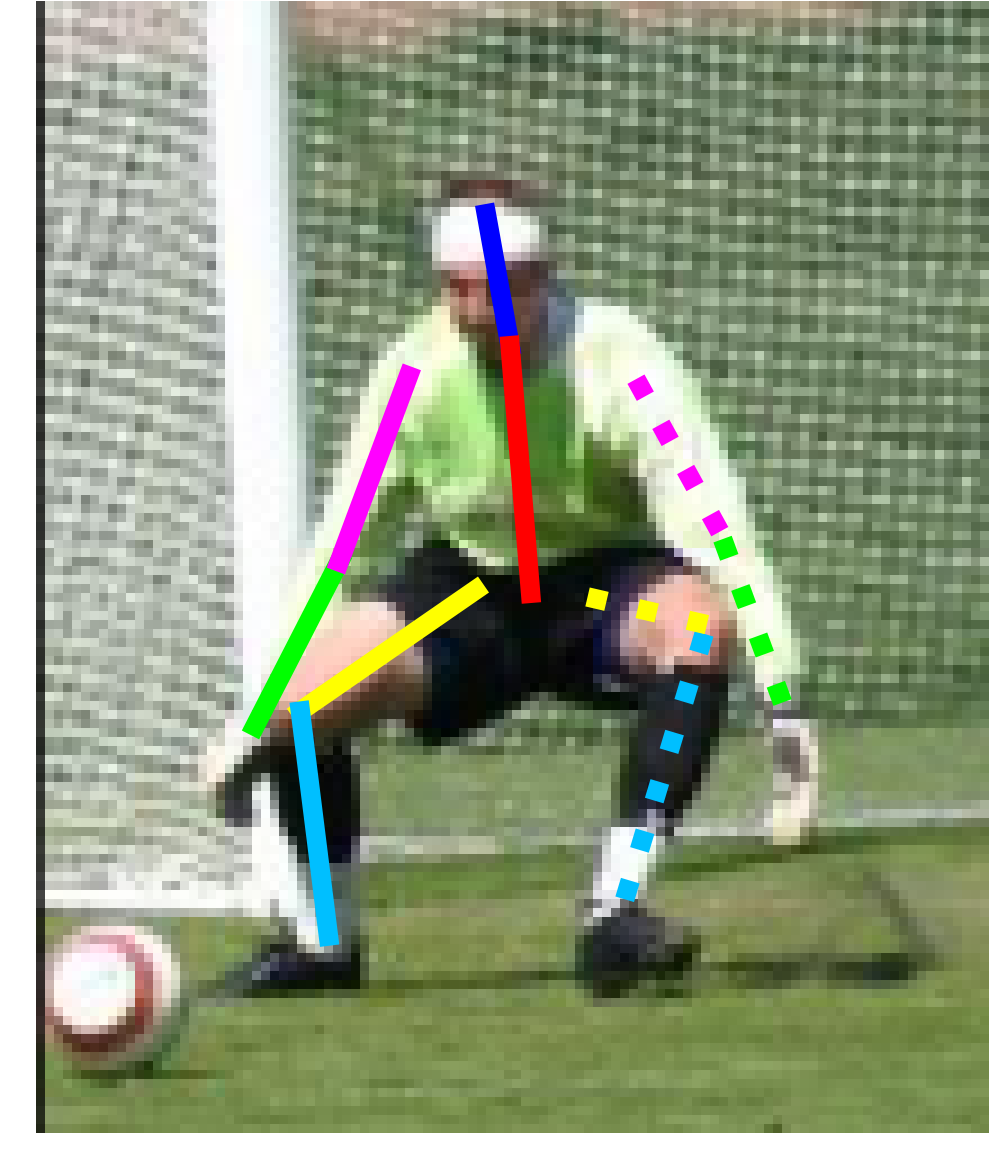} \\
\end{tabular*}
\caption{Our approach employs learned structural and geometric priors at a very coarse resolution using a CNN that refines the prediction hierarchically.
Top row shows the prediction before keypoint-specific features are updated.
Second row shows prediction after features are updated in a predefined order, where high confidence keypoints influence poorly localized ones.
Third row shows the final refined pose prediction at the finest resolution.
}
\label{fig:teaser_image}
\end{figure}

A key physiological property that is shared across humans is the kinematic structure of the body. Our understanding of this structure allows us to accurately estimate the location of all body parts of other people even under occlusions. This naturally brings up the question: \emph{could a deep neural network (DNN) make use of this kinematic structure to achieve highly accurate human body pose estimation while keeping the model complexity small?}

In fact, we are not the first to wonder about utilizing the kinematic structure of the body in a machine learning model. Indeed, early computer vision techniques for human pose estimation were using part-based graphical models \cite{felzenszwalb2005pictorial}.
While the kinematic structure of the human body is well defined, the distribution of joint distances and angles is complex and hard to model explicitly when projected on 2D. Therefore, these earlier approaches often simplified this distribution for example to Gaussian potentials~\cite{felzenszwalb2005pictorial}. In contrast, our approach only encodes the kinematic structure in a network architecture and lets the network learn the priors from data.

Some more recent deep learning approaches have also made use of a kinematic prior by approximating a maximum a posteriori (MAP) solver within a neural network \cite{de2018deep,deformableparts16}, typically through implementing the max-sum algorithm within the network. Running a MAP solver is computationally expensive and requires an explicit description of the distribution. Our approach avoids these issues by employing a kinematically structured network. This allows us to incorporate the structural prior without incurring the computational penalty. We encode this structure at a coarse resolution where a small receptive field is large enough to capture and spatially correlate neighbouring pose keypoints. Moreover, employing the kinematic feature update module at a coarse resolution keeps our network lightweight. Finally, our method successfully refines the predicted pose hierarchically through a feature pyramid until the finest resolution is reached. Figure \ref{fig:teaser_image} illustrates how the predicted pose improves throughout the various updates.

To summarize, our main contributions are as follows:
\begin{enumerate}
    \item A novel network architecture that encodes the kinematic structure  via feature updates at coarse resolution, without the need for including any approximate MAP inference steps.
    \item A lightweight kinematic feature update module that achieves a significant improvement in accuracy, while only adding a small number of learnable parameters compared to state-of-the-art approaches.
    \item Extensive evaluation showing state-of-the art results on the LSP dataset and competitive results on the MPII dataset using a lightweight network without using model compression techniques such as distillation.
\end{enumerate}

\section{Related Work}

Human pose estimation is a fundamental problem in computer vision and an active field of research. Early attempts to solving this problem were based on kinematically inspired statistical graphical models, \eg~\cite{felzenszwalb2005pictorial,tompson2014joint,articulatedgraphmodel14,deformableparts16}, for modeling geometric and structural priors between keypoints, \eg~elbow and wrist, in 2D images.

These techniques either imposed limiting assumptions on the modeled distribution, or relied on sub-optimal methods for solving such graphical models within a deep learning framework~\cite{tompson2014joint,articulatedgraphmodel14,deformableparts16}. For example, \cite{felzenszwalb2005pictorial} assumed that the distance between a pair of keypoints could be modeled as a Gaussian distribution. Although efficient optimization methods exist for such a model, in practice the model is fairly simple and does not capture the complex global-relation between keypoints especially in 2D image space.

More recent approaches such as~\cite{deformableparts16} applied loopy belief propagation, without any guarantees of optimality or convergence, in an effort to infer the MAP-estimate of a pose within a deep learning framework. The used loopy belief propagation in~\cite{deformableparts16} or dynamic programming in~\cite{articulatedgraphmodel14} are computationally expensive. Furthermore, such networks are harder to train in general~\cite{differentiableDP2018,niculae2017regularized}, and the inferred MAP-estimate is not informative during the early stages of training when networks are learning to extract low level features. 

The top performing pose estimation methods are based on DNNs~\cite{tang2019does,avddataugment18,featurepyramids17,attention17,stackedhourglass16,poseGAN2017}, which are capable of modeling complex geometric and appearance distributions between keypoints.
In search for better performance on benchmarks, novel architectures and strategies were devised, such as adversarial data augmentation~\cite{avddataugment18}, feature-pyramids \cite{featurepyramids17}, pose GANs~\cite{poseGAN2017} and network stacking~\cite{stackedhourglass16}, which is a commonly used strategy that other methods~\cite{avddataugment18,featurepyramids17,attention17,bulat2016human,poseGAN2017} build on due to its simplicity and effectiveness. 

In general, better pose estimates could be reached by successively refining the estimated pose. Carreira \etal~\cite{iterfeedback16} refined their initially estimated keypoints' heatmaps by using a \textit{single} additional refinement network and repeatedly using it to predict a new pose estimate given the current one. 
The stacking used in~\cite{stackedhourglass16} could be seen as \textit{unrolling} of Carreira \etal~\cite{iterfeedback16} refinement approach, where there are seven consecutive refinement networks that do not share weights.
Although refinement unrolling achieves significantly better results than a single repeated refinement step~\cite{iterfeedback16}, it is very expensive, e.g.~\cite{stackedhourglass16},~\cite{featurepyramids17} and~\cite{deepercutECCV16} require 18/38~\cite{avddataugment18}, 28 and 60+ million parameters, respectively.

There are DDNs that aim to learn spatial priors between keypoints without resolving to MAP inference approximation. In \cite{tang2019does} keypoints are clustered into sets of correlated keypoints and each set has its independent features, \eg knee features do not directly affect hip features. The clustering was based on mutual information measure, but the clustering threshold was heuristically chosen. In contrast, RePose allow neighbouring keypoints to directly influence each others features. Furthermore, \cite{tang2019does} relies heavily on network stacking, while stacking \textit{slightly} improves RePose's accuracy.
Unlike RePose, \cite{chu2016structured} does not apply hierarchical pose refinement and relies on a handcrafted post-processing step to suppress false positives in heatmaps. Finally, \cite{tang2019does,chu2016structured} are  significantly larger networks than RePose.

\begin{figure*}[!ht]
\centering
\includegraphics[width=\textwidth]{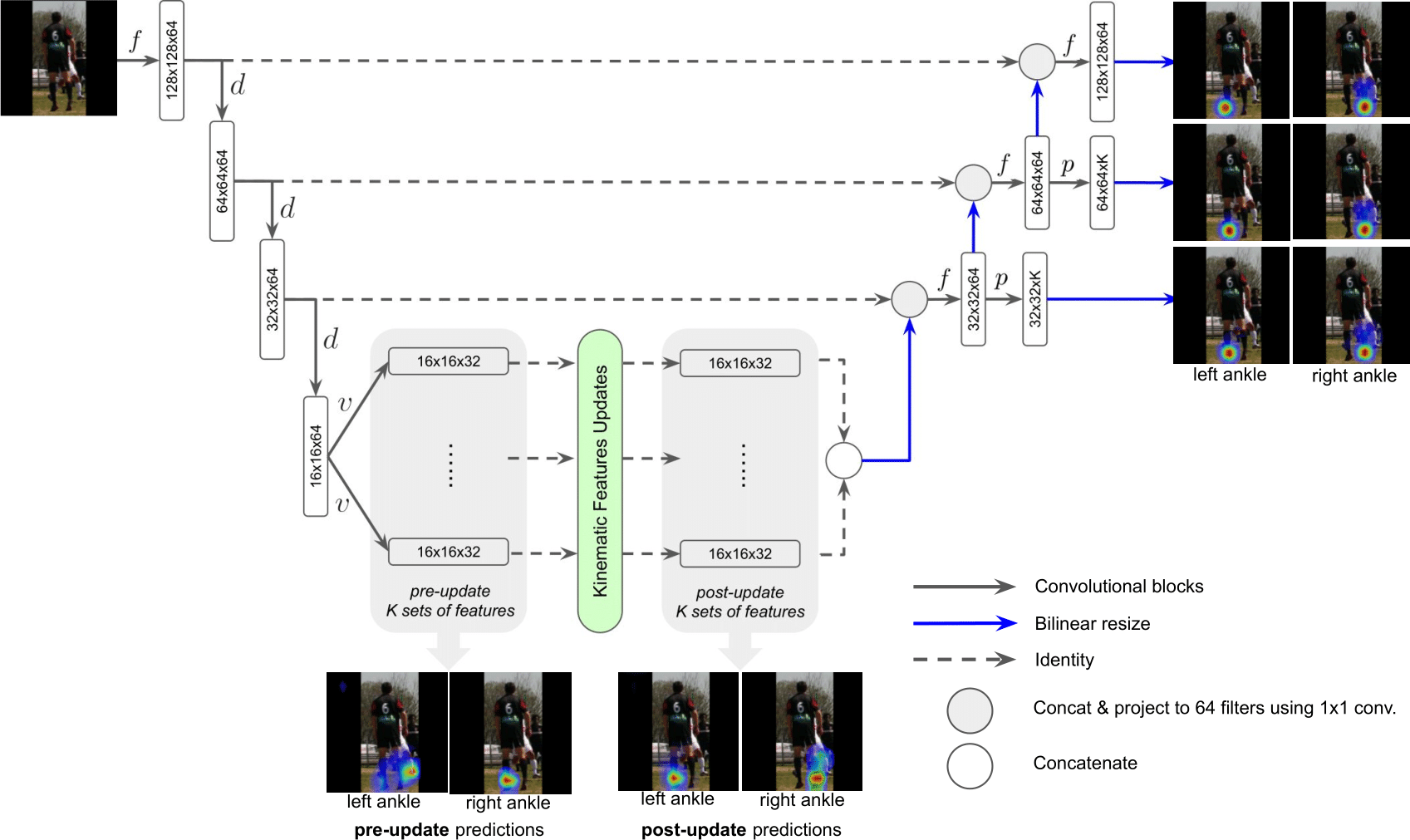}
\caption{The network architecture of \reposenospace. We use $f$, $d$, $v$ and $p$ to denote $(3, 1, 64)^4$, $(5, 2, 64)$, $(1, 1, 32)$ and $(3, 1, K)_u$ \textit{\textbf{convolutional blocks}} respectively; see text for definition. All convolution blocks are unique, \ie~no weight sharing, and we dropped their identifying indices for simplicity. Also, all predicted heatmaps are resized to $128 \times 128$.
As shown, after applying the kinematic features updates, our approach was able to correctly recover the ankles; see full predicted pose in Figure~\ref{fig:teaser_image}.}
\label{fig:network}
\end{figure*}

In reality those approaches sacrificed practicality, in terms of network size, for better benchmark performance metrics. There are a number of recent attempts to find lightweight pose estimation networks, while achieving close to state-of-the-art performance~\cite{bulat2017binarized,fastpose19}. In~\cite{bulat2017binarized}, the authors explored weight binarization~\cite{rastegari2016xnor,courbariaux2016binarized}, which enabled them to replace multiplications with bitwise XOR operations. Their approach, however, resulted in a significant drop in performance. Recently, \cite{fastpose19} was successful in distilling the stacked-hourglass~\cite{stackedhourglass16} network with minimal drop in performance. 



In Section \ref{sc:method} we describe our approach, \reposenospace,  for encoding geometric and structural priors via convolutional feature updates. Then we compare our approach to various state-of-the-art methods in Section \ref{sc:exp} and run an  extensive ablation studies of our model components. Finally, Section \ref{sc:conclusion} concludes our findings and contributions.
\section{Method}
\label{sc:method}

Let $I$ denote an $n \times n$ image. In our work, a human pose is represented by $K$ 2D keypoints, \eg~head, left ankle, \etc, where $\keypoint^i_k=(x^i_k, y^i_k) \in \mathbb{R}^2_{\geq 0}$  is the $k^{th}$ keypoint of example~$i$ in the dataset.
Our approach predicts a set of $K$ heatmaps, one for each keypoint. The ground truth heatmap $H_k^i \in \mathbb{R}\strut^{n\times n}_{\geq 0}$ of keypoint $\keypoint_k^i$ is an unnormalized Gaussian centered at $\keypoint_k^i$ with standard deviation~$\sigma$\footnote{In our experiments, we set $\sigma$ to 5 for a $128 \times 128$ input image size.}.

\vspace{1.5ex}
\subsection{Network}
\vspace{0.5ex}
\label{sc:network}
To simplify our network description, we define a \mbox{\textit{\textbf{convolutional block}}} as a 2D convolution followed by ReLU activation and batch normalization layer.
A series of $r$ convolutional blocks are denoted by $(\kernel, \stride, \filters)^r$, where $\kernel$, $\stride$ and $\filters$ are kernel size, stride and the number of output filters, respectively. 
In addition, $(\kernel, \stride, \filters)_u$ denotes a convolutional block without batch normalization layer.

Figure~\ref{fig:network} shows our network architecture. At the coarsest resolution the features are decoupled into $K$ independent sets of features.
To encourage that each set of features corresponds to a unique keypoint, we predict a single heatmap\footnote{We used $(3, 1, 32)^4$ and $(3, 1, 1)_u$ convolutional blocks per heatmap.} from each set of features out of the $K$ sets. 
Afterwards we concatenate all predicted $K$ heatmaps to form the pre-update heatmaps.

Next, we update the decoupled sets of features according to a predefined \textit{ordering} and kinematic \textit{connectivity}, which is covered in Section~\ref{sc:kin_updates}.
Then we use the updated features to compute post-update heatmaps, in the same manner as the pre-update heatmaps were computed. 
At this point we concatenate all the features used to predict the post-update heatmaps, this step is shown as a white circle in Figure~\ref{fig:network}.

The concatenated features are then bilinearly upsampled and concatenated with the skip connection, and projected back to $64$ channels. At each resolution $K$ heatmaps are predicted which are then bilinearly upsampled to full resolution. The refinement procedure continues as depicted in Figure~\ref{fig:network} until full resolution is achieved. 
Finally, loss \eqref{eq:loss} is applied to all predicted heatmaps.

Without feature decoupling and kinematic updates, which are discussed in Section~\ref{sc:kin_updates}, RePose reduces a UNet/Hourglass style architecture with intermediate supervision.

\subsection{Kinematic Features Updates}
\label{sc:kin_updates}
\def\neighbset{\mathcal{N}}
\def\edges{\mathcal{E}}
As shown in Figure~\ref{fig:network}, the kinematic features updates part of our network receives the decoupled $K$ sets of features $\{\feature_k \}_1^K$.
The basic idea at this stage is to update these sets of features in a way that enables the network to learn kinematic priors between keypoints and how to correlate them.
As such we update the decoupled keypoints' features according to a predefined ordering and kinematic connectivity. 

\begin{figure}[t]
\centering
\includegraphics[height=0.18\textheight]{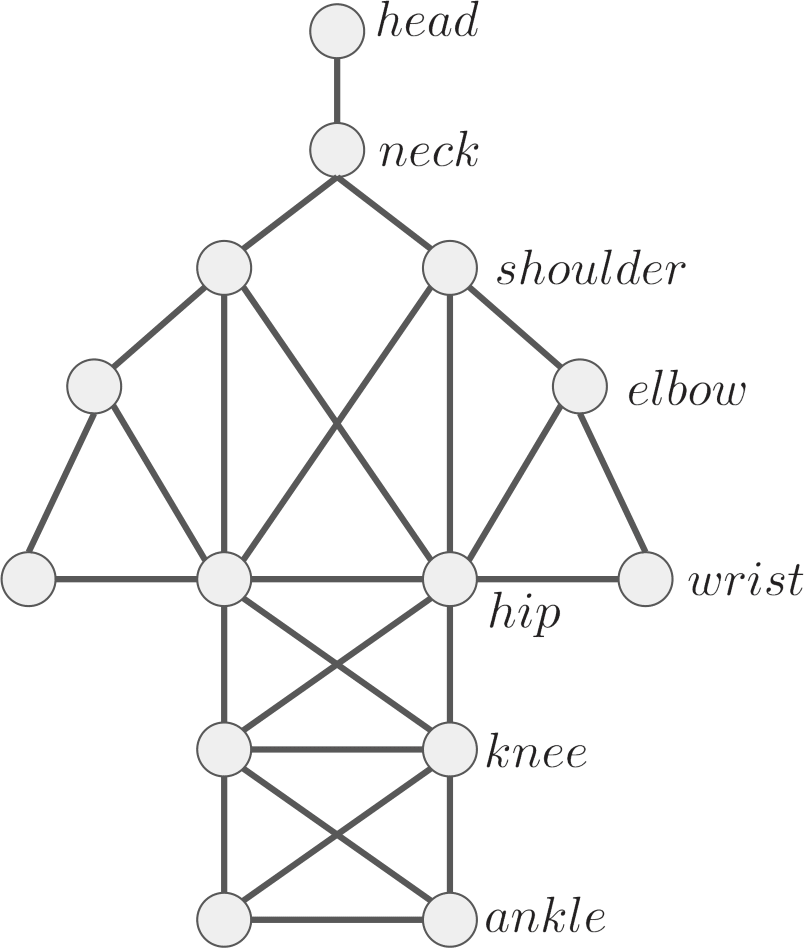}
\caption{The keypoint connectivity. At the coarsest resolution, features of each keypoint are updated based on the features of its neighbors. Our update ordering is hips, shoulders, knees, elbows, ankles, wrists, neck and then head, where the right keypoint comes just before its left counterpart.}
\label{fig:connectivity}
\end{figure}

Our predefined \textit{ordering} starts with keypoints that are more likely to be predicted with high fidelity, \eg~hips or head, and ends with usually poorly predicted ones, \eg~wrists or ankles, see Figure~\ref{fig:connectivity} for the predefined ordering used in our approach.

The \textit{connectivity} defines which keypoints we expect the network to learn to correlate. 
In our method connectivity is not restricted to trees.
We used an undirected graph to define such connectivity, where each keypoint is represented by a unique node, and the set of edges $\edges$ encodes the desired connectivity; see Figure~\ref{fig:connectivity}.
For a keypoint/node $k$ let 
\begin{equation*}
\neighbset(k) = (u \; |\; \forall \; (k, u) \in \edges)
\end{equation*}
be the ordered set of its neighbouring keypoints \wrt~$\edges$.

We update the keypoints one at a time following the predefined ordering.
The features $\feature_k$ of keypoint $k$ are updated as follows:
\begin{align}
\concat_k \gets & \; [\feature_k, \feature_j \; | \; \forall j \in \neighbset(k)] \label{eq:concat}\\
\feature_k \gets  & \; \feature_k + \lambda_k \; g_k(h_k(\concat_k))  \label{eq:update},
\end{align}
where $\lambda_k$ is a trainable parameter, $h_k$ and $g_k$ are $(1,1,32)$ and $(3,1,32)^4$ convolutional blocks, respectively.
In \eqref{eq:concat} we simply concatenate $\feature_k$ and all the features of its neighbouring keypoints.
Then $h_k$ projects the concatenated features to $32$ channels, which then pass through four convolutional blocks.
The features are updated via a residual connection \eqref{eq:update}  with a trainable weight. 
Finally, inspired by message passing techniques, we update the features one more time \wrt~the reversed ordering. It should be noted that the two passes do not share any trainable parameters.

\subsection{Loss}
Our loss is partial Mean Squared Error (MSE)

\begin{equation} 
L = \frac{1}{M}\sum_{m=1}^{M}\frac{1}{|\mathbf{a}_i|}\sum_{k=1}^{K} \mathbf{1}_{\mathbf{a}_i}(k)\; \frac{\| H_k^{i} - O_k^{i}\|_2^{2}}{n^2} \label{eq:loss},
\end{equation}
where $M$ is the batch size and $O_k^{i}$ is the heatmap predicted by the network. Some of the images in the datasets are not fully annotated, as such we define $\mathbf{a}_i$ to be the set of annotated keypoints of example $i$.
It should be noted that MSE is a fairly standard loss for pose estimation but its partial counterpart was not used before to the best of our knowledge. As shown in Figure \ref{fig:network}, \repose produces multiple heatmaps/predictions for intermediate supervision. Our total loss is the sum of \eqref{eq:loss} for all the predicted heatmaps.
\section{Experiments}
\label{sc:exp}
\paragraph{Datasets} We evaluated our \repose network on two standard pose estimation datasets, namely Leeds Sports Pose (LSP)~\cite{johnson2010clustered,johnson2011learning} and MPII Human Pose~\cite{andriluka14cvpr}.
MPII is more challenging compared to LSP, as poses in MPII cover a large number of activities. Furthermore, MPII has a large number of spatially inseparable poses, which frequently occur in crowded scenes. 
MPII provides an estimate of pose center and scale, while LSP does not.
To allow for joint training on both datasets we used an estimated pose center and scale for the LSP training set, as done in \cite{featurepyramids17,stackedhourglass16,convposemachinesCVPR16}.
For LSP testing set, the scale and center were set to the image's size and center, respectively.

\begin{table*}[t]
\vspace{-4ex}
\begin{tabular}{|*{11}{c@{\hspace{1.1ex}}|}}
\hline 
{\small } & {\small \textbf{Head}} & {\small \textbf{Shoulder}} & {\small \textbf{Elbow}} & {\small \textbf{Wrist}} & {\small \textbf{Hip}} & {\small \textbf{Knee}} & {\small \textbf{Ankle}} & {\small \textbf{Mean}} & {\small \textbf{\# Param}} & {\small \textbf{FLOPS}} \\ 
 \hline 
{\footnotesize Tompson \etal~NIPS 14 \cite{tompson2014joint}}$\dagger$ &  $90.60$ & $79.20$ & $67.90$ & $63.40$ & $69.50$ & $71.00$ & $64.20$ & $72.30$ &- & - \\
\hline 
{\small Rafi \etal~BMVC 16 \cite{rafi2016efficient}} &  $95.80$ & $86.20$ & $79.30$ & $75.00$ & $86.60$ & $83.80$ & $79.80$ & $83.80$ &56M & 28G \\
\hline 
{\small Yang \etal~CVPR 16 \cite{deformableparts16}}$\dagger$ &  $90.60$ & $78.10$ & $73.80$ & $68.80$ & $74.80$ & $69.90$ & $58.90$ & $73.60$ &- & - \\
\hline 
{\small Yu \etal~ECCV 16 \cite{yu2016deep}} &  $87.20$ & $88.20$ & $82.40$ & $76.30$ & $91.40$ & $85.80$ & $78.70$ & $84.30$ &- & - \\
\hline 
{\small Carreira \etal~CVPR 16 \cite{iterfeedback16}} &  $90.50$ & $81.80$ & $65.80$ & $59.80$ & $81.60$ & $70.60$ & $62.00$ & $73.10$ &- & - \\
\hline 
{\small Yang \etal~ICCV 17 \cite{featurepyramids17}} &  $98.30$ & $94.50$ & $92.20$ & $88.90$ & $94.40$ & $95.00$ & $93.70$ & $93.90$ &28M & 46G \\
\hline 
{\small Peng \etal~CVPR 18 \cite{avddataugment18}} &  $\mathbf{98.6}$ & $\mathbf{95.3}$ & $\mathbf{92.8}$ & $\mathbf{90.0}$ & $\mathbf{94.8}$ & $\mathbf{95.3}$ & $\mathbf{94.5}$ & $\mathbf{94.50}$ &26M & 55G \\
\hline 
\multicolumn{11}{c}{{\small \em lightweight pose estimation approaches}}\\
\hline 
{\small Fast Pose CVPR 19 \cite{fastpose19}} &  $97.30$ & $92.30$ & $86.80$ & $84.20$ & $\mathbf{91.9}$ & $92.20$ & $90.90$ & $90.80$ &\textbf{3M} & \textbf{9G} \\
\hline 
{\small \repose} &  $\mathbf{97.75}$ & $\mathbf{92.9}$ & $\mathbf{88.5}$ & $\mathbf{86.3}$ & $91.19$ & $\mathbf{93.0}$ & $\mathbf{92.0}$ & $\mathbf{91.66}$ & 4M & 13.5G \\
\hline 
\end{tabular}
\vspace{1ex}
\caption{A comparison of various methods on the LSP dataset using the PCK@0.2 metric~\cite{yang2012articulated}. \repose achieved better results \wrt~other state-of-the-art CNN methods that are based on statistical graphical methods$^\dagger$ (trained on LSP). M and G stand for Million and Giga.}
\label{tbl:leeds_vs_sota}
\end{table*}

\begin{table*}[ht]
\begin{tabular}{|*{11}{c@{\hspace{1.1ex}}|}}
\hline 
{\small } & {\small \textbf{Head}} & {\small \textbf{Shoulder}} & {\small \textbf{Elbow}} & {\small \textbf{Wrist}} & {\small \textbf{Hip}} & {\small \textbf{Knee}} & {\small \textbf{Ankle}} & {\small \textbf{Mean}} & {\small \textbf{\# Param}} & {\small \textbf{FLOPS}} \\ 
 \hline 
{\footnotesize Insafutdinov \etal~ECCV 16 \cite{deepercutECCV16}} &  $96.80$ & $95.20$ & $89.30$ & $84.40$ & $88.40$ & $83.40$ & $78.00$ & $88.50$ &66M & 286G \\
\hline 
{\small Rafi \etal~BMVC 16 \cite{rafi2016efficient}} &  $97.20$ & $93.90$ & $86.40$ & $81.30$ & $86.80$ & $80.60$ & $73.40$ & $86.30$ &56M & 28G \\
\hline 
{\small Wei \etal~ CVPR 16 \cite{convposemachinesCVPR16}} &  $97.80$ & $95.00$ & $88.70$ & $84.00$ & $88.40$ & $82.80$ & $79.40$ & $88.50$ &31M & 351G \\
\hline 
{\small Newell \etal~ECCV 16 \cite{stackedhourglass16}} &  $98.20$ & $96.30$ & $91.20$ & $87.10$ & $90.10$ & $87.40$ & $83.60$ & $90.90$ &26M & 55G \\
\hline 
{\small Chu \etal, CVPR 17 \cite{attention17}} &  $98.50$ & $96.30$ & $91.90$ & $88.10$ & $90.60$ & $88.00$ & $85.00$ & $91.50$ &58M & 128G \\
\hline 
{\small Yang \etal~ICCV 17 \cite{featurepyramids17}} &  $98.50$ & $96.70$ & $92.50$ & $88.70$ & $91.10$ & $88.60$ & $86.00$ & $92.00$ &28M & 46G \\
\hline 
{\small Nie \etal~CVPR 18 \cite{nie2018human}} &  $\mathbf{98.6}$ & $\mathbf{96.9}$ & $\mathbf{93.0}$ & $\mathbf{89.1}$ & $\mathbf{91.7}$ & $\mathbf{89.0}$ & $\mathbf{86.2}$ & $\mathbf{92.4}$ &26M & 63G \\
\hline 
{\small Peng \etal~CVPR’18 \cite{avddataugment18}} &  $98.10$ & $96.60$ & $92.50$ & $88.40$ & $90.70$ & $87.70$ & $83.50$ & $91.50$ &26M & 55G \\
\hline 
\multicolumn{11}{c}{{\small \em lightweight pose estimation approaches}}\\
\hline 
{\small Sekii~ECCV18 \cite{sekii2018pose}} &  - & - & - & - & - & - & - & $88.10$ &16M & \textbf{6G} \\
\hline 
{\small Fast Pose CVPR 19 \cite{fastpose19}} &  $\mathbf{98.3}$ & $\mathbf{96.4}$ & $\mathbf{91.5}$ & $\mathbf{87.4}$ & $90.90$ & $\mathbf{87.1}$ & $\mathbf{83.7}$ & $\mathbf{91.1}$ &\textbf{3M} & 9G \\
\hline 
{\small \repose} &  $97.65$ & $96.39$ & $91.00$ & $85.20$ & $\mathbf{92.95}$ & $86.40$ & $82.50$ & $90.29$ &4M & 13.5G \\
\hline 
\end{tabular}
\vspace{1ex}
\caption{A comparison of various methods on the MPII dataset using the PCKh@0.5 metric~\cite{andriluka14cvpr}. \repose achieved comparable results to Stacked-hourglass \cite{stackedhourglass16} and its distilled version, \ie~Fast Pose~\cite{fastpose19}.  M and G stand for Million and Giga.}
\vspace{-2ex}
\label{tbl:mpii_vs_sota}
\end{table*}

\paragraph{Training} Similar to~\cite{featurepyramids17,stackedhourglass16,convposemachinesCVPR16}, we augmented the training data by cropping according to the provided pose scale and center, and resized the crop to be $128\times128$.
Furthermore, the training data was augmented by scaling $[0.7, 1.3]$, rotation between $\pm 60^{\circ}$, horizontal flipping, and color noise (\ie jitter, brightness and contrast).
Our network described in Section~\ref{sc:network} results in a model with $4$M parameters and $13.5$GFLOPS, which was trained \textit{jointly} on LSP and MPII.
We used Adam optimizer to train our network with a batch size of 64 and a predefined stopping criterion at 2M steps.
The initial learning rate was set to $10^{-3}$ and was dropped to $5 \times 10^{-4}$ and $10^{-6}$ at 1M and 1.3M steps, respectively. Contrary to other approaches we did not fine-tune our model on a specific dataset.

\paragraph{Metrics} For evaluation, we used commonly adopted single pose estimation metrics in the literature.
As per the LSP and MPII benchmarks, we used two variants of the Probability of Correct Keypoints \cite{yang2012articulated,andriluka14cvpr} metric, \ie PCK@0.2 for LSP and PCKh@0.5 for MPII. The former uses $20\%$ of the torso diameter to define the threshold used in identifying correctly predicted keypoints, while the latter uses  $50\%$ of the head length. The validation set of~\cite{convposemachinesCVPR16,featurepyramids17,fastpose19} was used for evaluating our model on MPII.

\paragraph{Quantitative Evaluations} Quantitative results are shown in Tables~\ref{tbl:leeds_vs_sota} and~\ref{tbl:mpii_vs_sota} comparing our trained model to various state-of-the-art approaches on the LSP and MPII datasets, respectively.
As shown in Table~\ref{tbl:leeds_vs_sota}, \repose was able to surpass Yang \etal~\cite{deformableparts16} and Tompson \etal~\cite{tompson2014joint} by a large margin, which try to approximate a MAP-solver of a statistical graphical model within a deep neural network framework. Furthermore, our approach was able to perform better than Fast Pose~\cite{fastpose19} by $0.86\%$ on average. As shown in Table~\ref{tbl:mpii_vs_sota}, \repose reached comparable results to Fast Pose~\cite{fastpose19} and the Stacked-hourglass~\cite{stackedhourglass16}. Our network reaches better performance on MPII at the expense of increasing the number of trainable parameters and FLOPS; see Table~\ref{tbl:repose_stacking}. However, the gain in performance does not seem to justify doubling the network size.
\vspace{-2ex}

\begin{figure*}[ht]
\vspace{-4ex}
\newcommand{\leedsa}{0.12}
\begin{tabular*}{\textwidth}{*{7}{c}}
    \multirow{1}{*}[-8ex]{\rotatebox[origin=c]{90}{\textbf{LSP}}} &
    \includegraphics[height=\leedsa\textheight,frame]{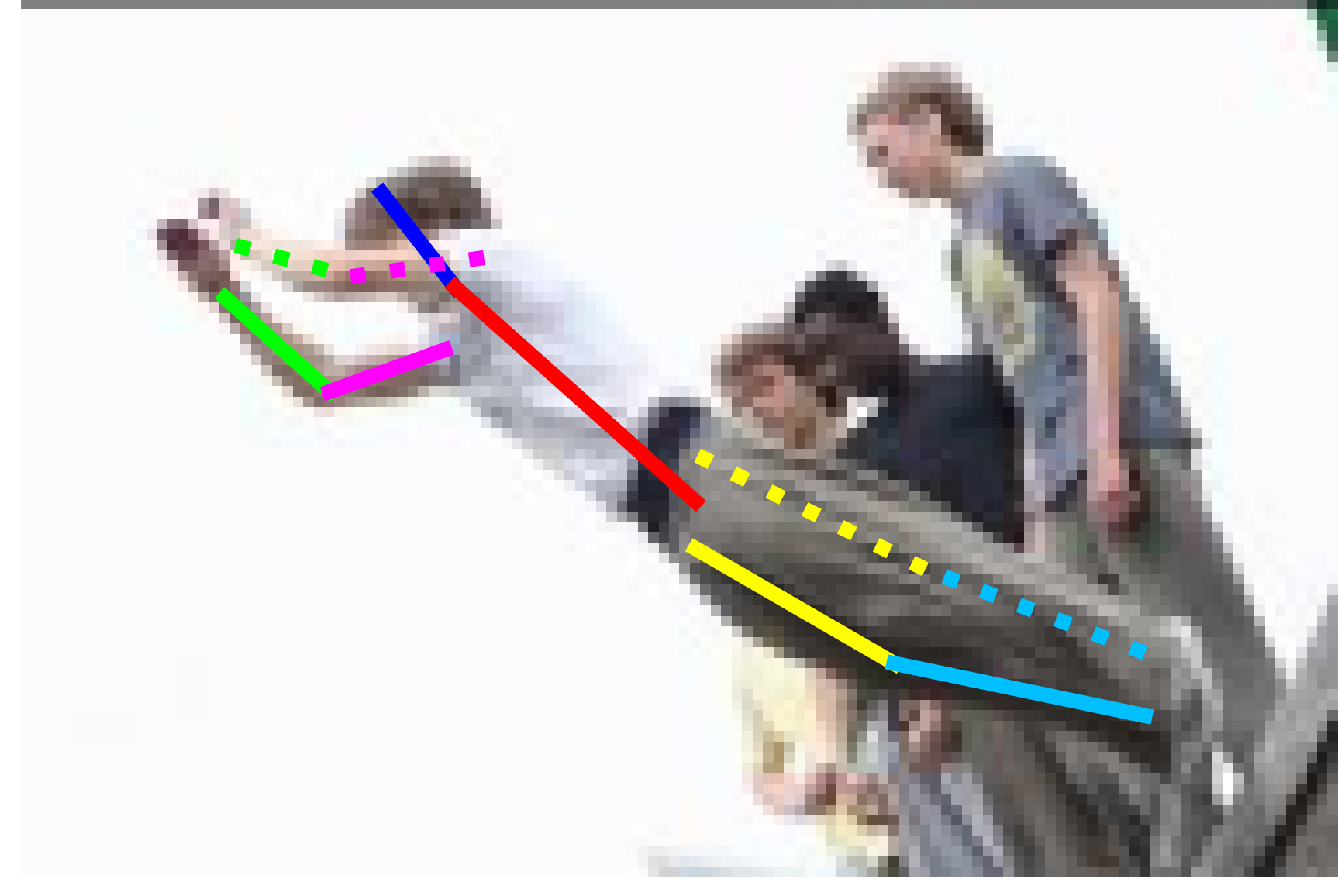} &
    \includegraphics[height=\leedsa\textheight]{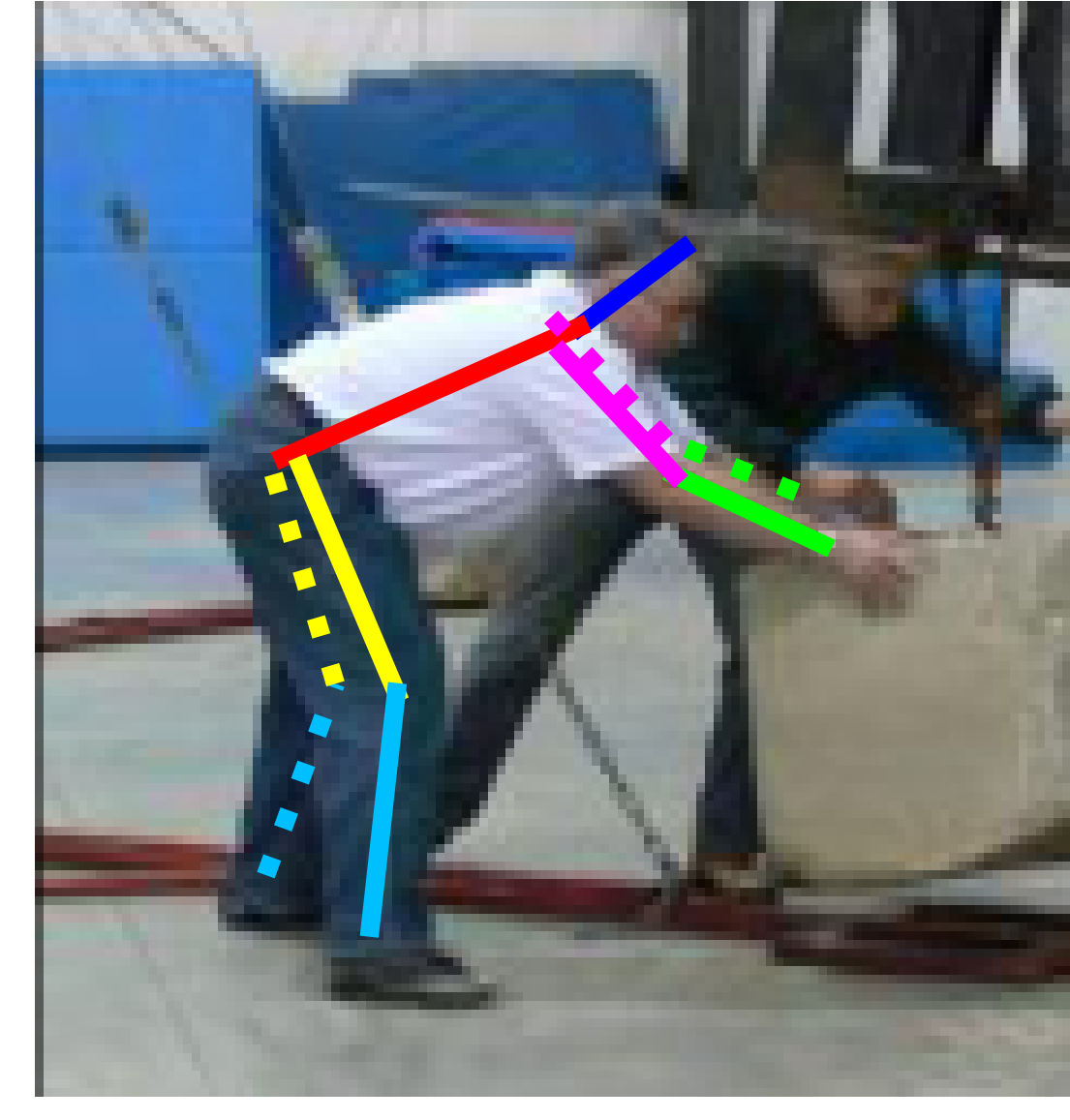} &
    \includegraphics[height=\leedsa\textheight]{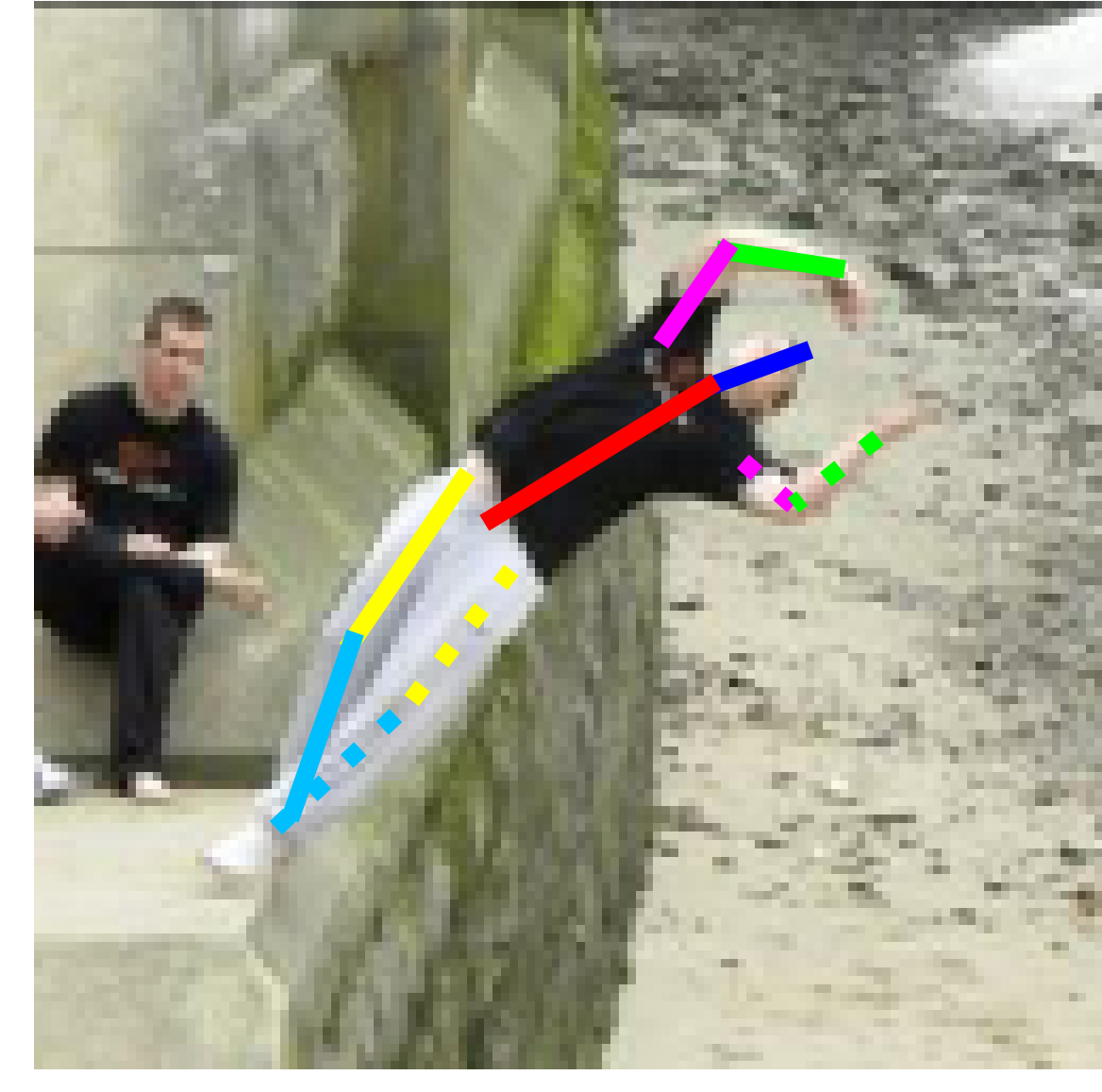} &
    \includegraphics[height=\leedsa\textheight]{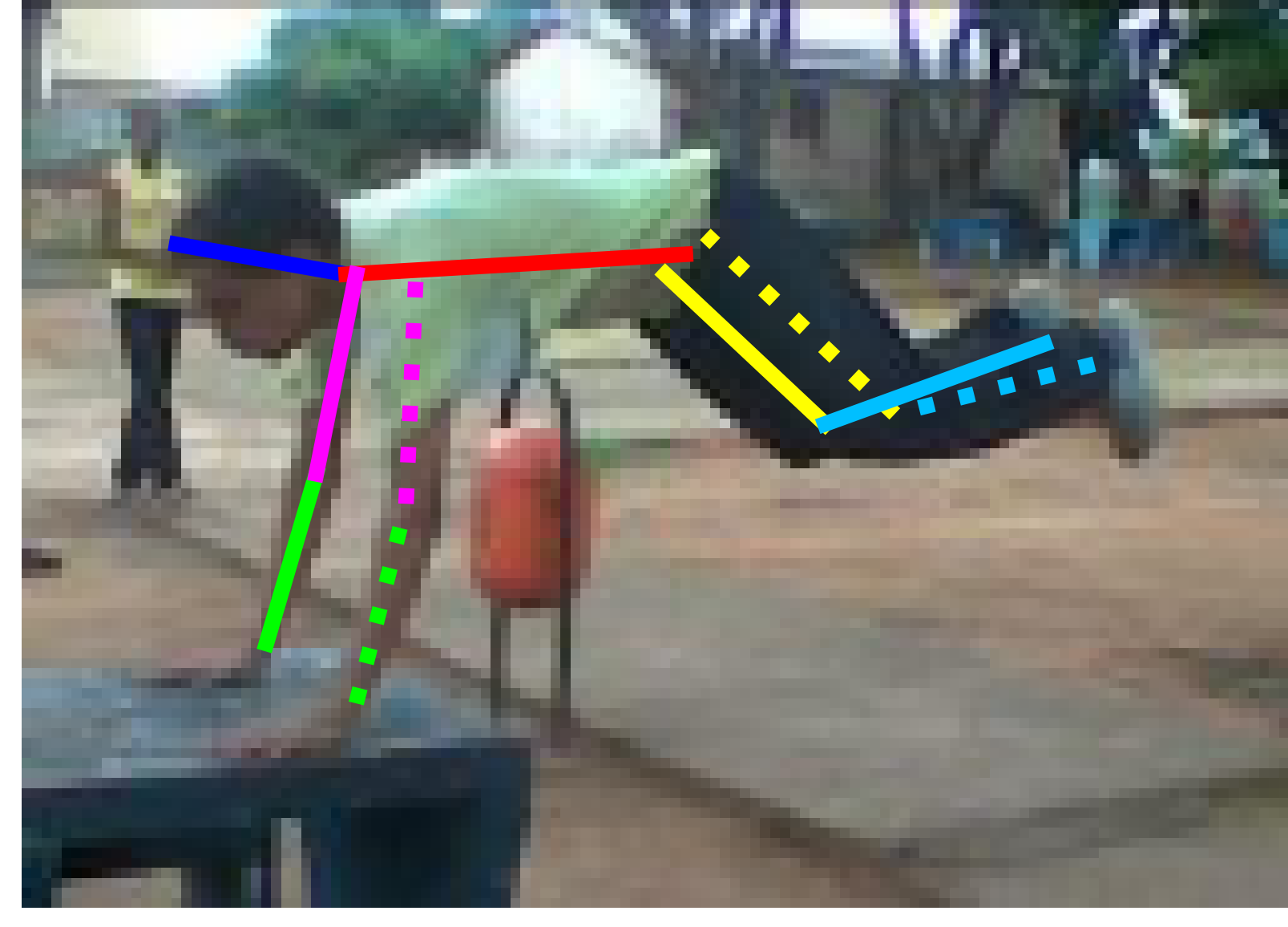} &
    \includegraphics[height=\leedsa\textheight]{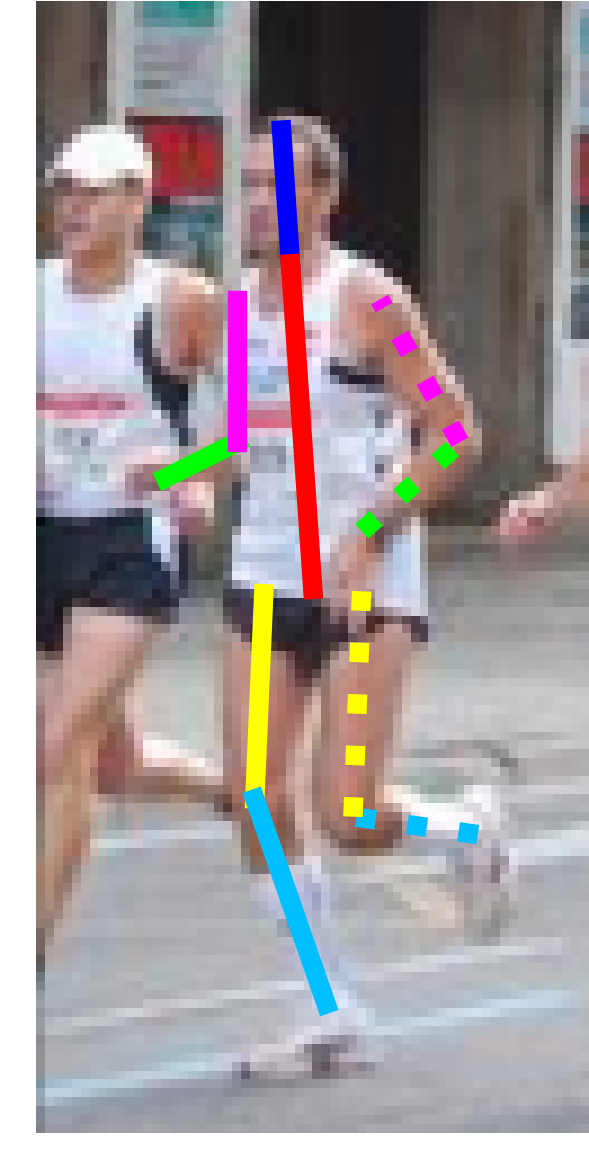} \\
\end{tabular*}\\
\newcommand{\leedsb}{0.117}
\begin{tabular*}{\textwidth}{*{7}{c}}
    \multirow{1}{*}[0ex]{\rotatebox[origin=c]{90}{\phantom{\textbf{LSP}}}} &
    \includegraphics[height=\leedsb\textheight]{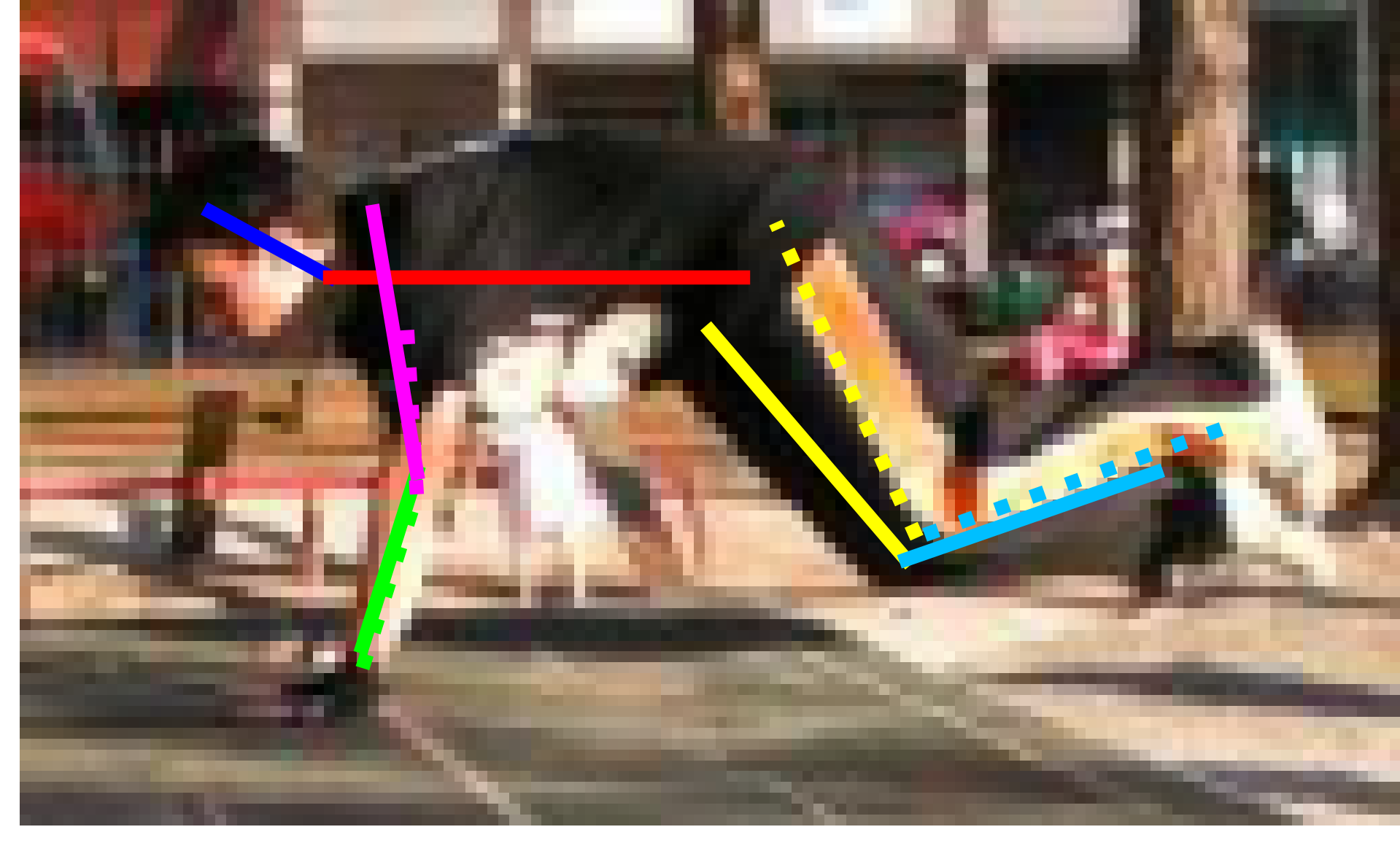} &
    \includegraphics[height=\leedsb\textheight]{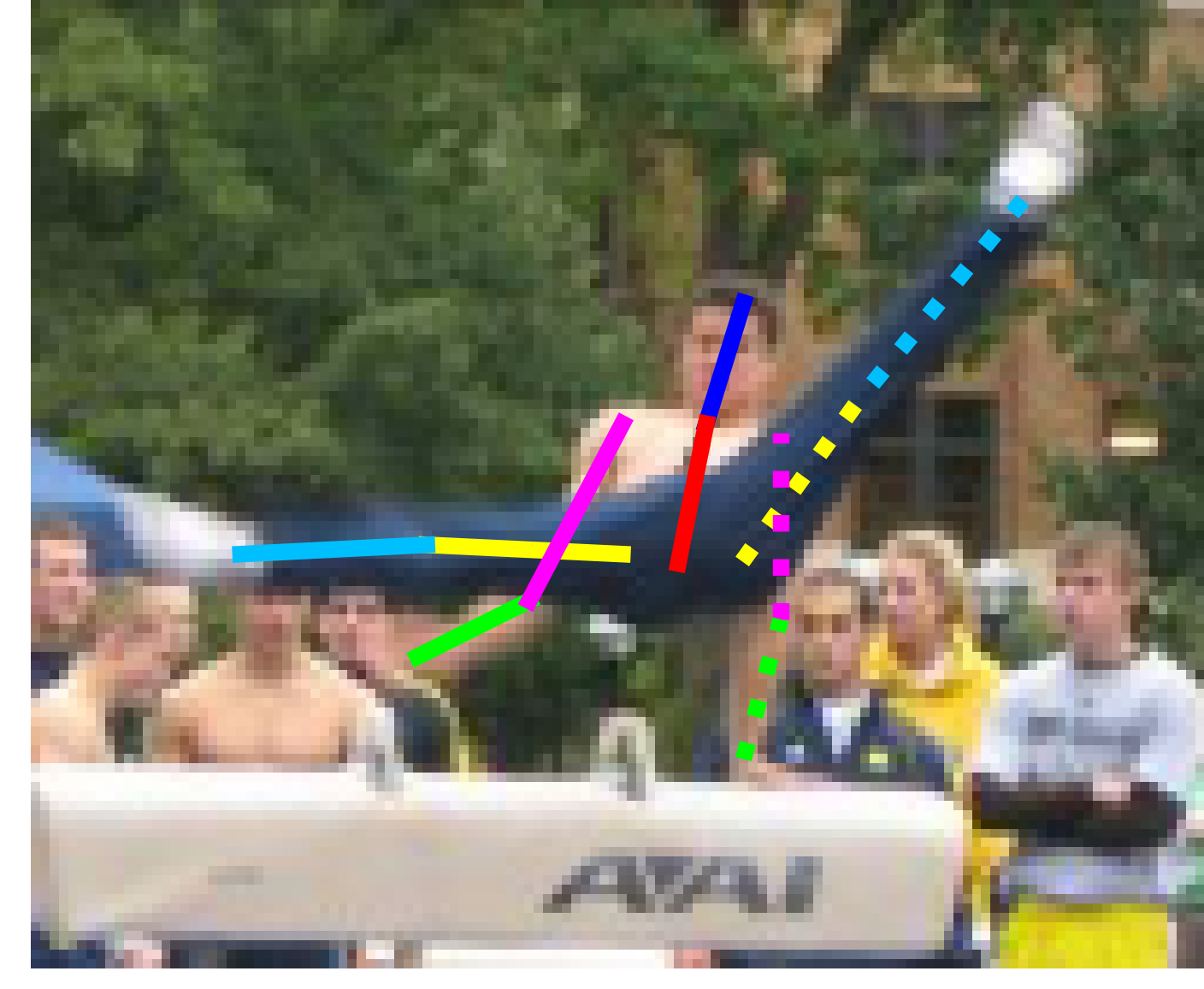} &
    \includegraphics[height=\leedsb\textheight]{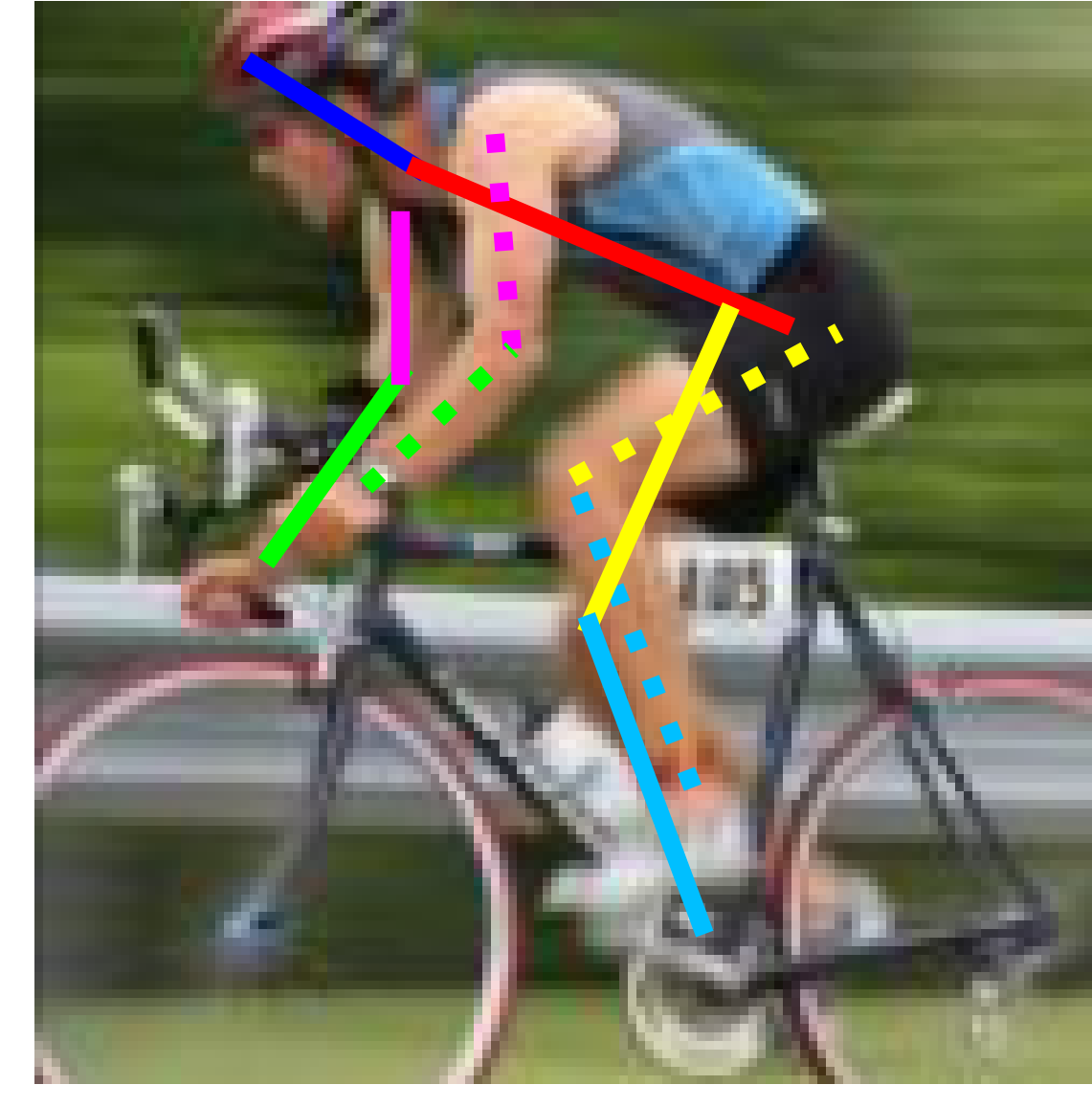} &
    \includegraphics[height=\leedsb\textheight]{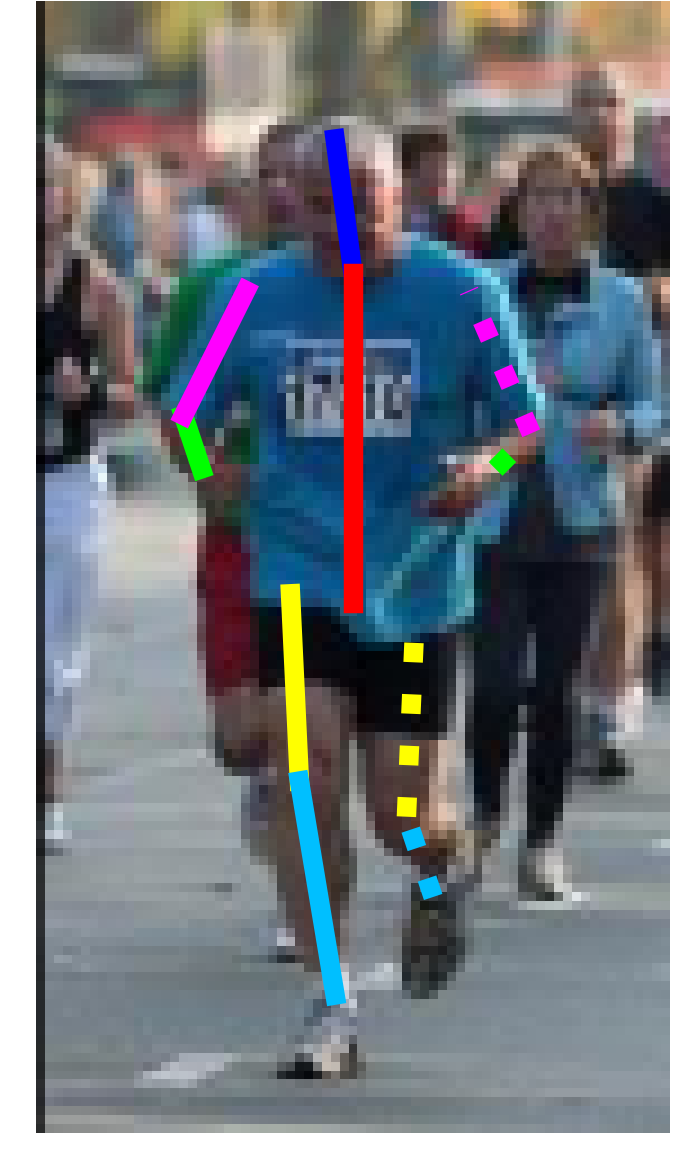} &
    \includegraphics[height=\leedsb\textheight]{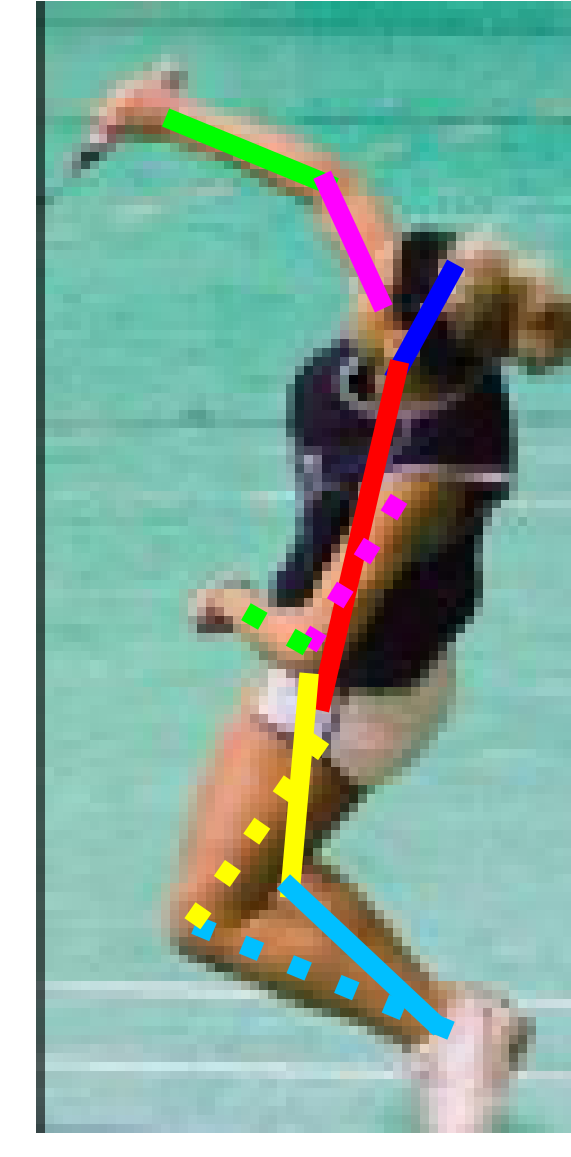} &
    \includegraphics[height=\leedsb\textheight]{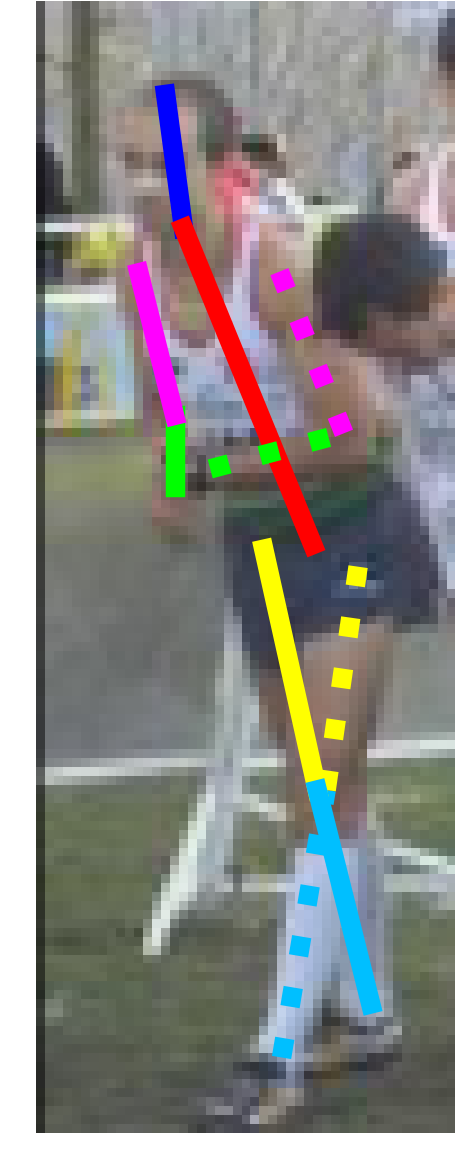} \\
\end{tabular*}

\newcommand{\leedsc}{0.0935}
\begin{tabular*}{\textwidth}{*{8}{c}}
    \multirow{1}{*}[0ex]{\rotatebox[origin=c]{90}{\phantom{\textbf{LSP}}}} &
    \includegraphics[height=\leedsc\textheight]{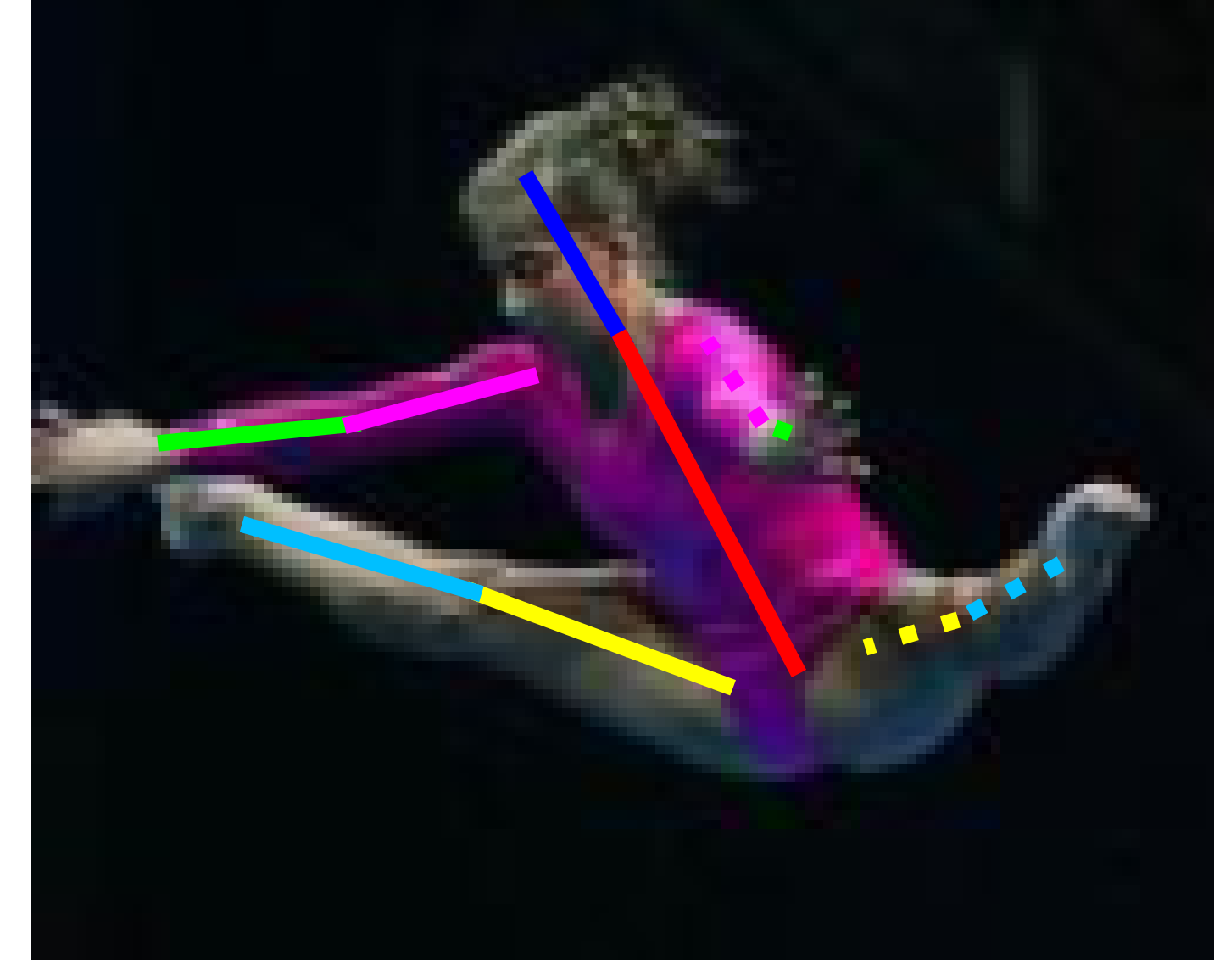} &
    \includegraphics[height=\leedsc\textheight]{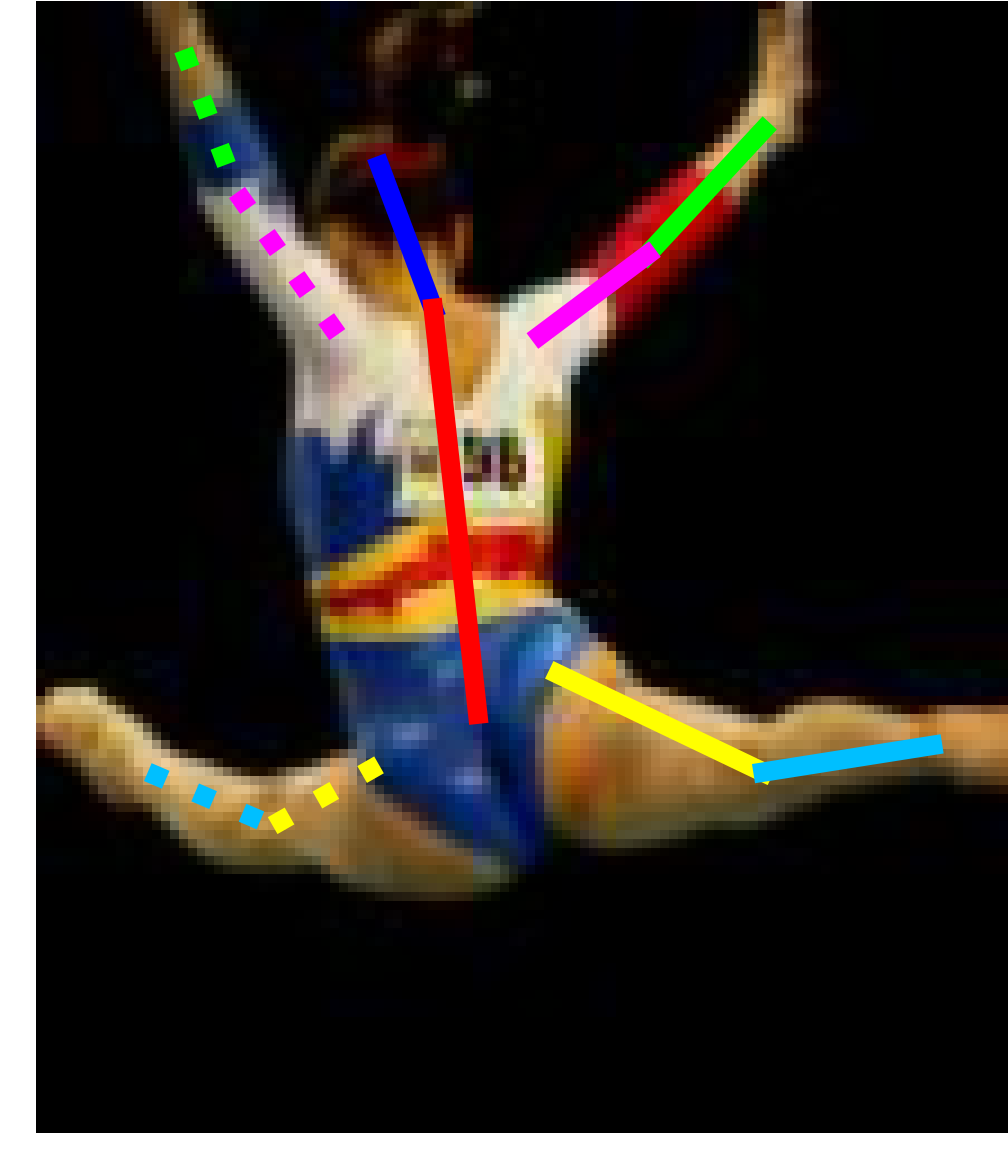} &
    \includegraphics[height=\leedsc\textheight]{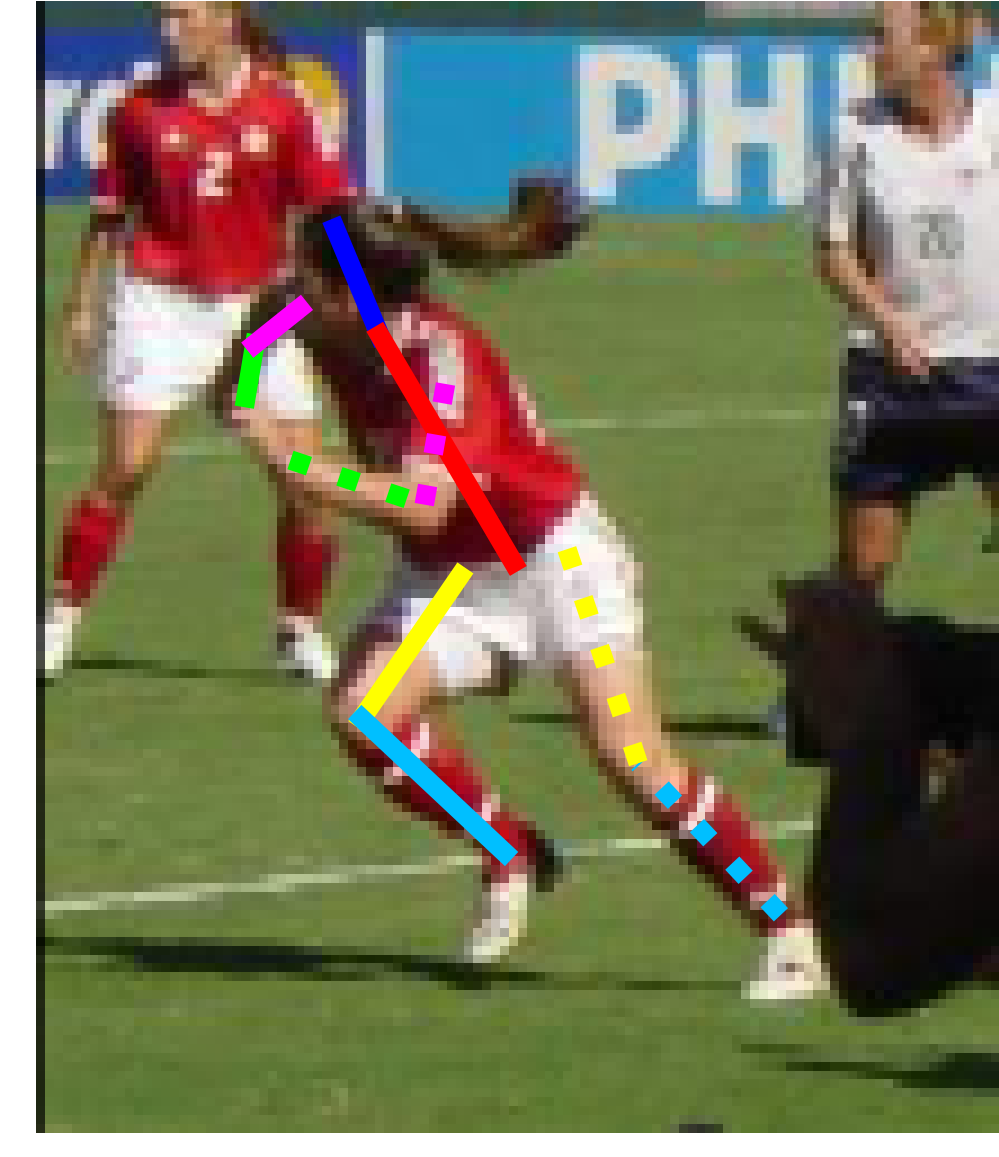} &
    \includegraphics[height=\leedsc\textheight]{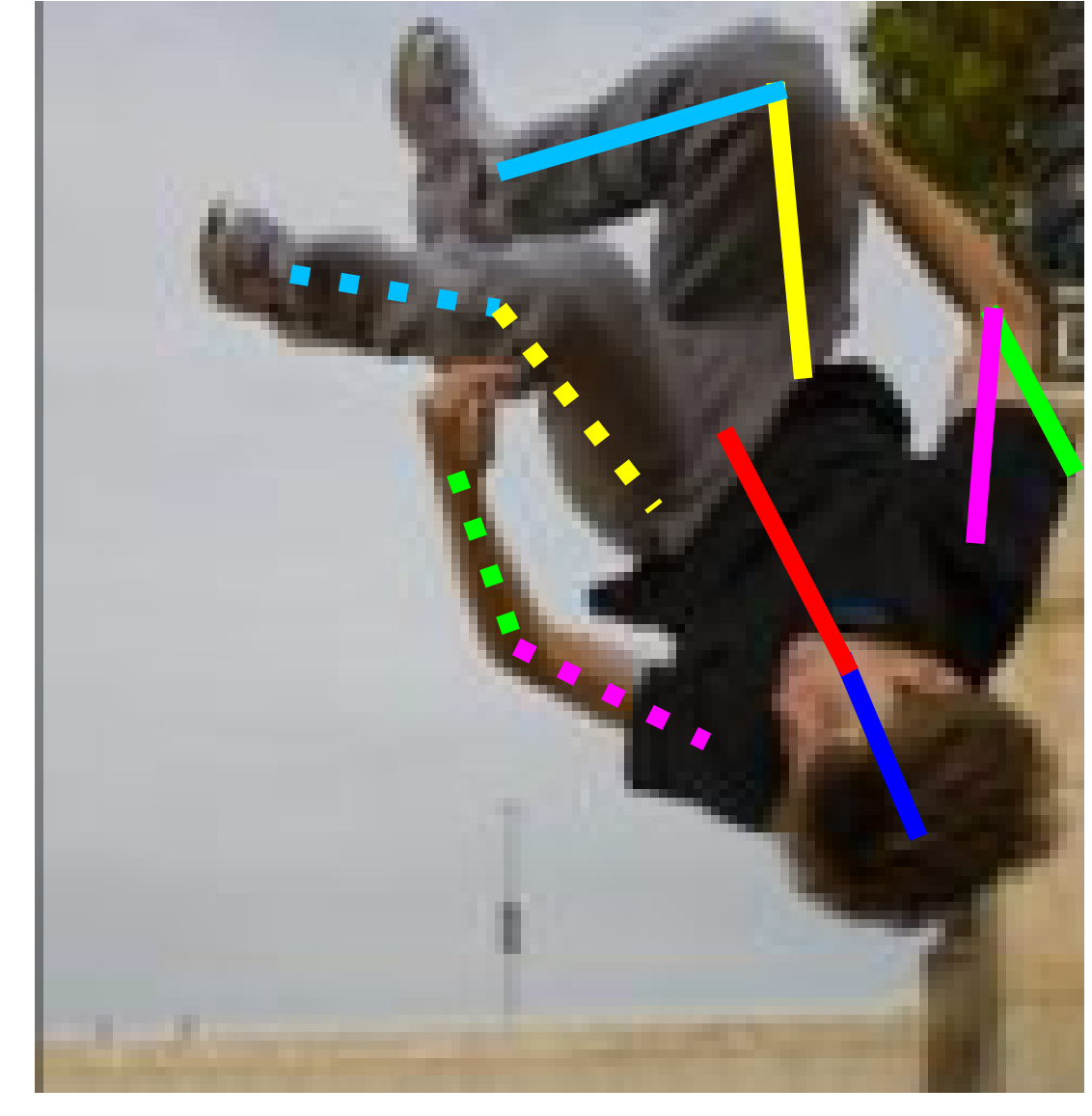} &
    \includegraphics[height=\leedsc\textheight]{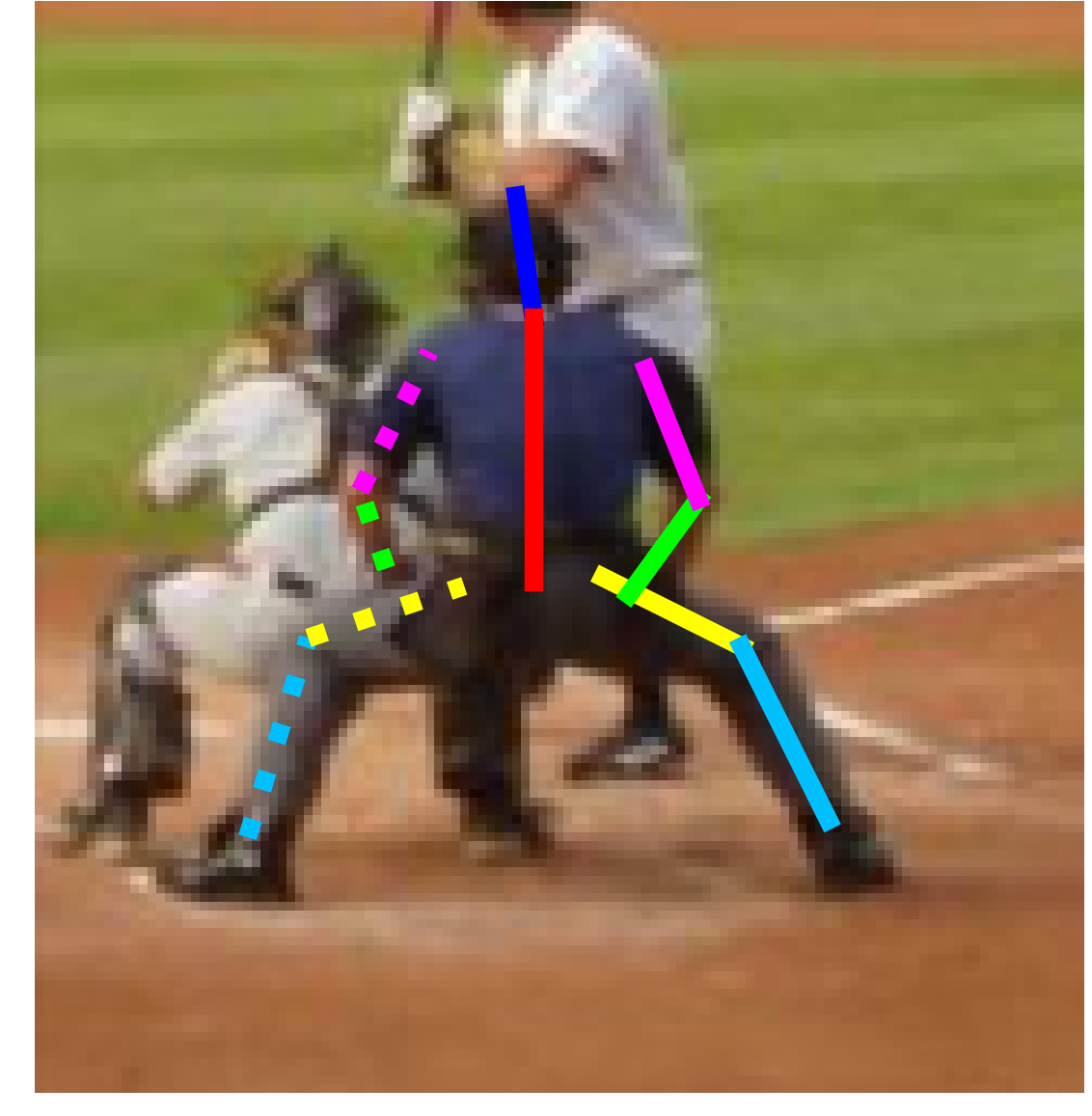} &
    \includegraphics[height=\leedsc\textheight]{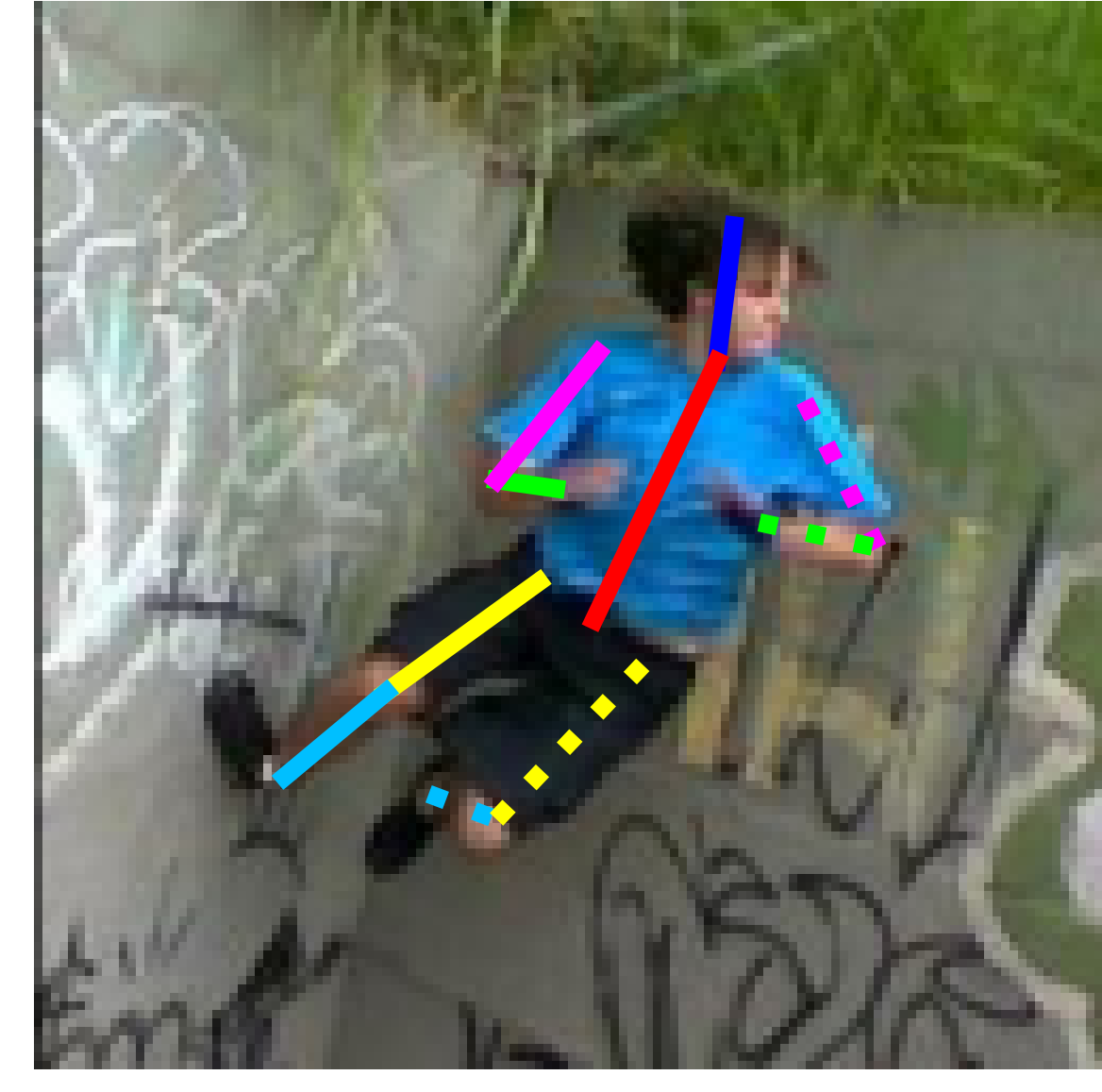} &
    \includegraphics[height=\leedsc\textheight]{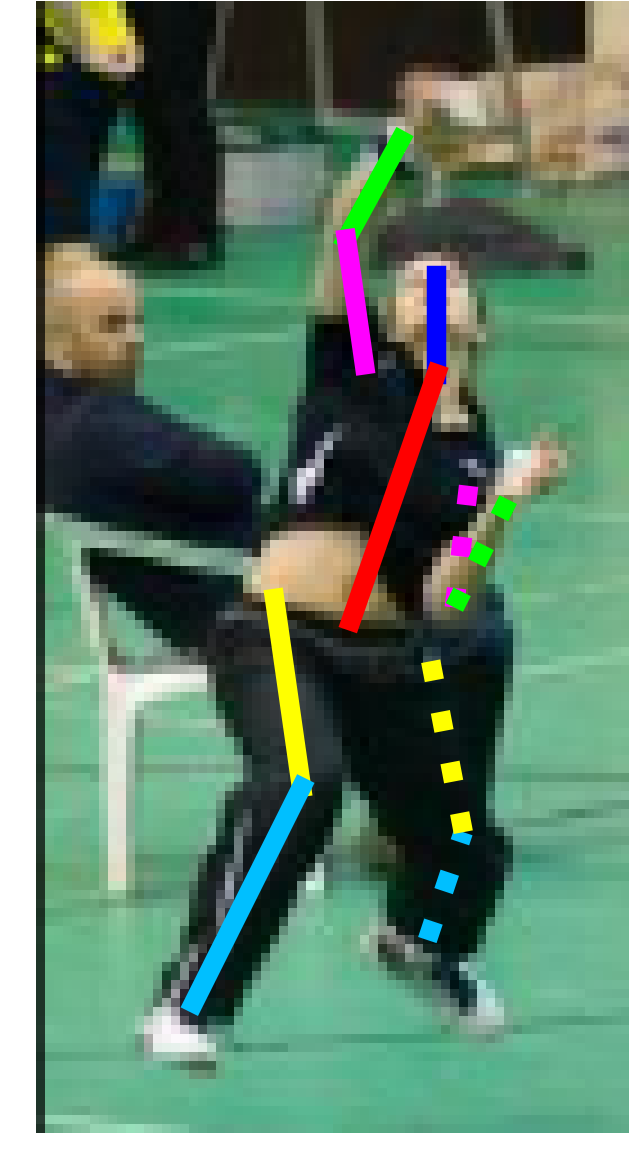} \\
\end{tabular*}
\newcommand{\leedsd}{0.1255}
\begin{tabular*}{\textwidth}{*{9}{c}}
    \multirow{1}{*}[0ex]{\rotatebox[origin=c]{90}{\phantom{\textbf{LSP}}}} &
    \includegraphics[height=\leedsd\textheight]{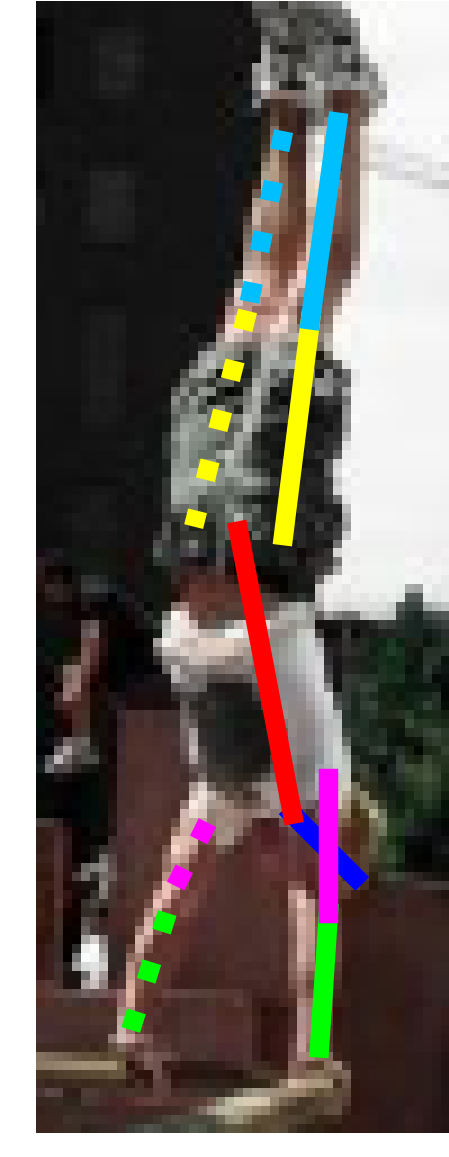} &
    \includegraphics[height=\leedsd\textheight]{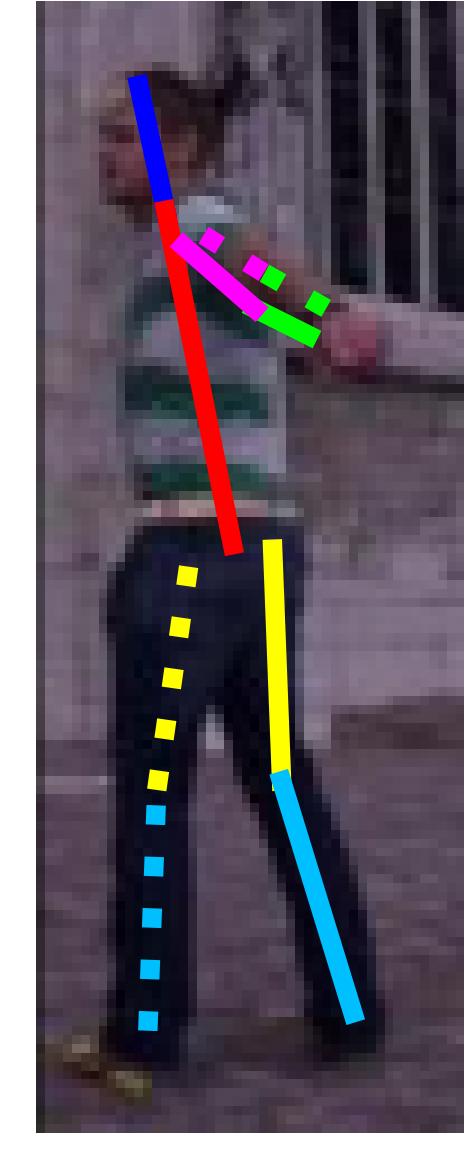} &
    \includegraphics[height=\leedsd\textheight]{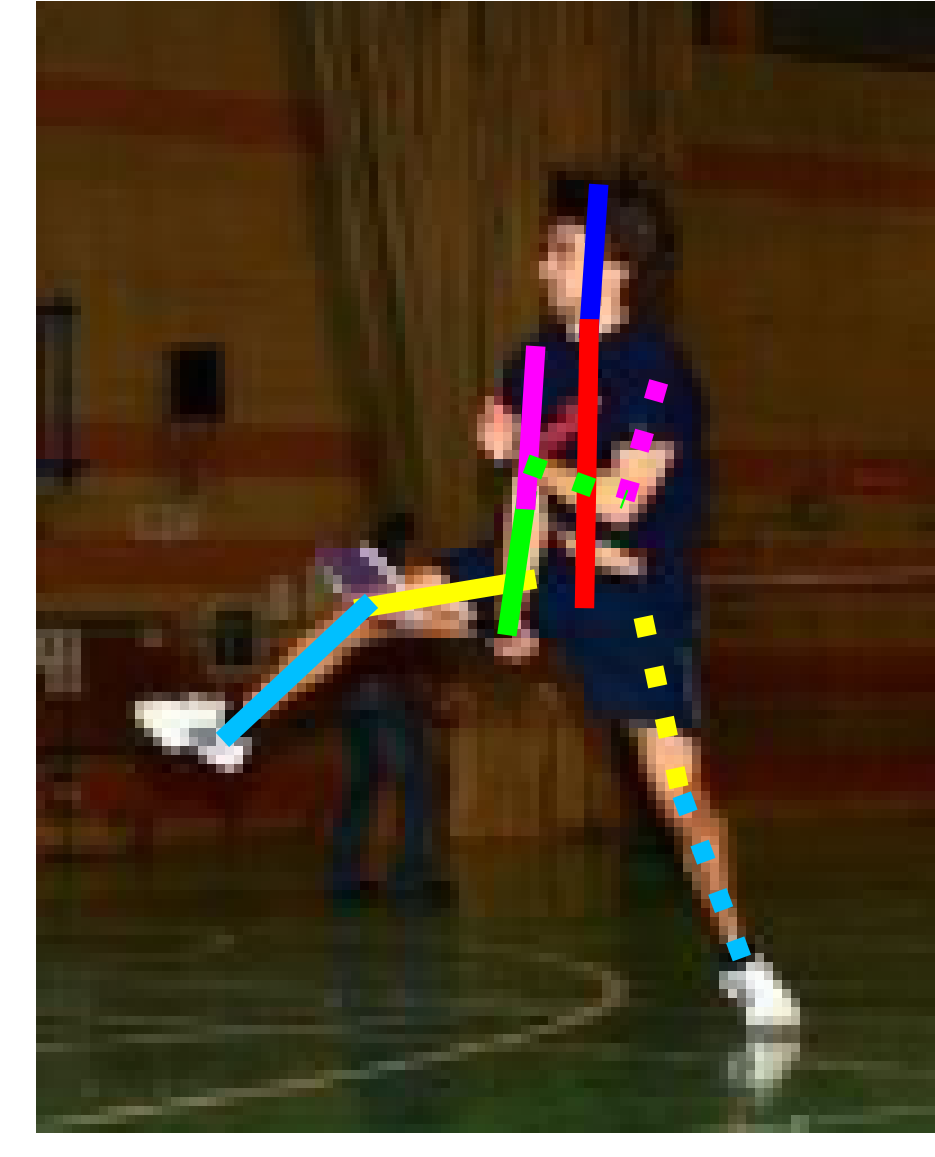} &
    \includegraphics[height=\leedsd\textheight]{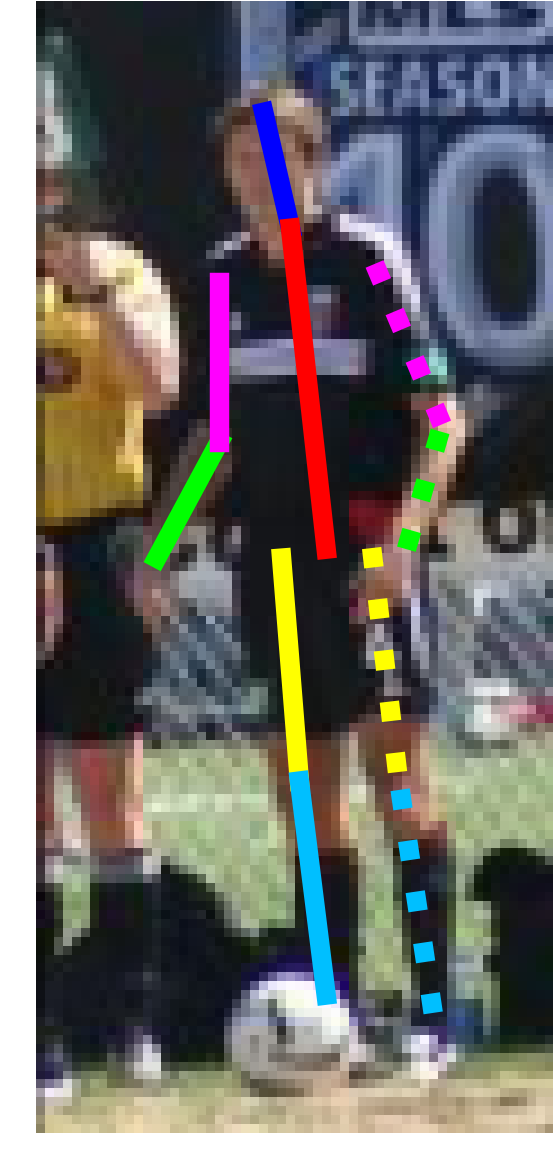} &
    \includegraphics[height=\leedsd\textheight]{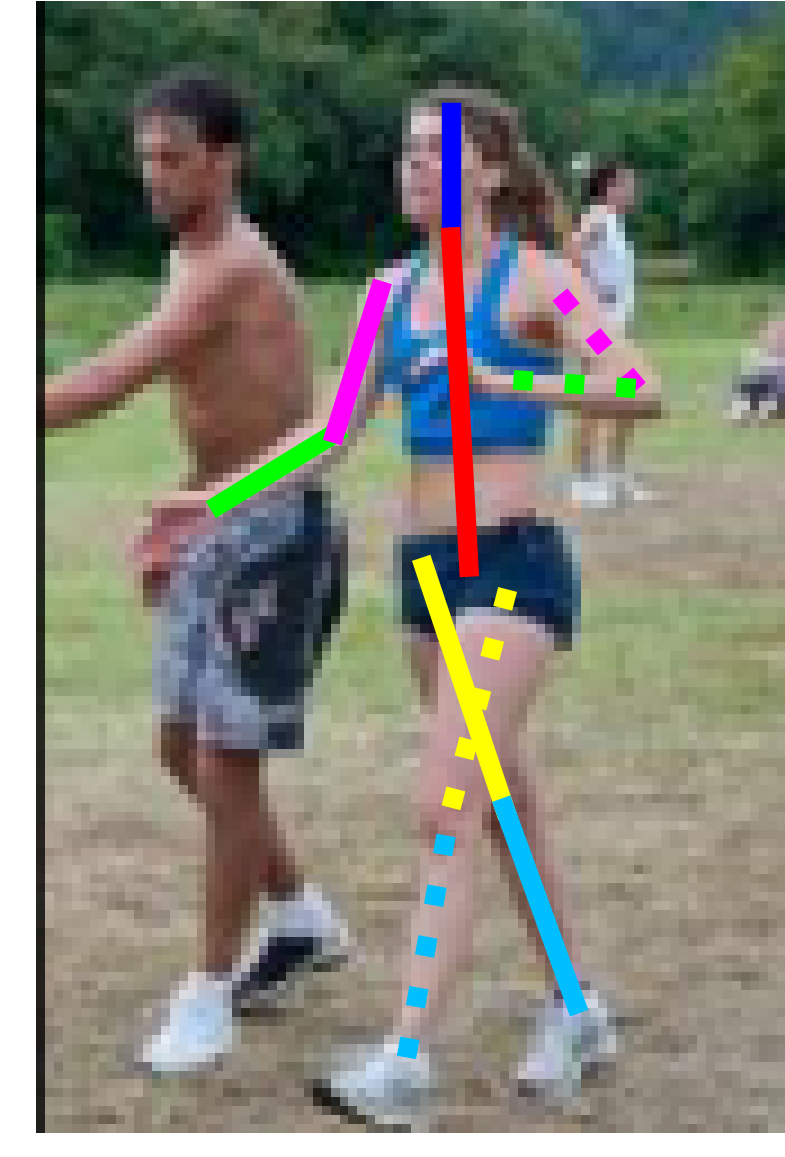} &
    \includegraphics[height=\leedsd\textheight]{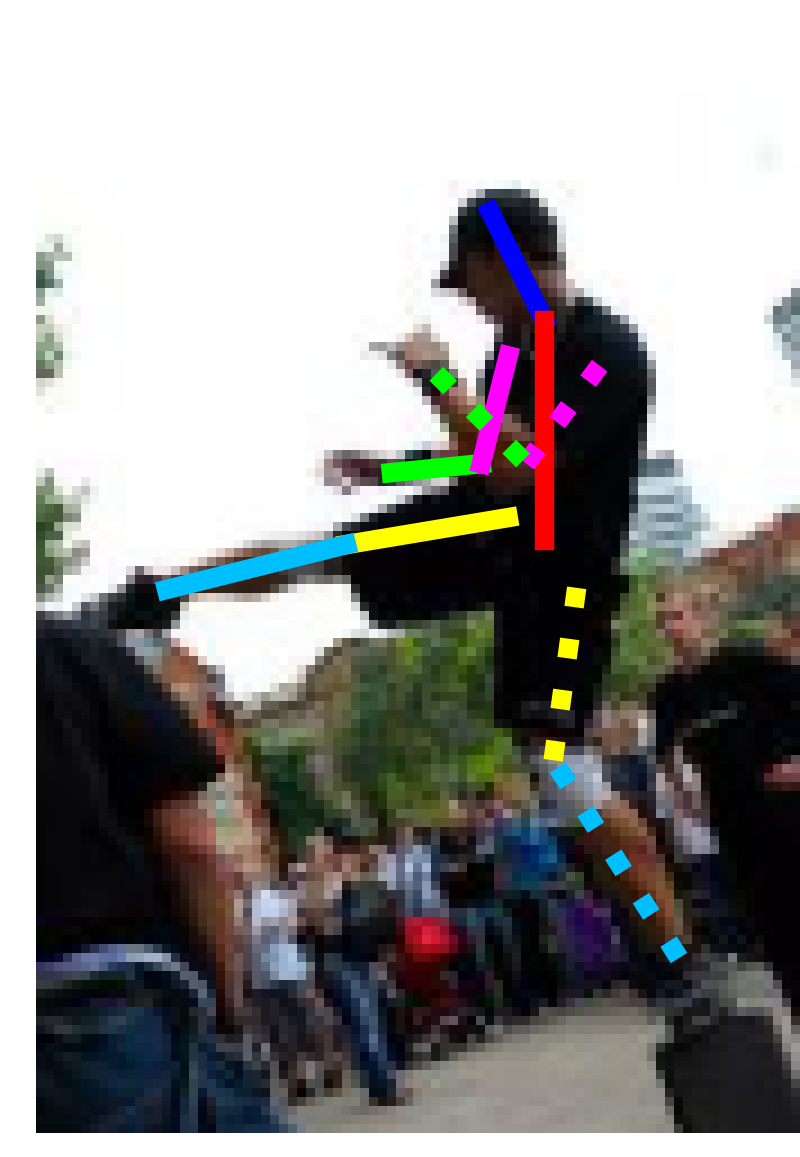} &
    \includegraphics[height=\leedsd\textheight]{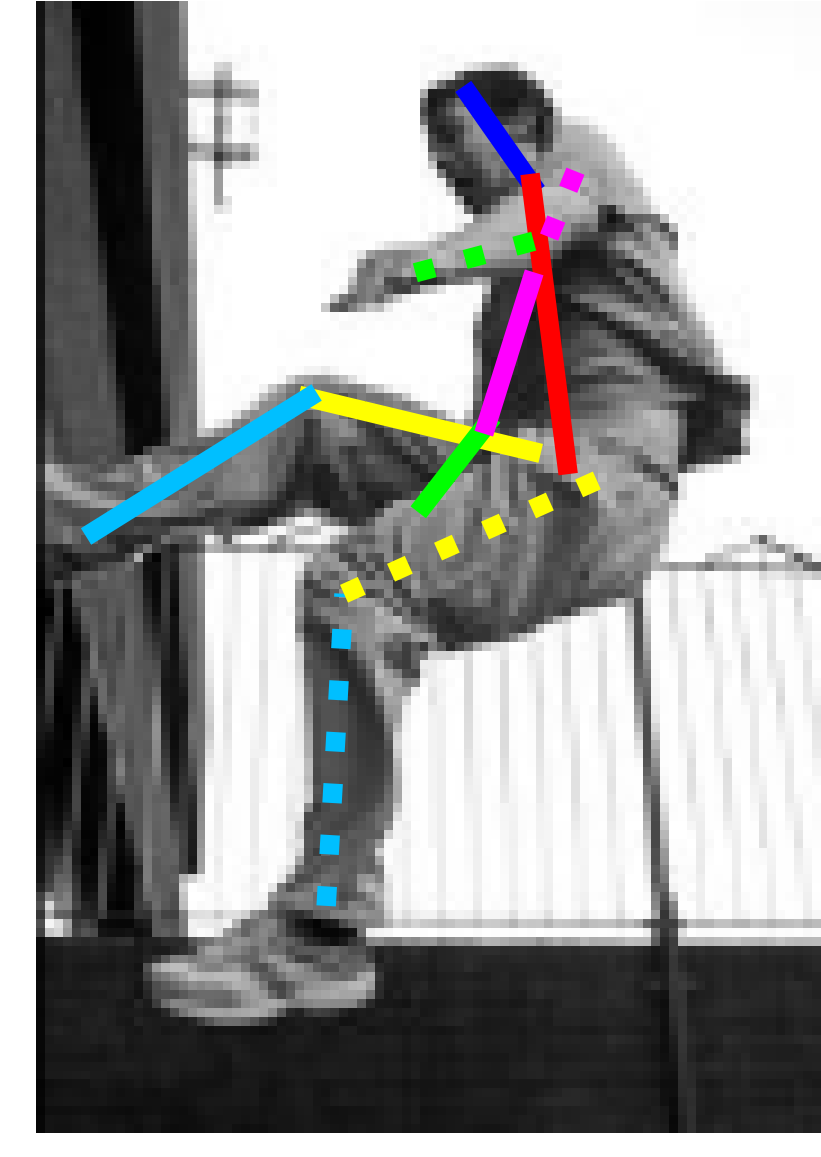} &
    \includegraphics[height=\leedsd\textheight]{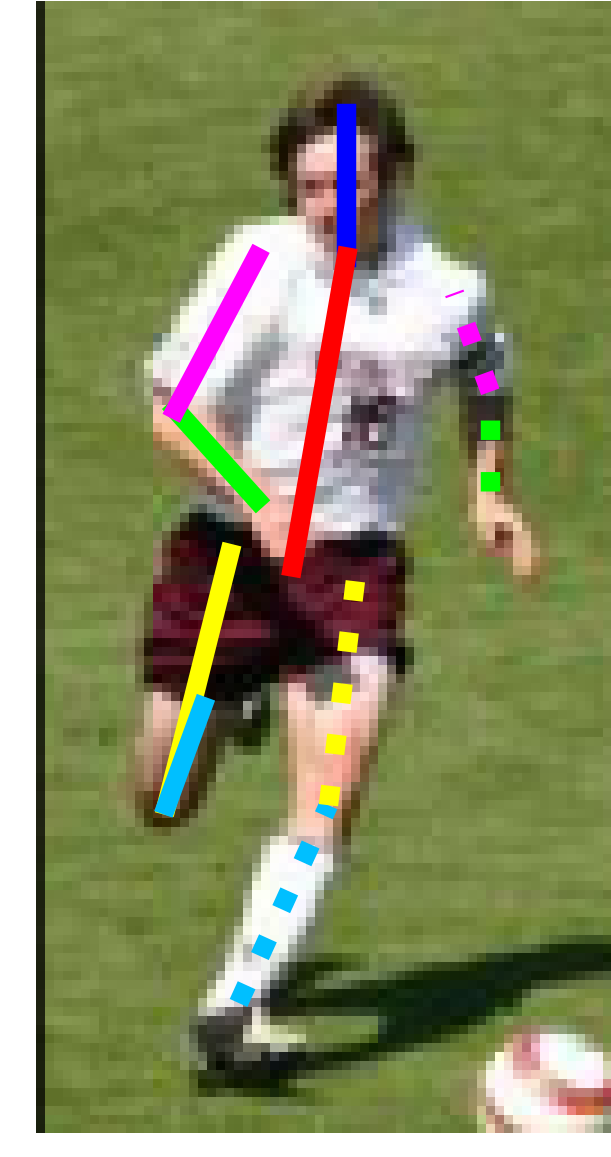}\\
\end{tabular*}
\begin{tabular*}{\textwidth}{c}
\hline
\end{tabular*}
\newcommand{\mpiia}{0.124}
\begin{tabular*}{\textwidth}{*{7}{c}}
    \multirow{1}{*}[0ex]{\rotatebox[origin=c]{90}{\textbf{MPII}}} &
    \includegraphics[height=\mpiia\textheight]{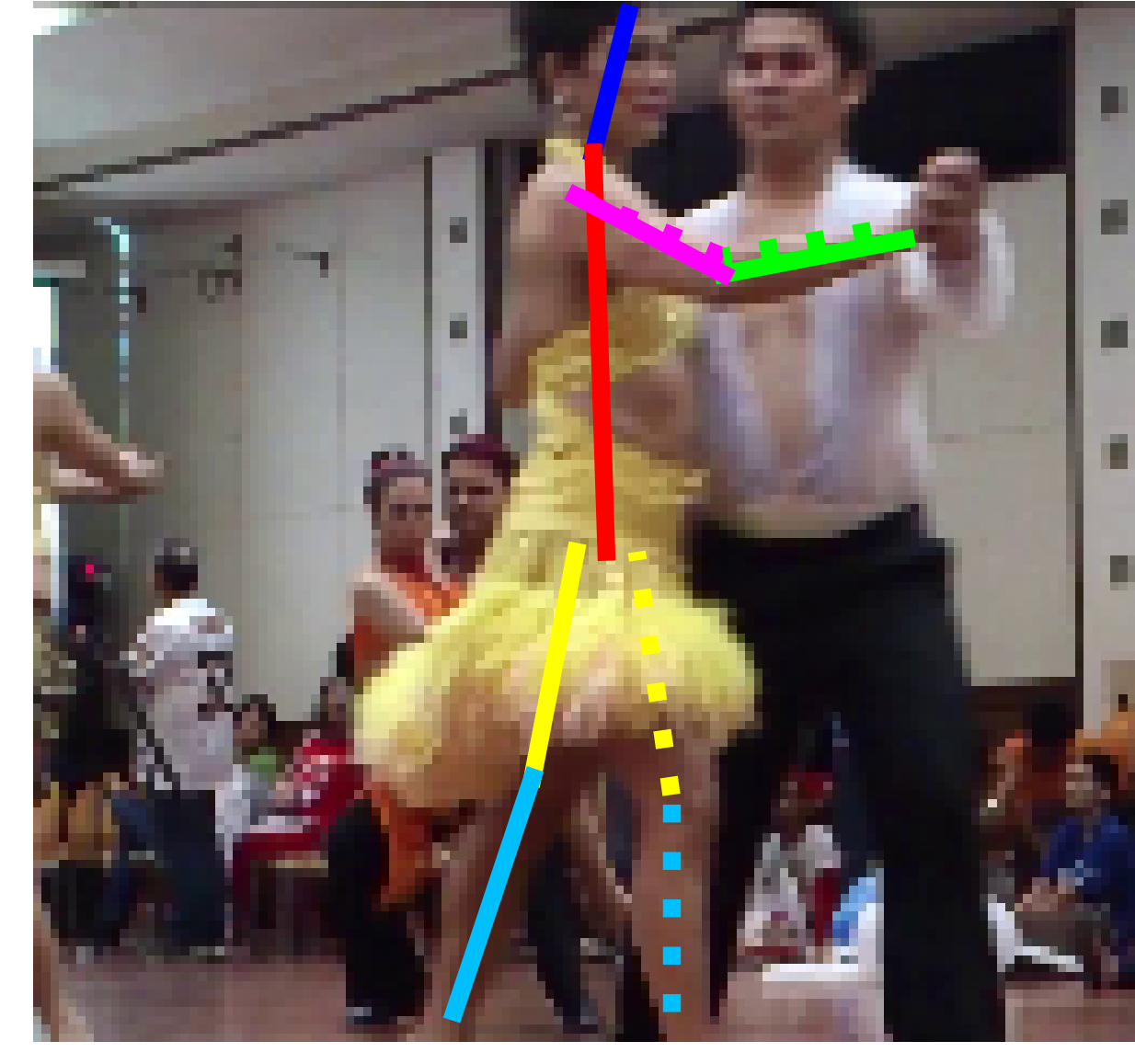} &
    \includegraphics[height=\mpiia\textheight]{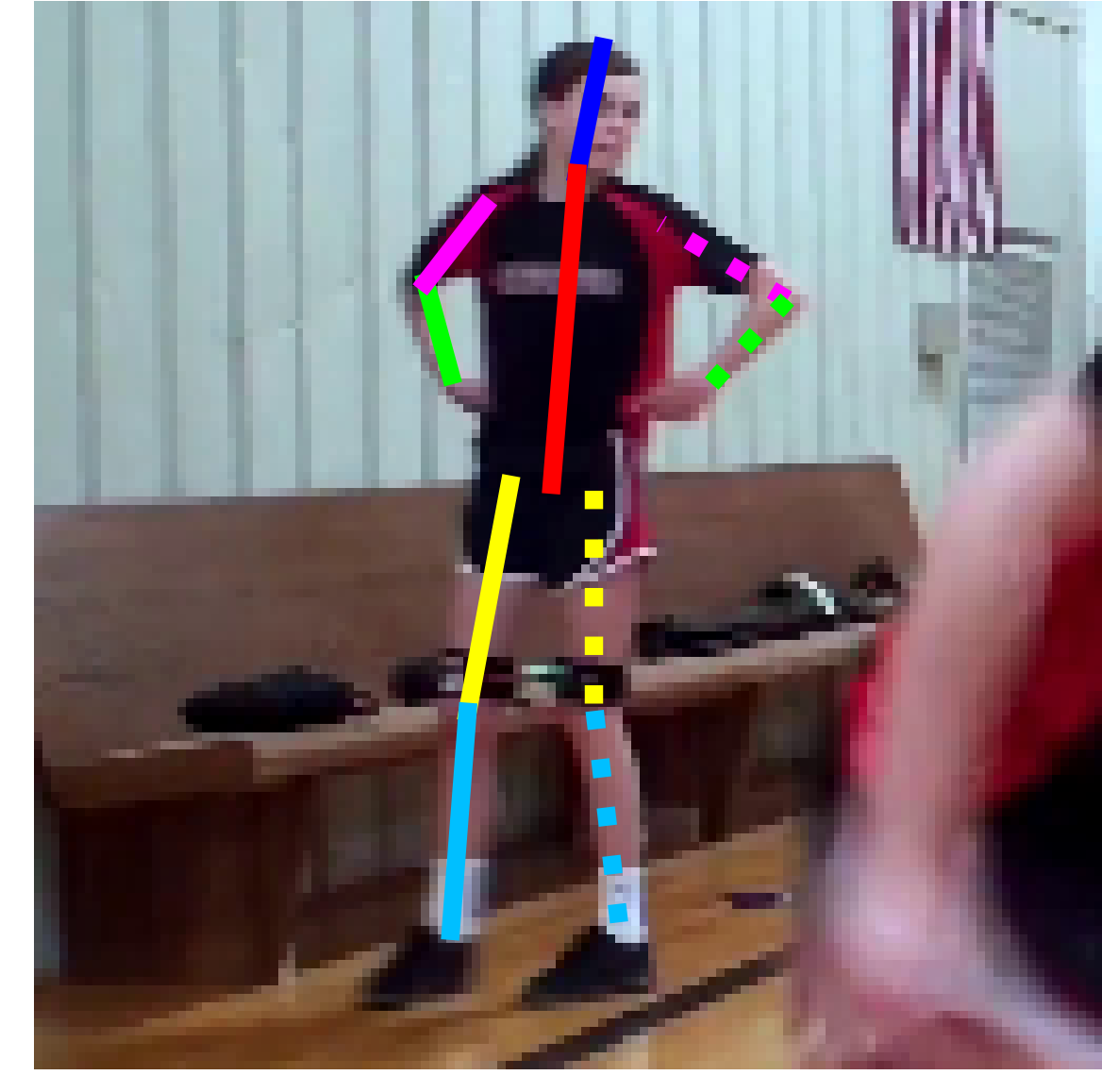} &
    \includegraphics[height=\mpiia\textheight]{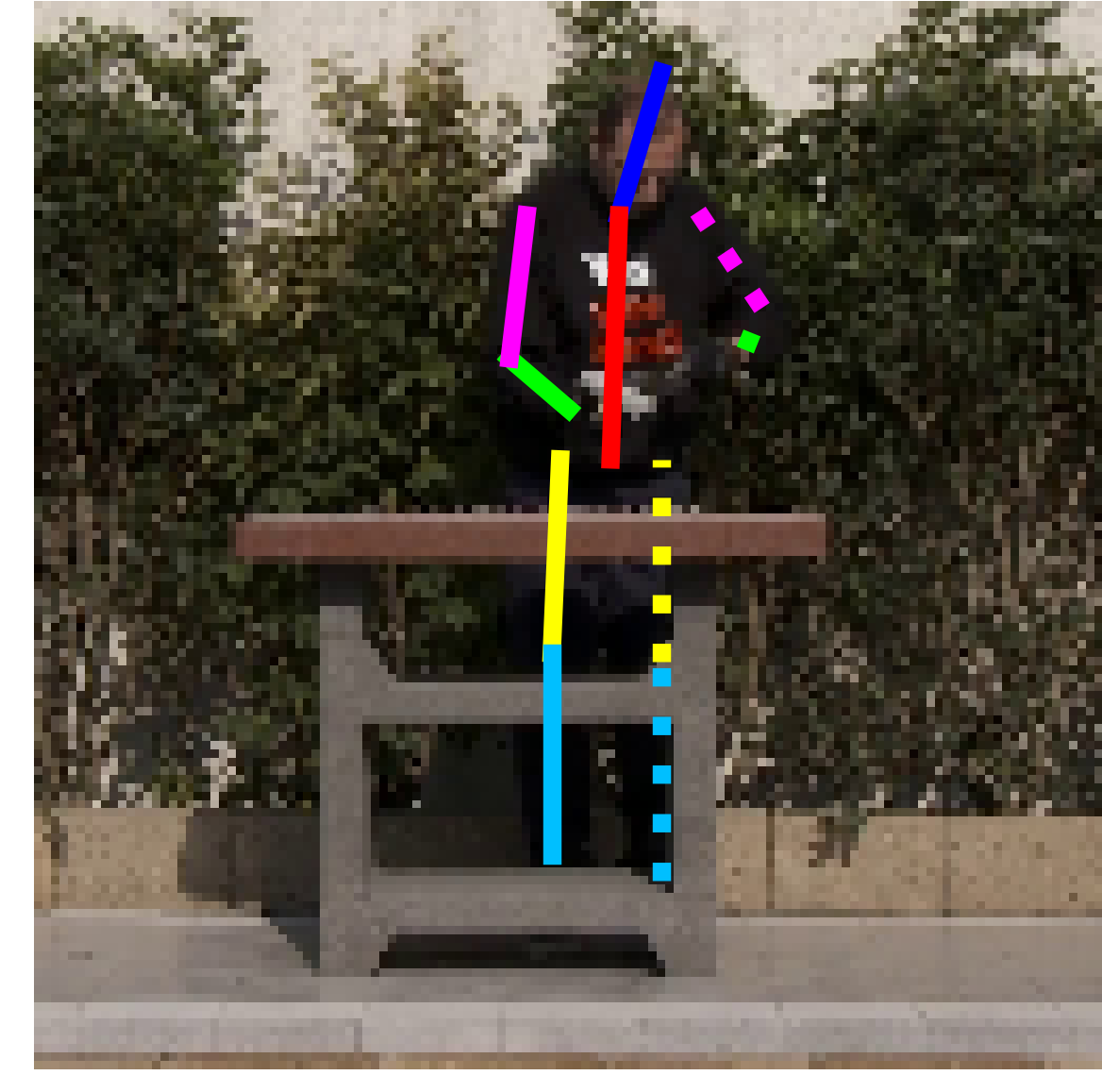} &
    \includegraphics[height=\mpiia\textheight]{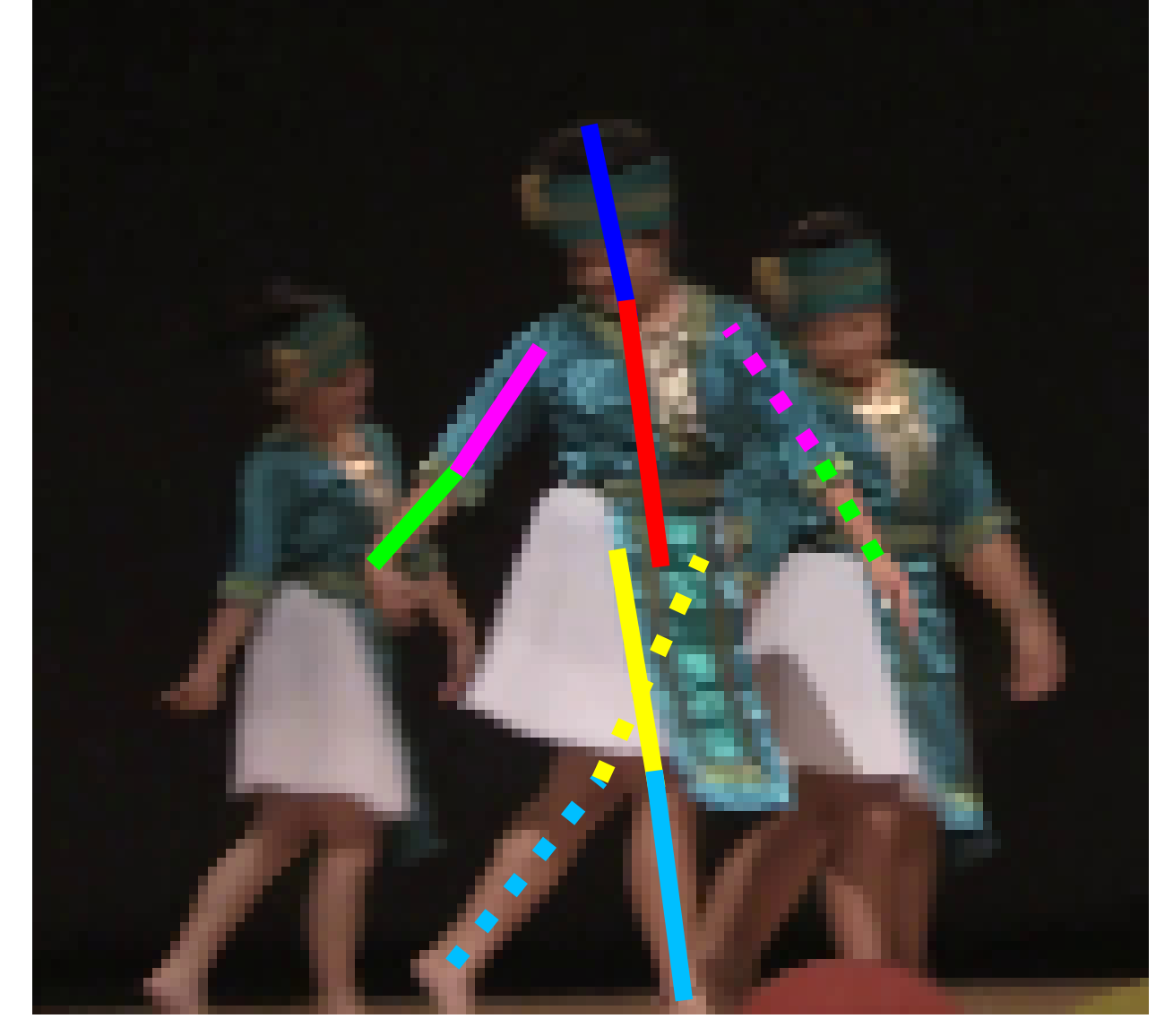} &
    \includegraphics[height=\mpiia\textheight]{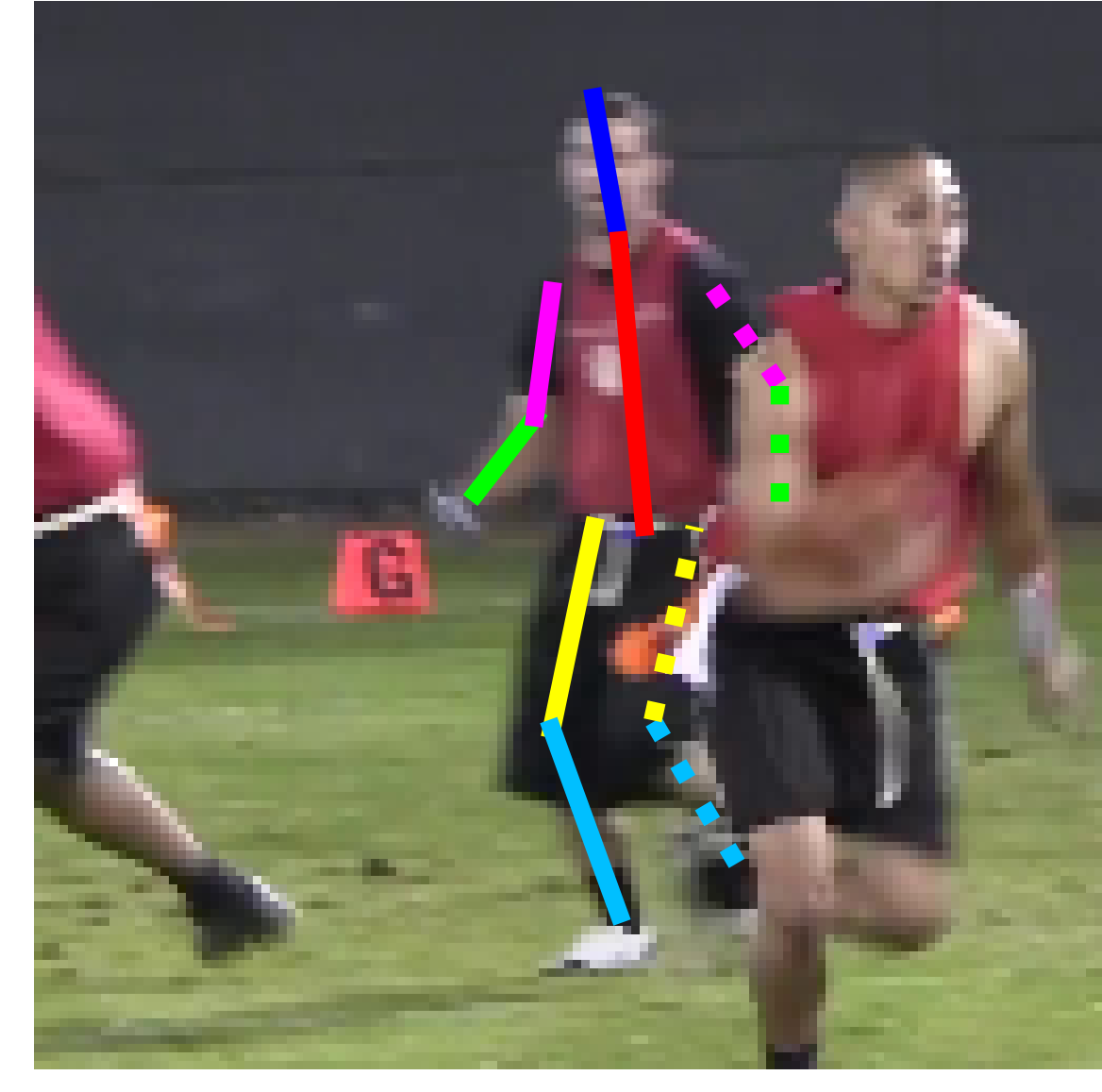} \\
\end{tabular*}
\newcommand{\mpiib}{0.113}
\begin{tabular*}{\textwidth}{*{7}{c}}
    \multirow{1}{*}[0ex]{\rotatebox[origin=c]{90}{\phantom{\textbf{MPII}}}} &
    \includegraphics[height=\mpiib\textheight]{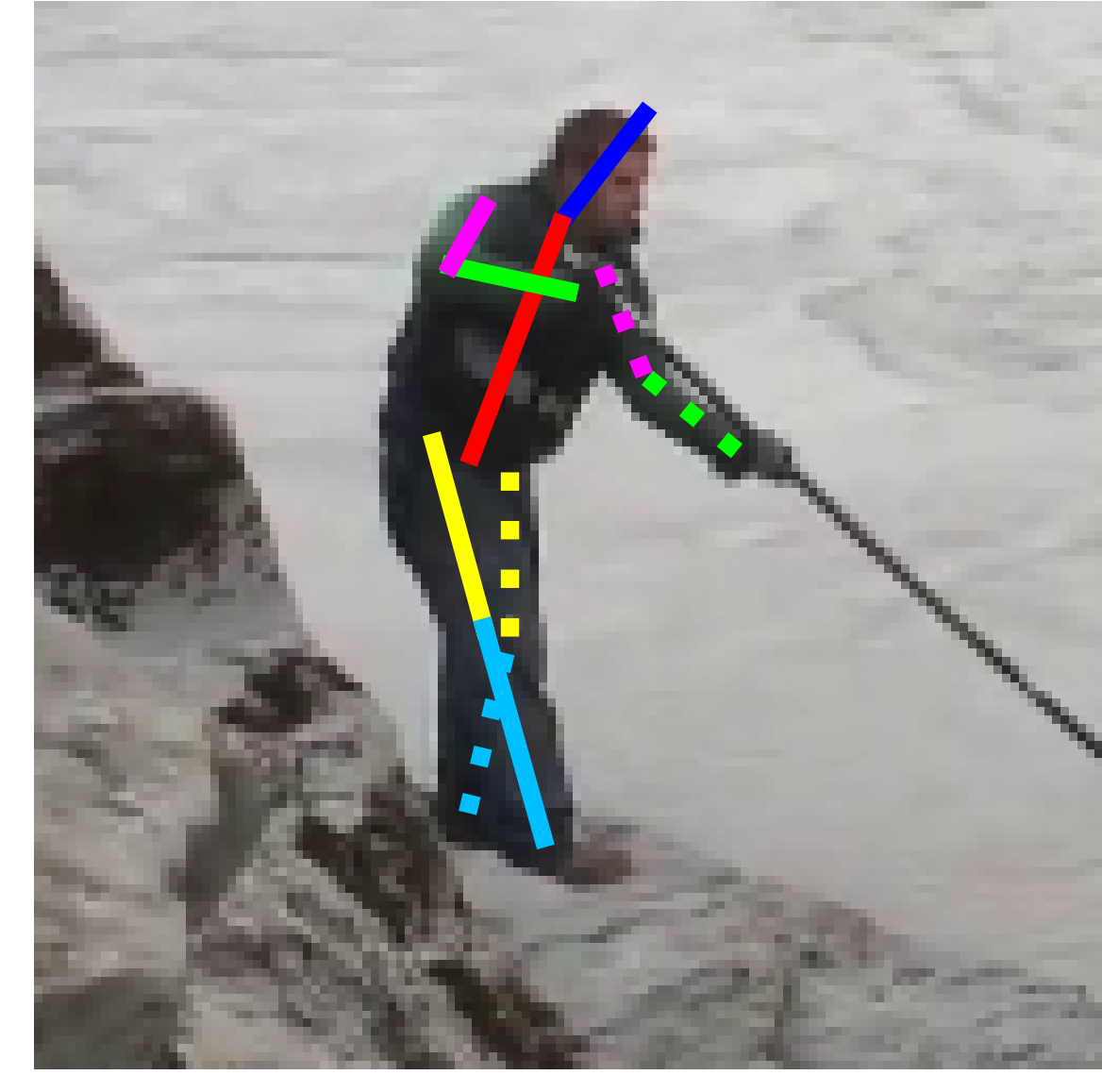} &
    \includegraphics[height=\mpiib\textheight]{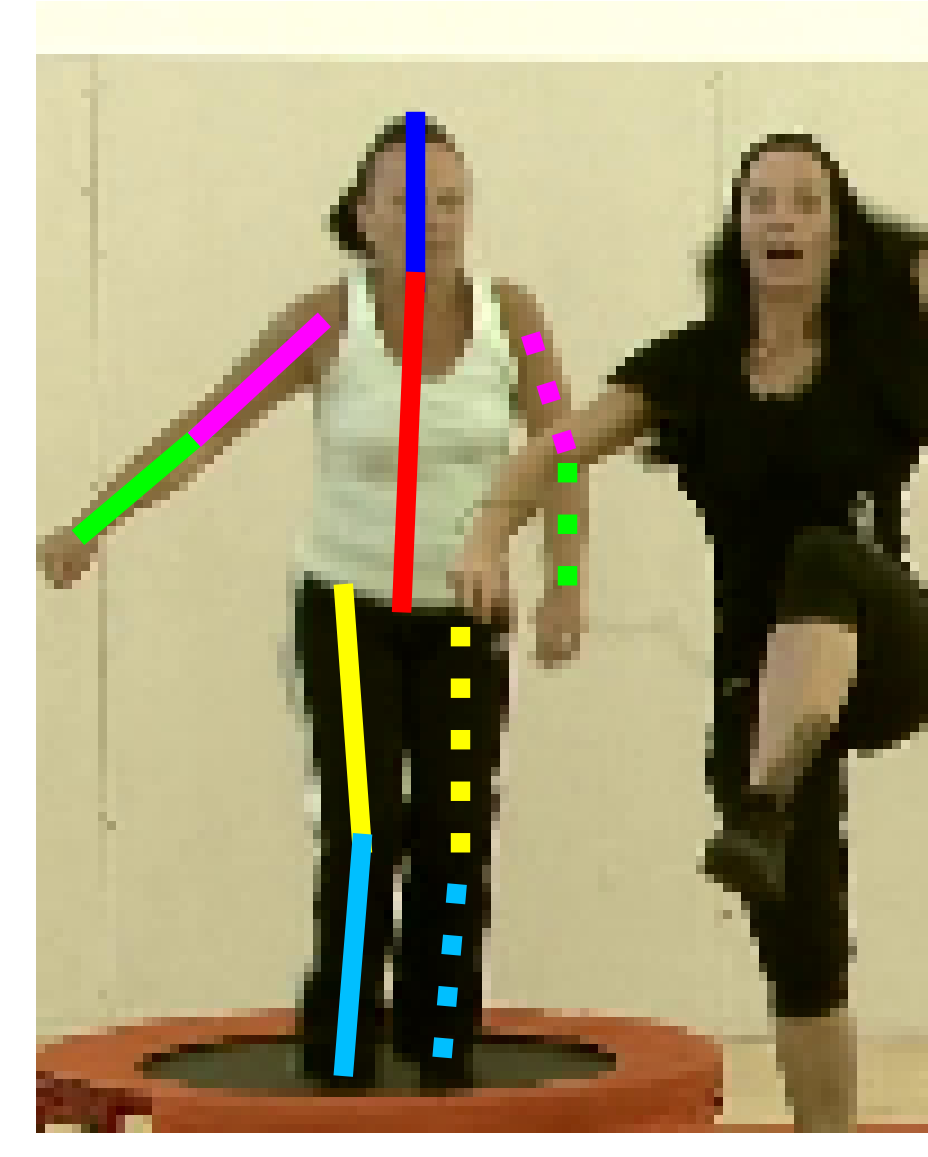} &
    \includegraphics[height=\mpiib\textheight]{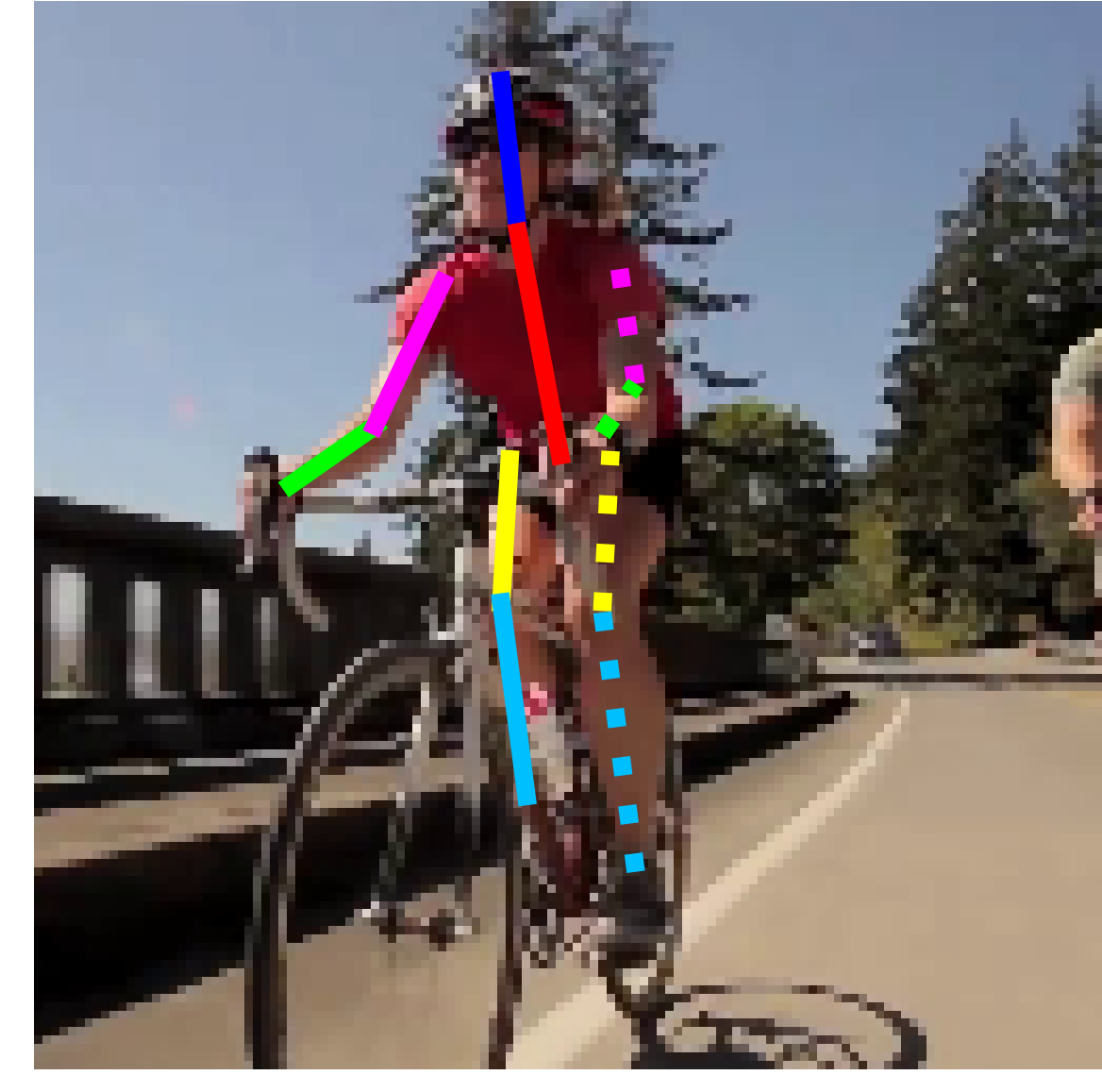} &
    \includegraphics[height=\mpiib\textheight]{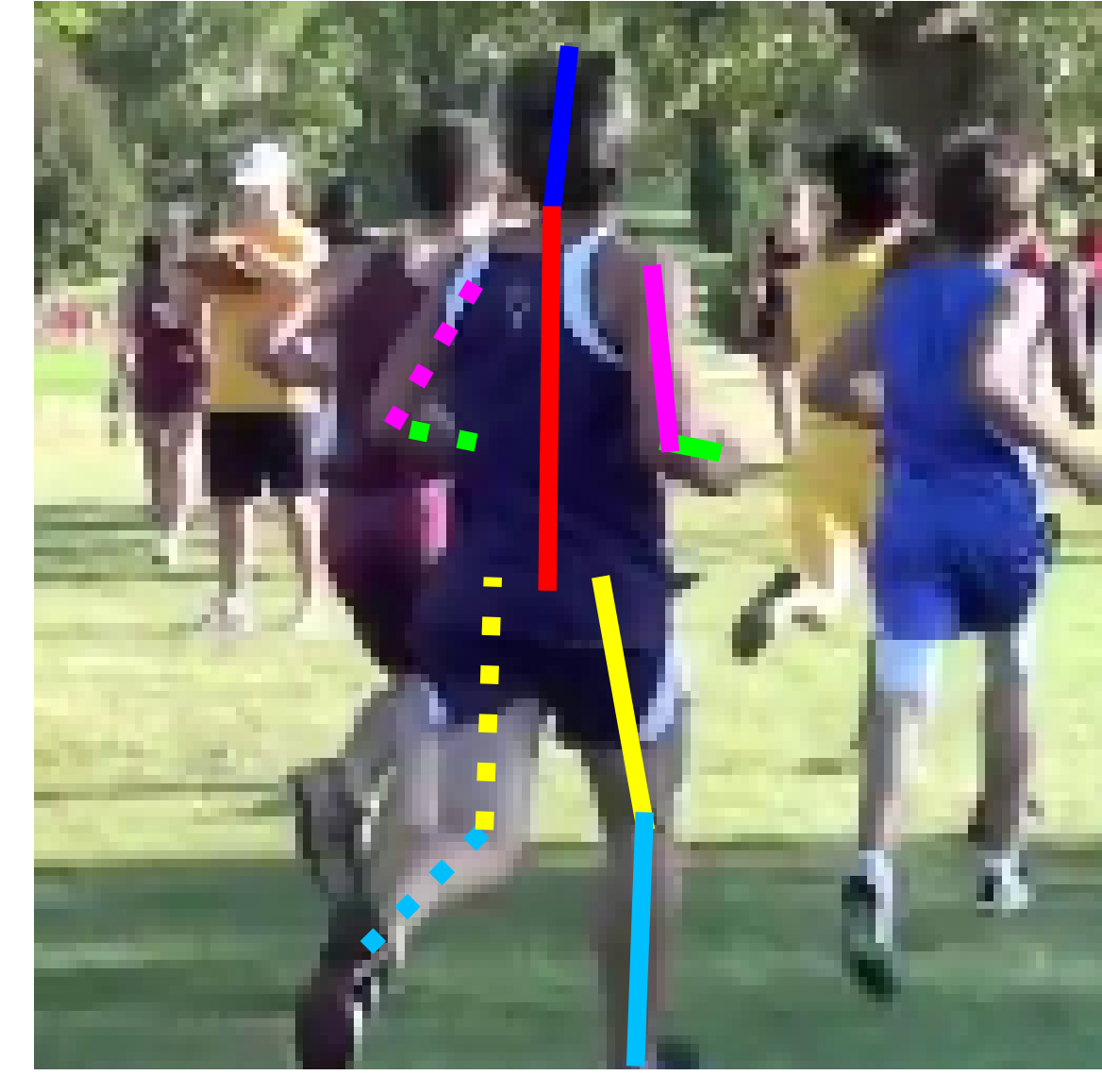} &
    \includegraphics[height=\mpiib\textheight]{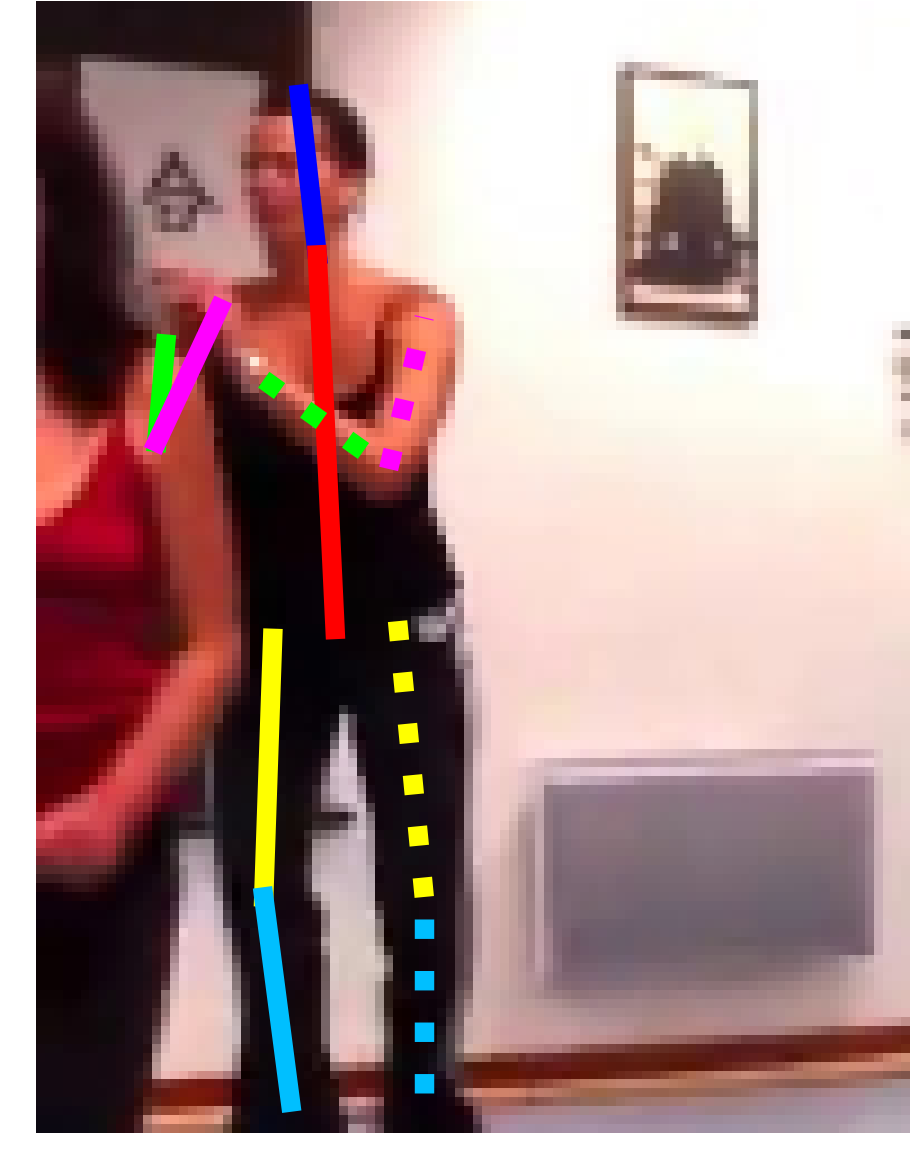} &
    \includegraphics[height=\mpiib\textheight]{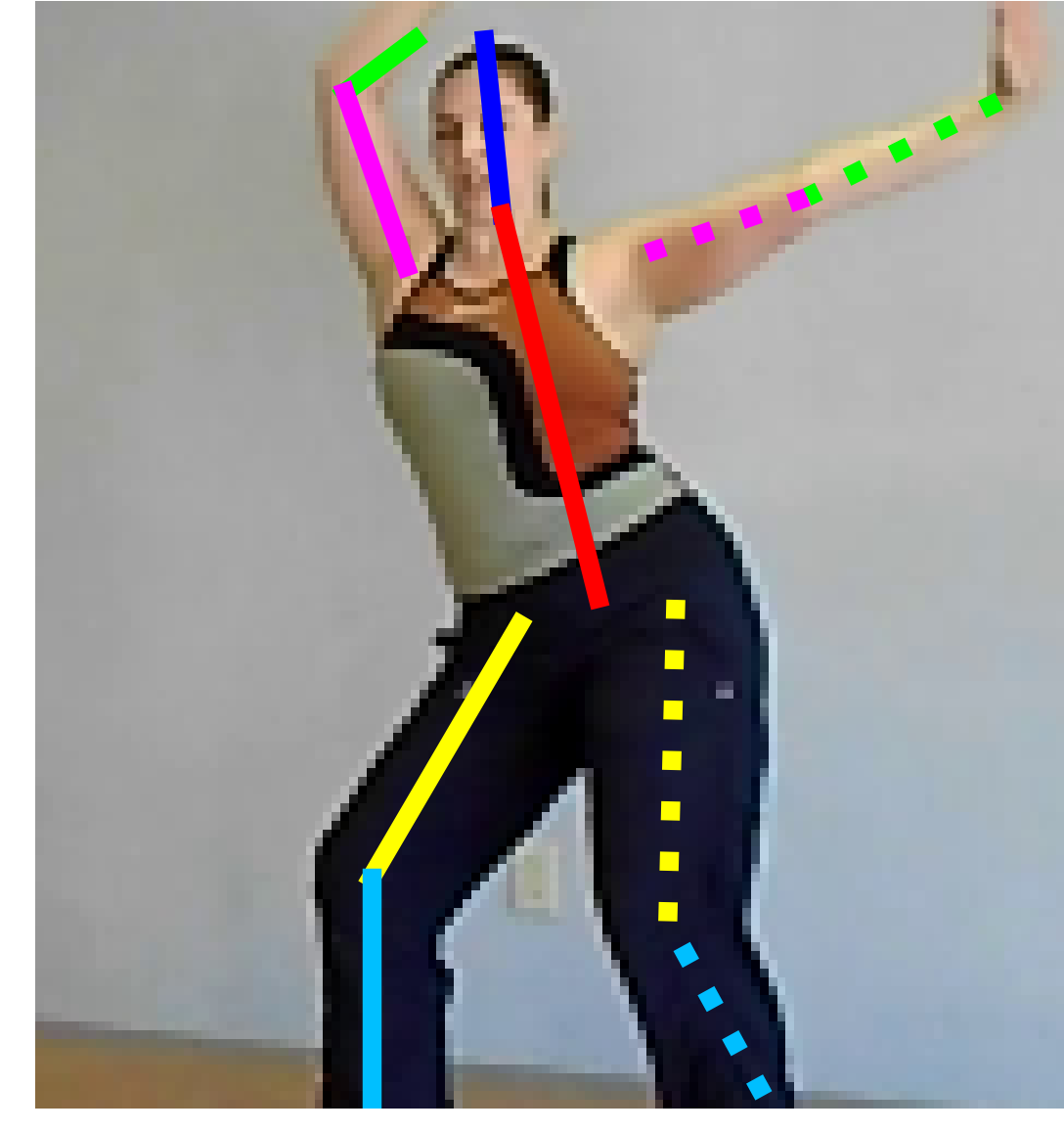} \\
\end{tabular*}
\caption{A sample of correctly predicted poses using \repose on LSP and MPII examples not seen during training. The body parts are uniquely color coded, while line style encodes the person centric orientation of a part, \ie~solid and dashed for right and left, respectively.}
\label{fig:repose_correct}
\end{figure*}

\begin{figure*}[h]
\vspace{-2ex}
\newcommand{\leedsa}{0.12}
\begin{tabular*}{\textwidth}{c@{\hspace{0.5ex}}cccc|cc}
    \multirow{1}{*}[12ex]{\rotatebox[origin=c]{90}{Ground truth}} &
    \includegraphics[height=\leedsa\textheight]{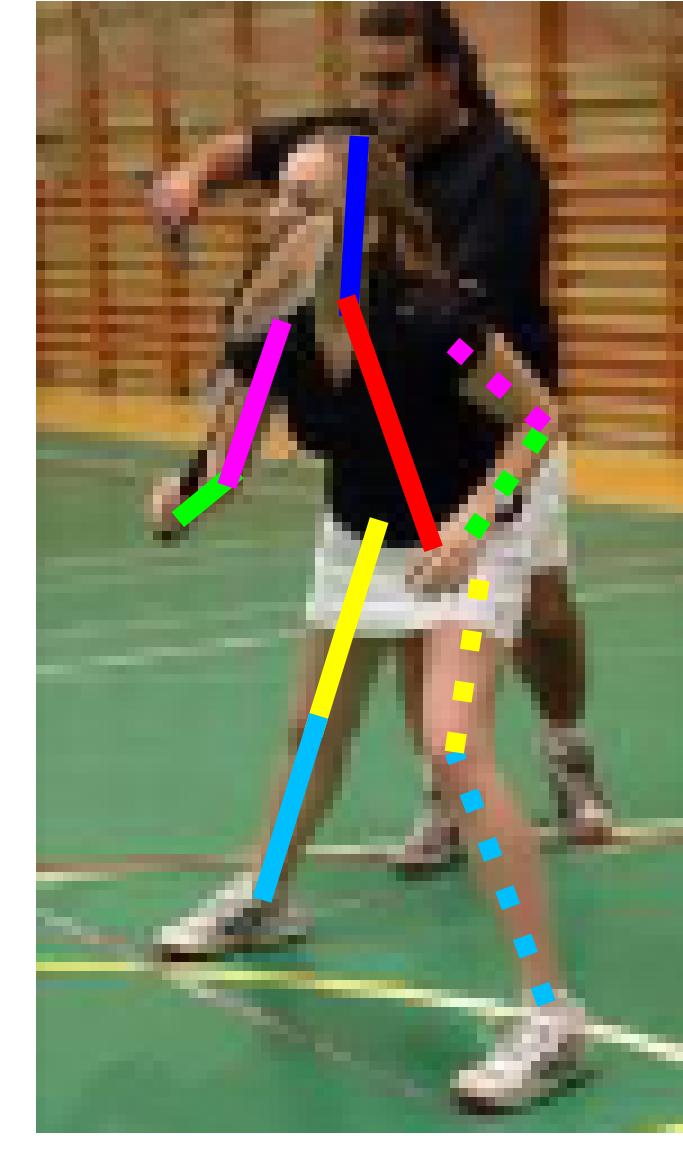} &
    \includegraphics[height=\leedsa\textheight]{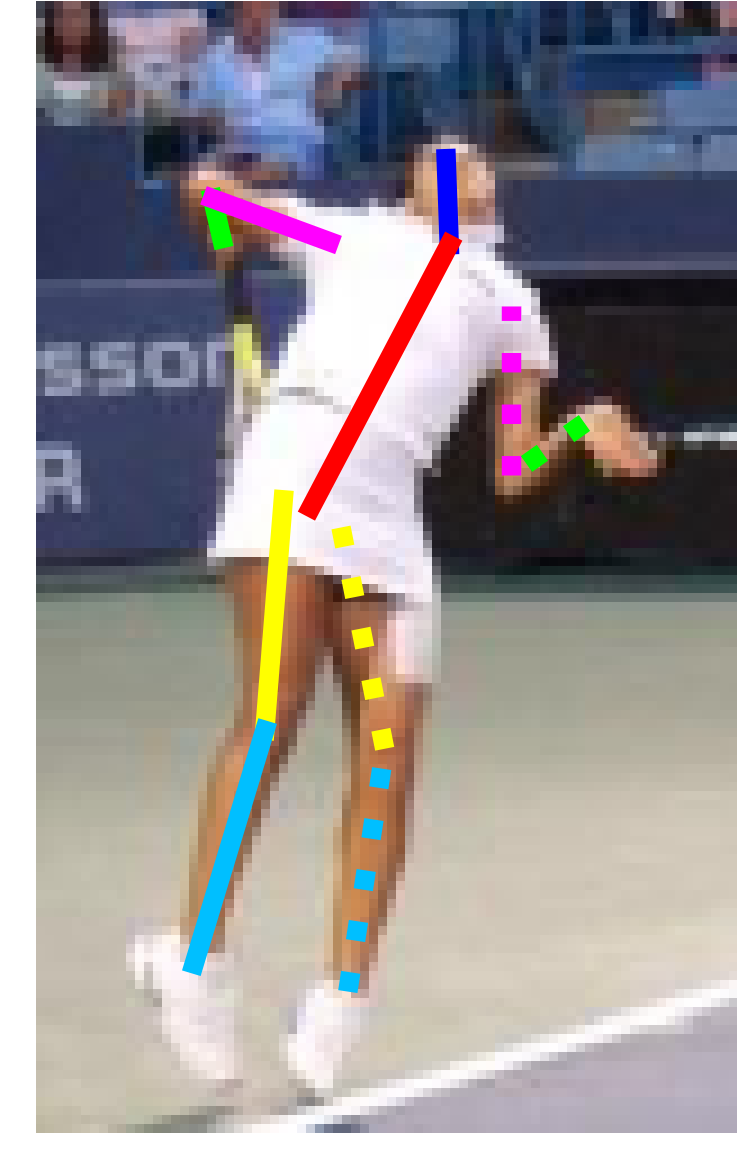} &
    \includegraphics[height=\leedsa\textheight]{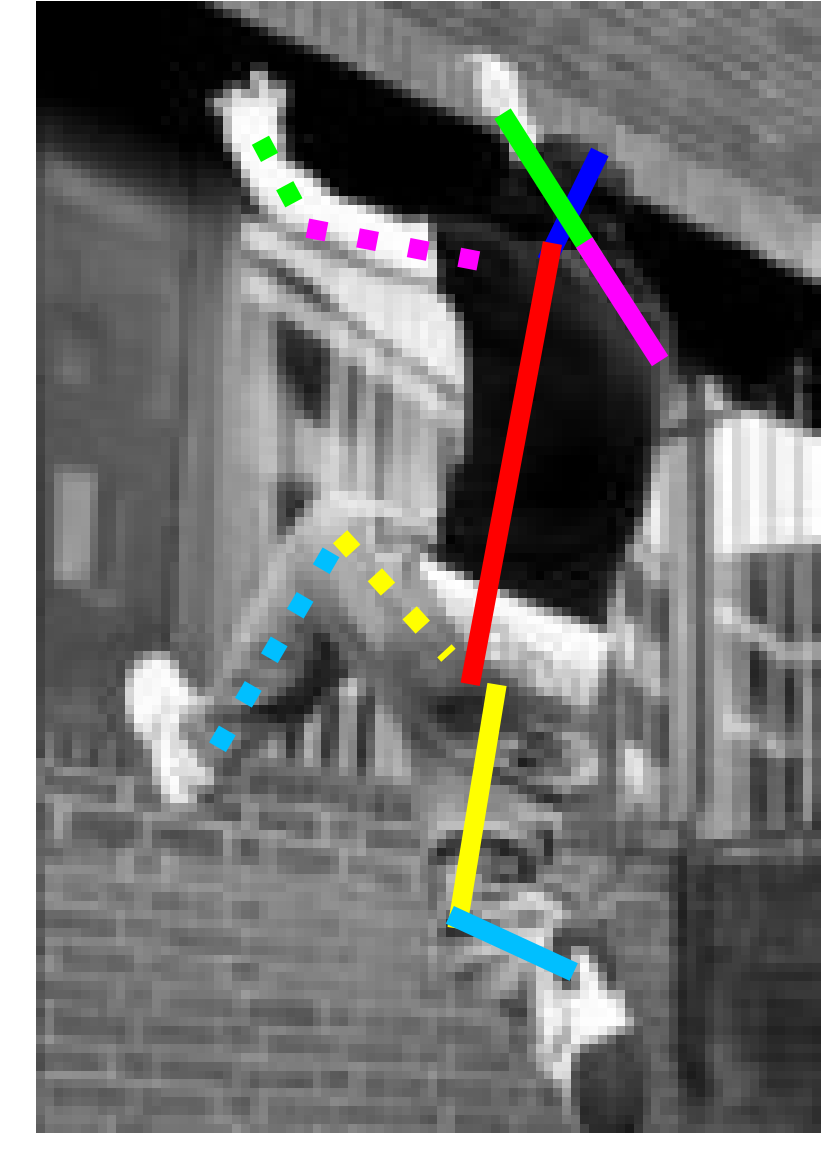} &
    \includegraphics[height=\leedsa\textheight]{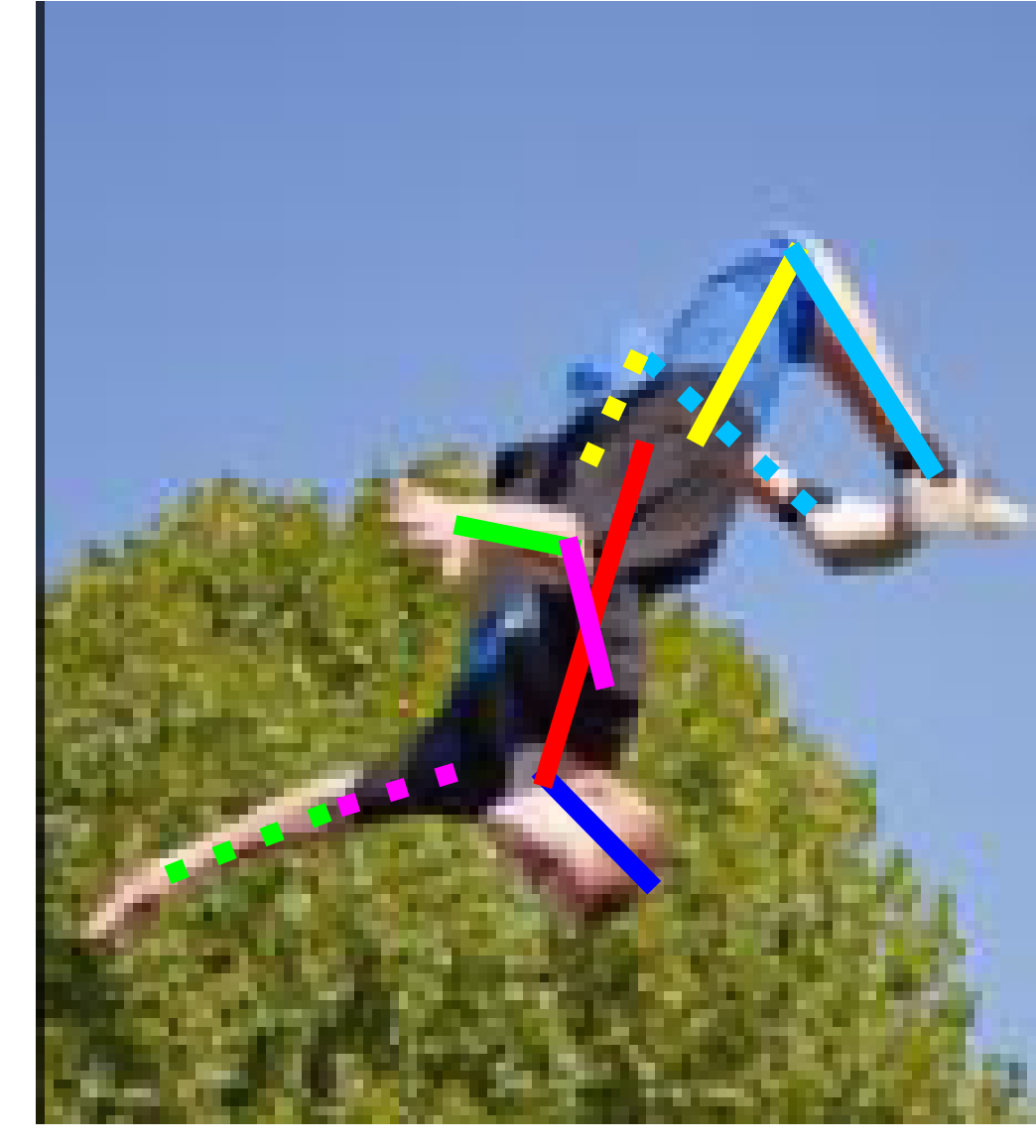} &
    \includegraphics[height=\leedsa\textheight]{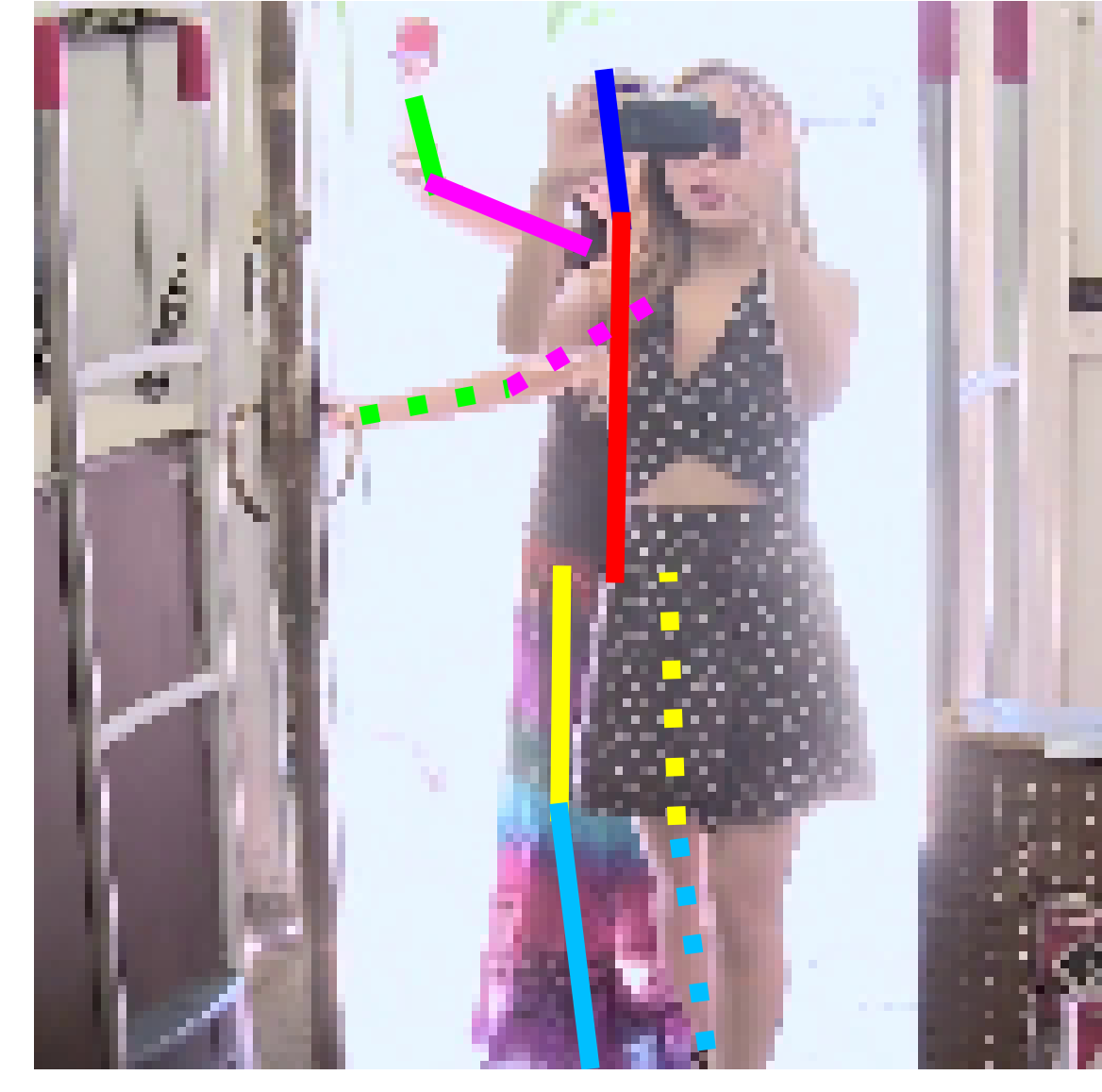} &
    \includegraphics[height=\leedsa\textheight]{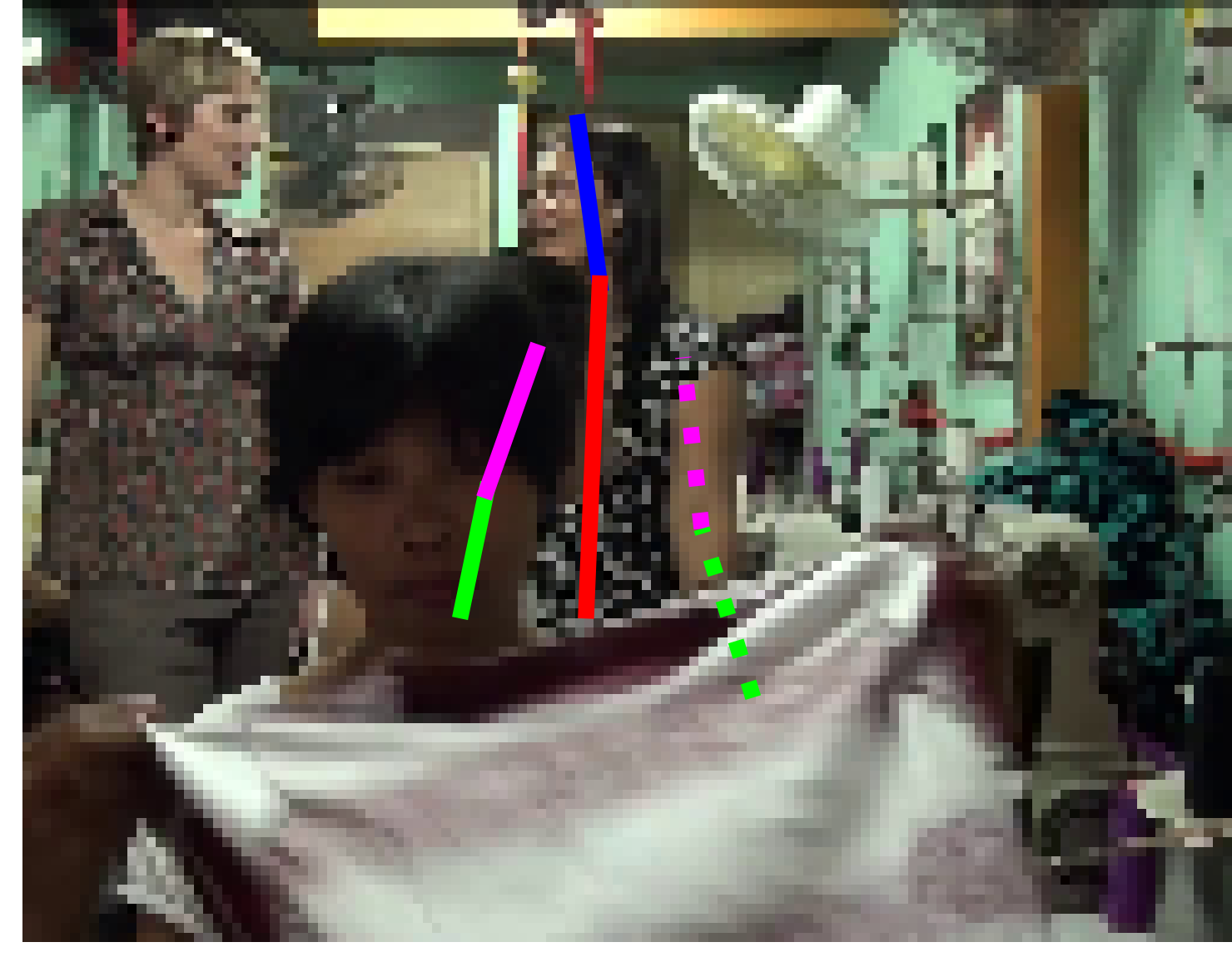}
    \\
    \multirow{1}{*}[13ex]{\rotatebox[origin=c]{90}{Incorrect pose}} &
    \includegraphics[height=\leedsa\textheight]{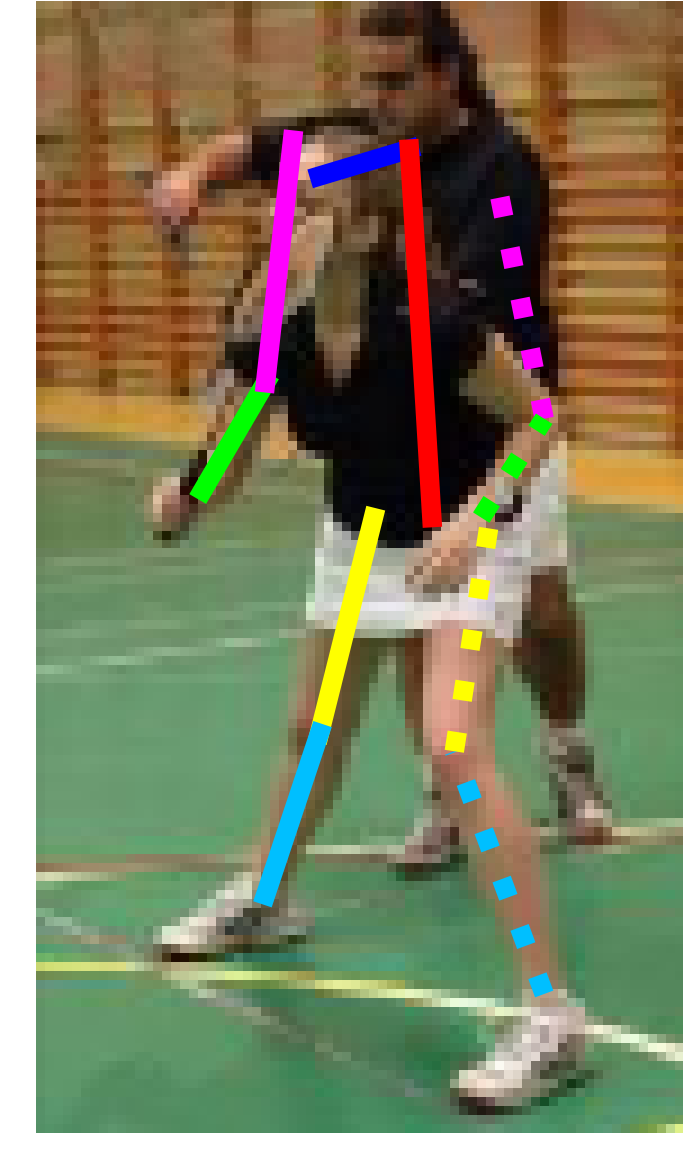} &
    \includegraphics[height=\leedsa\textheight]{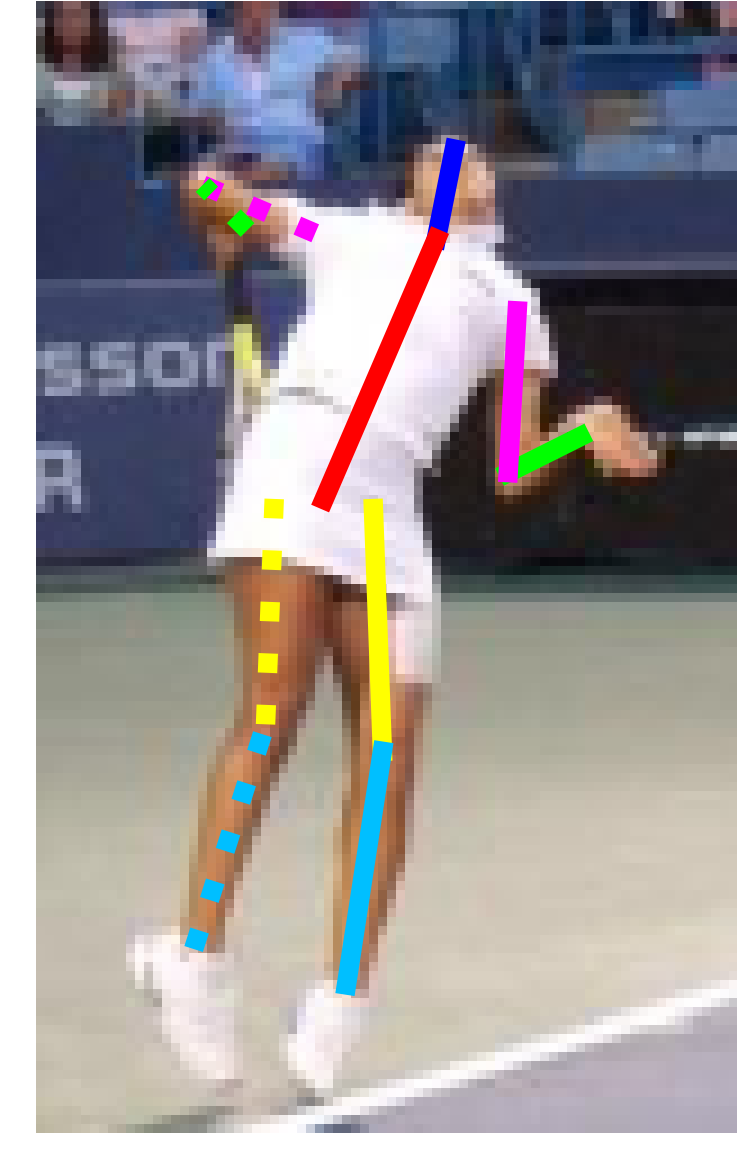} &
    \includegraphics[height=\leedsa\textheight]{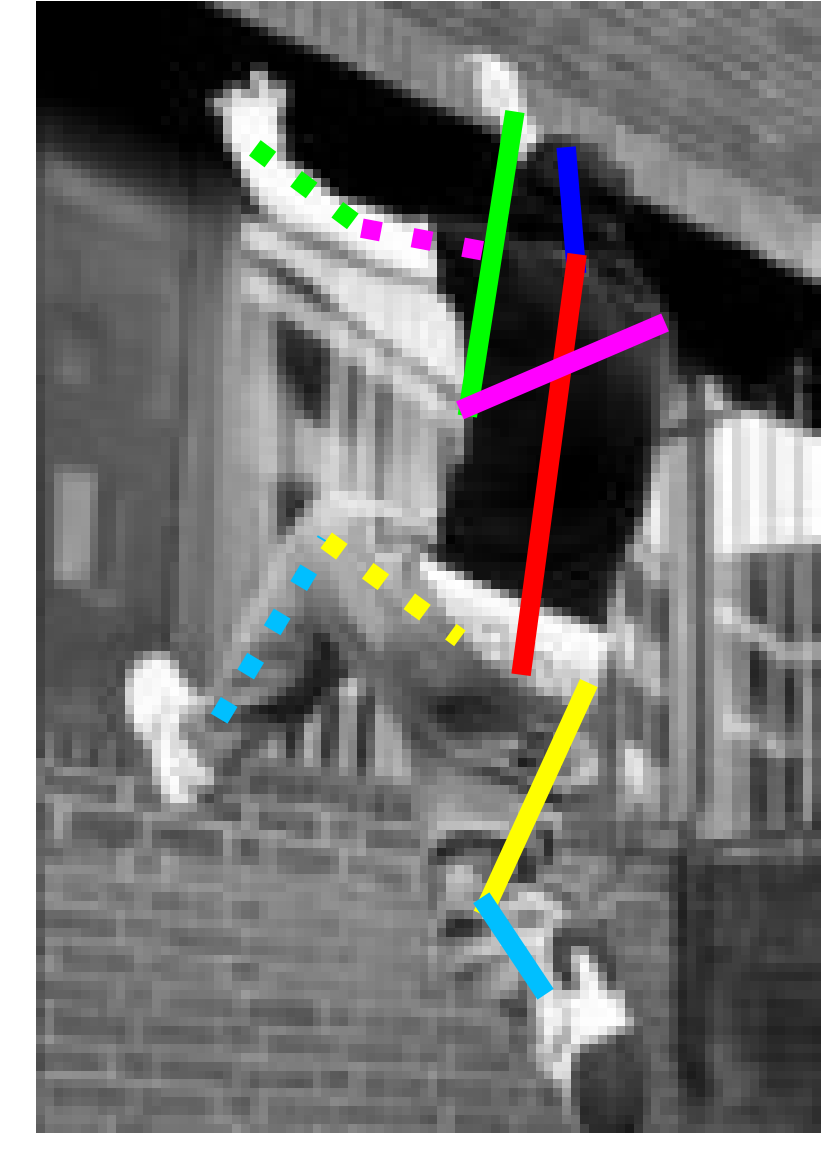} &
    \includegraphics[height=\leedsa\textheight]{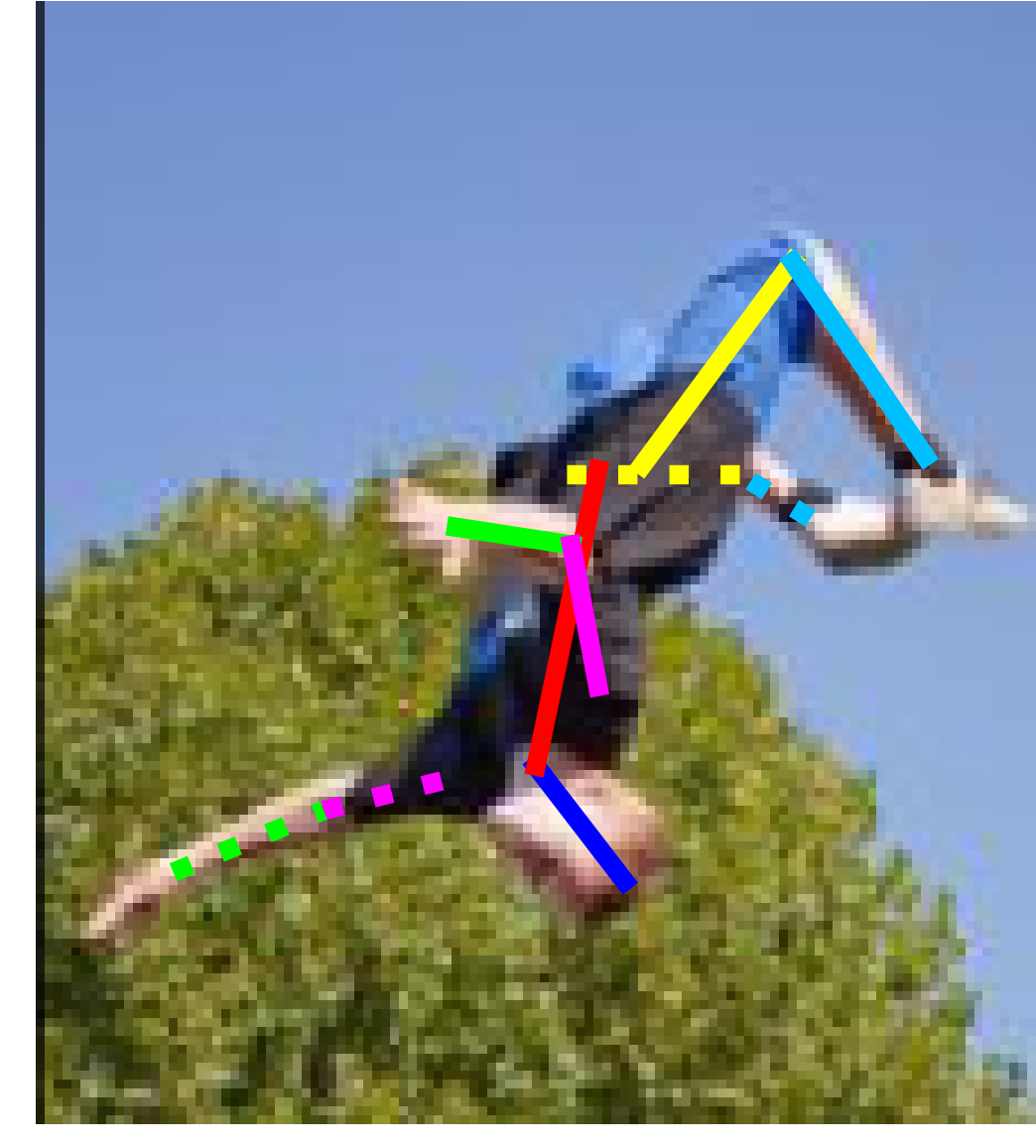} &
    \includegraphics[height=\leedsa\textheight]{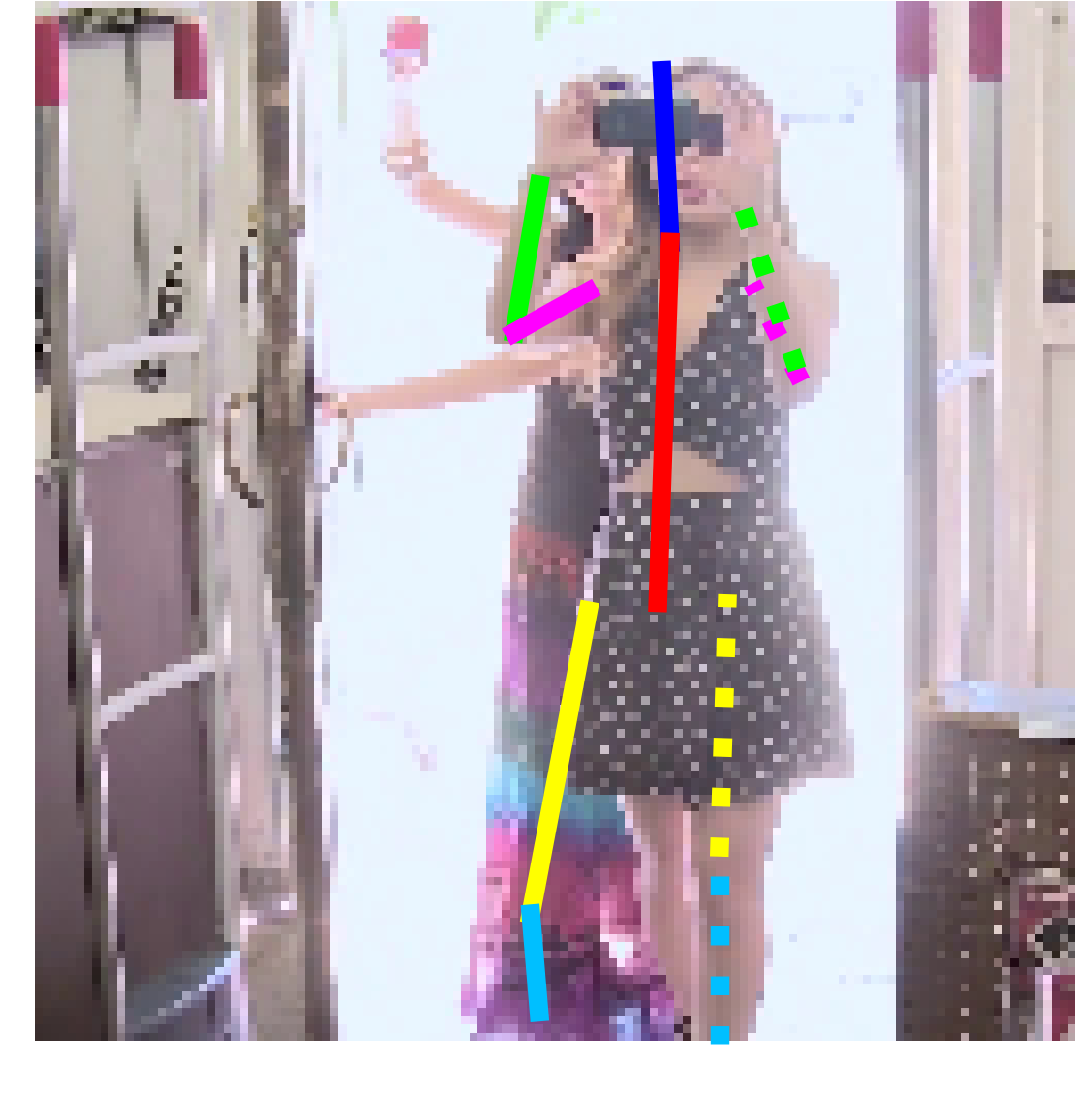} &
    \includegraphics[height=\leedsa\textheight]{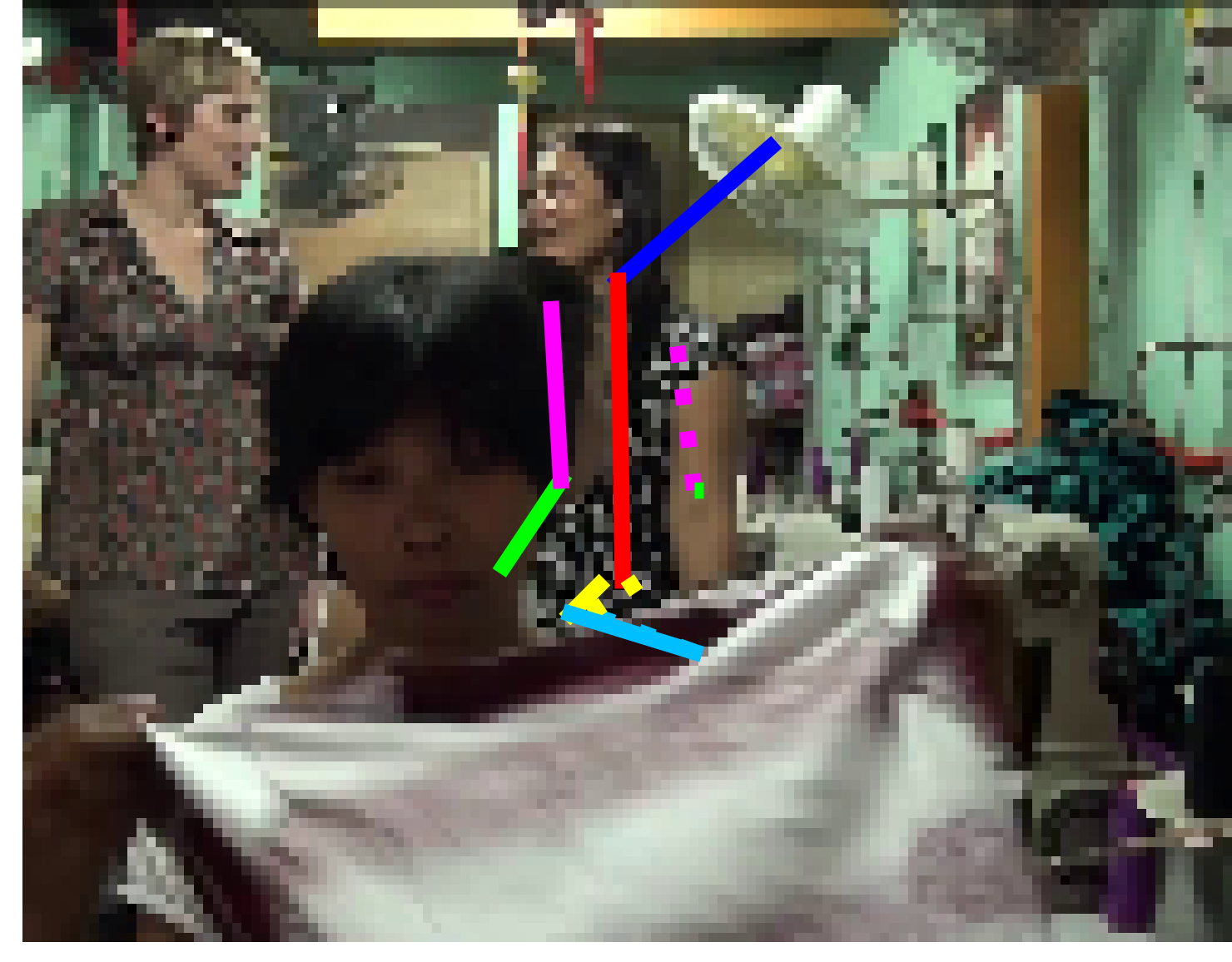}
    \\
    & \multicolumn{4}{c}{\textbf{LSP}} & \multicolumn{2}{c}{\textbf{MPII}}\\
\end{tabular*}
\caption{shows a sample of incorrectly predicted poses. The failure cases usually involve crowed scenes, highly occluded pose, or a left/right flip of the whole body pose.}
\label{fig:repose_failed}
\vspace{-2ex}
\end{figure*}

\paragraph{Qualitative Evaluations} Figure~\ref{fig:repose_correct} shows a sample of correctly predicted poses using examples not seen during training on both datasets. Our network failed predictions (see Figure~\ref{fig:repose_failed} for a sample) are skewed towards scenes with large number of occluded keypoints or spatially inseparable poses. Intuitively,  kinematically updating features in those cases does not perform as well, since there are not enough accurately localized keypoints to enhance the prediction of the remaining ones.

\subsection{Ablation Study}
\vspace{-1ex}
We conduct ablation study to show the effectiveness of different configurations and components of \reposenospace.
\vspace{-1ex}

\paragraph{Coarsest Resolution for Kinematic Updates}
One important question is, \emph{what is the coarsest resolution at which our kinematic features updates are the most effective?} 
Table~\ref{tbl:repose_n_scales} shows the results of applying the updates at different resolutions. On the one hand, it is clear that applying kinematic updates at $8 \times 8$ resolution degrades the performance significantly, on average by $2.46\%$. 
If we were to randomly place $14$ keypoints on an $8 \times 8$ pixel grid, then there is more than even chance\footnote{Assuming keypoints are i.i.d.~this chance is $ 1 - \binom{n^2}{K} K!/ n^{2K}$.} that two or more keypoints will be placed on the same pixel.
For $K=14$ as in LSP, this chance is $78\%$ and $30\%$ for the $8 \times 8$ and $16 \times 16$ resolutions, respectively.
On the other hand, at the $32 \times 32$ resolution the number of FLOPS increases by $32\%$ compared to the $16 \times 16$ resolution.
Furthermore, applying the updates at higher resolutions could adversely affect the performance, since the receptive field would not be large enough to capture all neighbouring keypoints to properly correlate their features.

\paragraph{Feature Update Step}
We tried using different number of convolutional blocks in each kinematic update step \eqref{eq:update}.
As shown in Table~\ref{tbl:fus_n_convs}, increasing the number of blocks to more than four degrades performance. 
We also tested different strategies of applying the residual connection of the update. Table~\ref{tbl:fus_update_approach} shows the results of using trainable weights as in \eqref{eq:update}, adding the old features to the updated ones, or completely replacing the old ones. Using trainable weights leads to a significant performance gain, specially on MPII where occlusions are more common. 
\vspace{-0.5ex}

\paragraph{Network Stacking}
Network stacking~\cite{stackedhourglass16} is a popular technique to increase network performance. For completeness, Table~\ref{tbl:repose_stacking} shows results for stacking. \repose reaches comparable results to state-of-the-art methods~\cite{featurepyramids17,avddataugment18} on LSP, while only using $66\%$ of the required trainable weights.
Finally, stacked \repose networks train significantly faster, requiring less than half the number of steps compared to a single network.
\vspace{-0.5ex}

\paragraph{Kinematically Ordered vs Sequential Updates}
To show how ordering the convolutions helps performance, we replaced the features update step by a series of sequential convolutional blocks, such that the resulting model would have roughly the same number of parameters. 
The sequential model reached $85.82\%$ and $84.94\%$ on LSP and MPII, respectively, which is a significant reduction in performance compared to \repose with kinematically ordered updates. 
Thus, indicating how crucial it is to properly structure the convolutional blocks to get better pose estimation models.

\paragraph{Updates Ordering} Instead of using the predefined ordering in Figure~\ref{fig:connectivity}, where we started from the hips and propagated outwards, we tried a top down approach where started from the head and moved towards the ankles and wrists. The alternative ordering led to a decrease in performance by $0.42\%$ and $0.70\%$ on the LSP and MPII datasets, respectively.
\begin{table}[h]
\centering
\begin{tabular}{|c|c|c|c|c|}
\hline
\textbf{Coarsest}  &  
\multirow{2}{*}{\textbf{Leeds}} &
\multirow{2}{*}{\textbf{MPII}} &
\multirow{2}{*}{\textbf{\# Params}} &
\multirow{2}{*}{\textbf{FLOPS}}\\
\textbf{Resolution}  &  &  &  &  \\ 
\hline
$8 \times 8$ & $89.32$ & $87.72$ & $  4.3$M & $ 12.4$G\\
\hline
$16 \times 16$ & $91.66$ & $90.29$ & $  4.0$M & $ 13.5$G\\
\hline
$32 \times 32$ & $91.70$ & $90.06$ & $  3.8$M & $ 17.8$G\\
\hline
\end{tabular}
\vspace{2ex}
\caption{The results for applying kinematic features update steps at different resolutions.
At $32 \times 32$ resolution the receptive field is not large enough to fully capture all neighbouring keypoints.}
\label{tbl:repose_n_scales}
\end{table}

\begin{table}[h]
\centering
\begin{tabular}{|c|c|c|c|c|}
\hline
\textbf{\# Conv}  &  
\multirow{2}{*}{\textbf{Leeds}} &
\multirow{2}{*}{\textbf{MPII}} &
\multirow{2}{*}{\textbf{\# Params}} &
\multirow{2}{*}{\textbf{FLOPS}}\\
\textbf{Blocks}  &  &  &  &  \\ 
\hline
$1$ & $90.42$ & $89.46$ & $  3.3$M & $ 13.1$G\\
\hline
$2$ & $91.14$ & $89.30$ & $  3.5$M & $ 13.2$G\\
\hline
$3$ & $90.64$ & $89.61$ & $  3.8$M & $ 13.3$G\\
\hline
$4$ & $91.66$ & $90.29$ & $  4.0$M & $ 13.5$G\\
\hline
$5$ & $91.32$ & $90.06$ & $  4.3$M & $ 13.6$G\\
\hline
\end{tabular}
\vspace{1ex}
\caption{The results for different number of convolutional blocks used in the kinematic feature update step.}
\label{tbl:fus_n_convs}
\vspace{-0.5ex}
\end{table}

\begin{table}[h]
\centering
\begin{tabular}{|c|c|c|}
\hline
\textbf{Feature Update  Strategy} &  \textbf{Leeds} & \textbf{MPII}\\
 \hline
trainable & $91.66$ & $90.29$\\
\hline
add & $91.13$ & $89.59$\\
\hline
replace & $90.29$ & $88.94$\\
\hline
\end{tabular}
\vspace{1ex}
\caption{The results for different kinematic features update strategies. Using trainable mixing weights to learn how to weight old vs new features is the clear winner, while insignificantly increasing the number of trainable parameters.}
\label{tbl:fus_update_approach}
\vspace{-0.5ex}
\end{table}

\begin{table}[h]
\centering
\begin{tabular}{|c|c|c|c|c|}
\hline
\textbf{\# Stages} &  \textbf{Leeds} & \textbf{MPII} & \textbf{\# Params} & \textbf{FLOPS} \\
\hline
$1$ & $91.66$ & $90.29$ & $  4.0$M & $ 13.5$G\\
\hline
$2$ & $92.94$ & $91.55$ & $  8.4$M & $ 29.4$G\\
\hline
$4$ & $92.99$ & $91.45$ & $ 17.3$M & $ 61.4$G\\
\hline
\end{tabular}
\vspace{1ex}
\caption{The results for stacking multiple \repose architecture to create a multi-stage network, a la~\cite{stackedhourglass16}.
Our approach achieves comparable results to state-of-the-art methods~\cite{featurepyramids17,avddataugment18} on LSP, while using $66\%$ of the parameters required by~\cite{featurepyramids17,avddataugment18}.} 
\label{tbl:repose_stacking}
\vspace{-4ex}
\end{table}

\begin{table}[h]
\centering
{
\setlength{\extrarowheight}{8pt}
\small\addtolength{\tabcolsep}{-0.8pt}
\begin{tabular}{c|c|c|c|c|c|c|}
\cline{2-6}
& $\sigma$ & \textbf{Leeds} & \textbf{MPII} & \textbf{\# Params} & \textbf{FLOPS} \\
\cline{2-6}
\multirow{2}{*}{\rotatebox[origin=c]{90}{\footnotesize $128 \!\times\! 128$}}& 
5 & $91.66$ & $90.29$ & $  4.0$M & $ 13.5$G \\
\cline{2-6}
& 10 & $90.25$ & $89.49$ & $  4.0$M & $ 13.5$G \\
\cline{2-6}
\multirow{2}{*}{\rotatebox[origin=c]{90}{\footnotesize $256 \!\times\! 256$}}& 
5 & $90.76$ & $89.96$ & $  4.3$M & $ 49.5$G \\
\cline{2-6}
& 10 & $91.77$ & $90.83$ & $4.3$M & $49.5$G \\
\cline{2-6}
\end{tabular}
}
\vspace{2ex}
\caption{The results for different input resolutions and ground truth heatmap $\sigma$'s using \reposenospace.
Increasing the input image resolution leads to better results on MPII, but also increases the FLOPS by a factor of $3.7$.}
\label{tbl:res_sigma}
\end{table}

\paragraph{Post-features Update Predictions} As described in Section \ref{sc:method}, we independently predict one heatmap form each post-update feature sets. This configuration results in a $4M$ \repose model. Alternatively, jointly predicting $K$ heatmaps from the projected concatenation of all the post-update features reduced the model to $3.5M$ but degraded performance by $0.45\%$ and $0.35\%$ on LSP and MPII, respectively.

\paragraph{Input Image Resolution \& Ground Truth Heatmaps}
We tried two different values for $\sigma$, namely~$\{5, 10\}$, which is used in generating ground truth heatmaps.
We also tried two different input image resolutions, $128 \times 128$ and $256 \times 256$, but applied the kinematic features updates at the $16 \times 16$ resolution for both configurations.

On the one hand, as shown in Table~\ref{tbl:res_sigma}, increasing the resolution leads to an increase in performance by $0.55\%$, but on the other hand FLOPS increased by a factor of $3.7$.

\section{Conclusion}
\label{sc:conclusion}
We presented a novel lightweight model for pose estimation from a single image. Our model combines two main components to achieve competitive results at its scale: 1) a learned deep geometric prior that intuitively encourages predictions to have consistent configurations, and 2) hierarchical refinement of predictions through a multi-scale representation of the input image; both trained jointly and in an end-to-end fashion. Compared with various state-of-the-art models, our approach has a fraction of the parameter count and the computational cost, and achieves state-of-the-art results on a standard benchmark for models of its size. 

We carried out extensive ablation studies of our model components, evaluating across input resolutions, number of scales, and types of kinematic updates, among others, to provide a detailed report of the impact of the various design choices. Finally, recent state-of-the-art approaches to pose estimation incorporate adversarial loss or distillation, both of which are orthogonal to our contribution and will likely improve our model, which we leave to future work.

{\small
\bibliographystyle{ieee_fullname}
\bibliography{egbib}
}

\end{document}